\documentclass[a4paper,fleqn]{cas-dc}

\usepackage[authoryear]{natbib}
\usepackage{todonotes}

\def\tsc#1{\csdef{#1}{\textsc{\lowercase{#1}}\xspace}}
\tsc{WGM}
\tsc{QE}

\ExplSyntaxOn
\cs_gset:Npn \__first_footerline: { 
	\group_begin: 
	\small \sffamily \__short_authors: 
	\group_end: 
}
\ExplSyntaxOff

\begin{document}
\let\WriteBookmarks\relax
\def\floatpagepagefraction{.6}
\def\textpagefraction{.1}

\shorttitle{FiLMMeD: Feature-wise Linear Modulation for Cross-Problem MDVRP}    

\shortauthors{A. Corrêa et al.}  

\title [mode = title]{FiLMMeD: \underline{F}eature-wise \underline{L}inear \underline{M}odulation for Cross-Problem \underline{M}ulti-\underline{D}epot Vehicle Routing}



%

\author[1]{Arthur Corrêa}[orcid=0009-0000-4949-4791]
\cormark[1]
\ead{ajpcorrea@dem.uc.pt}
\credit{Conceptualization, Formal analysis, Investigation, Methodology, Software, Validation, Visualization, Writing - original draft, Writing - review \& editing}

\author[1]{Paulo Nascimento}[orcid=0000-0003-4139-1100]
\ead{pnascimento@dem.uc.pt}
\credit{Formal analysis, Project administration, Supervision, Writing - review \& editing}

\author[1]{Samuel Moniz}[orcid=0000-0002-7813-4514]
\ead{samuel.moniz@dem.uc.pt}
\credit{Funding acquisition, Project administration, Supervision, Writing - review \& editing}

\affiliation[1]{organization={University of Coimbra, CEMMPRE, ARISE},
	addressline={Department of Mechanical Engineering}, 
	city={Coimbra},
	postcode={3030-788}, 
	country={Portugal}}

\cortext[1]{Corresponding author}


\begin{abstract}
Solving practical multi-depot vehicle routing problems (MDVRP) is a challenging optimization task central to modern logistics, increasingly driven by e-commerce. To address the MDVRP's computational complexity, neural-based combinatorial optimization methods offer a promising scalable alternative to traditional approaches. However, neural-based methods typically rely on rigid architectures and input encodings tailored to specific problem formulations. In real-world settings, heterogeneous constraints create multiple MDVRP variants, limiting the applicability of such models. While multi-task learning (MTL) has begun to accelerate the development of unified neural-based solvers, prior works focus almost exclusively on single-depot VRPs, leaving the MDVRP unaddressed. To bridge this gap, we propose \underline{F}eature-wise \underline{L}inear \underline{M}odulation for Cross-Problem \underline{M}ulti-\underline{D}epot Vehicle Routing (FiLMMeD), a novel unified neural-based model for 24 different MDVRP variants. We introduce three main contributions: (1) to improve the model's generalization, we augment the standard Transformer encoder with Feature-wise Linear Modulation (FiLM), which dynamically conditions learned internal representations based on the active set of constraints; (2) we provide an initial demonstration of Preference Optimization in the MTL setting, establishing it as a superior alternative to Reinforcement Learning for future MTL works; (3) to mitigate the generalization gap caused by the introduction of multi-depot constraints, we introduce a targeted curriculum learning strategy that progressively exposes the model to increasingly more complex constraint interactions. Extensive experiments on 24 MDVRP variants (including 8 novel formulations) and 16 single-depot VRPs confirm the effectiveness of FiLMMeD, which consistently outperforms state-of-the-art baselines. Our code is available at: \url{https://github.com/AJ-Correa/FiLMMeD/tree/main}
\end{abstract}




\begin{keywords}
 Vehicle routing \sep Multi-task learning \sep Cross-problem generalization \sep Curriculum learning \sep Preference optimization
\end{keywords}

\maketitle

\section{Introduction}

The vehicle routing problem (VRP) belongs to a class of canonical combinatorial optimization (CO) problems that hold wide importance in both the operations research and computer science communities. VRPs arise in numerous real-world applications, such as logistics and drone delivery \citep{Cattaruzza2017,Li2019,Wang2019}. On one hand, due to the inherent NP-hard nature of the VRP, obtaining optimal solutions with exact algorithms is typically intractable for instances with more than a couple hundred nodes \citep{Pessoa2020}. On the other hand, heuristic algorithms offer a practical compromise, achieving (near-)optimal solutions within manageable runtimes, but their design and tuning often depend on specialized expert knowledge. Recently, neural-based approaches have been gaining traction as an alternative approach to solve VRPs \citep{Bengio2021,Bogyrbayeva2024}. These methods leverage deep learning to approximate powerful policies, capable of finding high-quality solutions with minimal computational overhead and domain expertise. However, like conventional approaches, most existing neural-based methods are tailored and trained for specific types of VRP variants, which limits their applicability across diverse scenarios \citep{Kool2019,Kwon2020,Zhou2023}. 

\subsection{Motivation}

Among existing VRP variants, the multi-depot VRP (MDVRP) is of particular practical relevance. It extends the classic VRP by considering multiple depots from where vehicles are dispatched. This setting has become increasingly important with the rise of e-commerce and growing customer expectations for faster deliveries, prompting companies to operate multiple depots \citep{Lyu2023}. Despite that, the MDVRP has been largely neglected by the neural-based CO literature, where most efforts have focused on the classic single-depot VRP. Furthermore, considering the sheer amount of existing VRP variants \citep{Elatar2023}, each defined by distinct sets of operational constraints, designing and training separate models for each one is impractical.

Crucially, in real-world logistics, these constraints are rarely static. A delivery fleet might face different daily requirements, from varying depot availability and time windows to backhaul requests and route length limits \citep{Pillac2013} — depending on customer needs, seasonality factors or regulatory changes. Notwithstanding this fact, traditional solution methods are typically designed for fixed problem structures and constraints \citep{Braekers2016}, lacking the flexibility to adapt to different operating settings, often requiring a separate method for each scenario. Consequently, there is an existing need for a single unified routing model capable of solving diverse constraint combinations without being retrained each time.

To address this, the literature has recently started to move towards the development of unified cross-problem models through the application of multi-task learning (MTL) \citep{Liu2024,Zhou2024,Berto2025}. MTL is a machine learning paradigm that allows a single model to learn from multiple related tasks simultaneously, leveraging shared representations to capture common structural patterns across them \citep{Zhang2022}. In the context of VRPs, this allows a model to generalize across different problem variants without having to be retrained for each formulation. In spite of recent efforts, prior MTL studies have not explicitly targeted MDVRP variants. While RouteFinder \citep{Berto2025} and CaDA \citep{Li2025} have addressed MDVRPs to some extent, they operate in a zero- or few-shot learning manner, leading to substantial performance gaps relative to traditional non-neural baselines. In addition, existing MTL models do not structurally condition learned representations based on the active constraints of each VRP variant. Although CaDA introduced constraint prompts, it relies on static feature concatenation that treats constraints as high-level context rather than directly modulating node embeddings. As a result, existing models still fail to distinguish the structural requirements of each variant, inducing gradient conflicts that limit their generalization across diverse constraint sets. Lastly, prior MTL works have relied on relatively simple training strategies. For example, MTPOMO \citep{Liu2024} and MVMoE \citep{Zhou2024} train almost exclusively on variants with a single active constraint. Conversely, RouteFinder \citep{Berto2025} samples all available variants simultaneously, regardless of the number of active constraints. While these strategies work well for single-depot VRPs, our experiments have shown that they are insufficient to achieve good generalization on MDVRP variants.

\subsection{Our contributions}

To mitigate the aforementioned challenges, this work proposes \underline{F}eature-wise \underline{L}inear \underline{M}odulation for Cross-Problem \underline{M}ulti-\underline{D}epot Vehicle Routing (FiLMMeD), the first MTL model explicitly targeting the MDVRP, to the best of our knowledge. Our main contributions are summarized as follows:
\begin{itemize}
	\item We introduce an MTL model capable of handling a wide range of MDVRP variants, including 8 new variants with inter-depot routes. In total, FiLMMeD can solve 24 distinct MDVRP variants, including backhauls, open routes, route length limits, time windows, inter-depot routes, and possible combinations between these constraints.
	\item We introduce a feature-wise linear modulation (FiLM) mechanism \citep{Perez2018} to condition node embeddings directly on the constraint attributes of each instance. FiLM is a conditioning technique that allows neural networks to adapt their internal representations based on auxiliary information. In our case, the conditioning input is a Boolean attribute vector indicating the presence or absence of each constraint. 
	\item To solve single-depot VRP variants with FiLMMeD, we fine-tune and train it using preference optimization (PO) instead of reinforcement learning (RL). PO is an emerging alternative to RL that has seen some applications to CO recently \citep{Pan2025,Liao2025}. As far as we know, this is the first application of PO in an MTL setting, being an initial demonstration of its potential for future MTL works. Through various experiments, we show that PO significantly outperforms the traditional RL training of MTL models, having the potential to become the standard for future MTL works.
	\item We propose a lightweight yet effective curriculum learning (CL) strategy for the MDVRP, which structures training by gradually increasing the number of active constraints, leading to improved generalization. We show that, unlike single-depot VRPs where uniformly sampling suffices, the MDVRP may induce more complex interactions between constraints, requiring a guided curriculum for effective generalization.
\end{itemize}

We report extensive experimental results on 24 MDVRP variants and 16 single-depot VRP variants, with FiLMMeD achieving state-of-the-art performance and surpassing existing MTL baselines. Ablation studies confirm the effectiveness of the proposed FiLM mechanism, CL training regimen and PO algorithm. 

The rest of this paper is organized as follows. Section~\ref{related_work} discusses the relevant literature. Section~\ref{preliminaries} establishes important background about our work. Section~\ref{methodology} details the main architecture and methodology in this study. Section~\ref{experimental_results} displays all computational experiments performed. Finally, Section~\ref{conclusion} draws conclusions and possible future work directions.

\section{Related Work}
\label{related_work}

\subsection{Neural-based CO for VRPs}

Recently, neural-based methods have been growing as a promising third paradigm for CO problems, complementing (meta-)heuristics and exact algorithms \citep{Bengio2021,Mazyavkina2021,Bogyrbayeva2024}. Generally speaking, neural-based methods are, most of the time, \textit{construction-based}, focusing on learning policies to incrementally build a solution in an end-to-end manner. One of the first prominent examples are Pointer Networks, a sequence-to-sequence method used to construct solutions autoregressively for the traveling salesman problem (TSP) \citep{Vinyals2015}. Pointer Networks marked an important paradigm shift in the literature, being one of the first demonstrations of the potential for neural-based CO solvers. Despite showing competitive performance on the TSP, it relied on supervised learning for training. Considering the NP-hard nature of many CO problems, computing (near)-optimal labels is non-trivial, especially for larger instances. Thus, to circumvent this, following works focused on training with RL instead of supervised learning \citep{Bello2017,Nazari2018}. Later, \citet{Kool2019} made a substantial contribution to the literature, developing an attention-based model (AM) inspired by the Transformer architecture \citep{Vaswani2017}. The AM is, up to this day, the inspiration and backbone for most existing VRP model architectures in the literature. After, \citet{Kwon2020} introduced POMO, a new training regimen for neural-based models that exploits the symmetries inherent in CO problems. Their key contributions are twofold. First, they proposed REINFORCE with shared baselines, an algorithm that generates multiple trajectories per instance, leveraging different solution starting nodes to compute a low-variance baseline. This stabilizes training and yields substantially faster convergence than the conventional REINFORCE algorithm. Second, they developed a lightweight instance augmentation technique using reflections and rotations. Together, both techniques have become a standard across almost all neural-based CO works, influencing most subsequent studies \citep{Bi2025,Chalumeau2023,Correa2026,Drakulic2023,Grinsztajn2023,Jiang2026,Lei2022,Kim2022,Zhou2023,Zhou2024b}.

Other less common approaches can be classified as \textit{improvement-based}, which learn to iteratively refine an initially generated solution through local search operators \citep{Chen2019neurewriter,Hottung2020,Ma2021dual,Hudson2022,Roberto20202-opt,Wu2022improvement,Ma2023search}. Typically, improvement-based methods find higher-quality solutions than construction-based, but at the cost of significantly slower inference. Lastly, \textit{divide-and-conquer} approaches have also been showing promise lately, learning to solve VRPs by decomposing an instance into multiple smaller sub-instances \citep{Cheng2023select,Fu2021generalize,Hou2023generalize,Kim2021collaborative,Li2021delegate,Ye2024glop,Zheng2024udc,Zong2022rbg}. However, these methods may suffer from errors at sub-problem boundaries and increased training and implementation complexity.

\subsection{Multi-task learning for VRPs}

In recent years, the neural-based CO community has started to investigate the application of MTL to improve the generalization across different VRP variants. \citet{Liu2024} introduced MTPOMO, the first MTL-based approach for VRPs capable of solving 16 different variants through attribute composition. This work established a new paradigm for cross-problem learning, significantly influencing later MTL developments. After, \citet{Zhou2024} proposed MVMoE, an approach built upon MTPOMO which incorporates mixture-of-experts in its architecture. In a mixture-of-experts model, the network is divided into multiple specialized sub-networks, called experts, each designed to learn distinct patterns or features. Then, a gating network selects the most relevant experts to be activated, depending on the given input. Another significant step forward is RouteFinder \citep{Berto2025}, which encompasses several technical innovations. RouteFinder adopts a modern Transformer-based architecture, including root mean square normalization \citep{Zhang2019rms} and SwiGLU activations \citep{Shazeer2020}. It also employs mixed batch training, a technique that allows multiple different VRP variants to be processed simultaneously within the same batch, which leads to more stable training. \citet{Li2025} presented CaDA, incorporating a dual-attention mechanism that adds a new multi-head attention module to the encoder architecture, focused on capturing the information of closely related nodes. Specifically, this is achieved using top-\textit{k} sparse attention, where each node only attends to its \textit{k} most relevant neighbors. The aforementioned studies are considered as the backbone of MTL VRP research, inspiring most following works.

\citet{Wang2025spsm} introduced a soft-parameter-sharing model to balance task-specific and shared representations across different variants. Although improving performance over MTPOMO and MVMoE, its multiple separate Multi-Head Attention modules introduce significant more computational burden and GPU memory footprint. \citet{Huang2025rethinking} presented ReLD, showing that adding a simple identity mapping with a feed-forward layer to the decoder can significantly enhance its capacity. ReLD surpasses state-of-the-art performance on single- and multi-task settings, providing a great lightweight alternative to expensive heavy decoder Transformer models. \citet{Goh2025shield} proposed SHIELD, improving robustness to both cross-problem and cross-distribution generalization through hierarchical and sparsity mechanisms. More recently, \citet{Pan2025decomposable} proposed MoSES, which reformulates VRP variants through a State-Decomposable Markov Decision Process and employs specialized LoRA experts with adaptive gating. Although it achieves strong performance, MoSES requires training and maintaining multiple task-specific expert networks, substantially increasing computational cost and memory footprint compared to previous architectures. In an effort to model the highly varied topologies inherent to different VRP variants, \citet{Liu2025curvature} introduced a novel pre-training framework based on mixed curvature. Rather than relying on a single geometric space, their approach maps nodes into a composite manifold that fuses Euclidean, spherical, and hyperbolic dimensions, providing the architectural flexibility required to capture the diverse underlying topological structures of different variants. Finally, \citet{Zheng2025mtlkd} presented MTL-KD, leveraging knowledge distillation for a better cross-size generalization. 

\textbf{Remark.} Recent MTL advances have improved cross-problem generalization for neural VRP solvers, yet significant gaps remain. First, existing approaches focus almost exclusively on single-depot variants, leaving the multi-depot setting --- critical for modern logistics --- largely unaddressed. Second, while some works incorporate constraint information through static feature concatenation or prompt-based conditioning, they fail to dynamically modulate internal representations based on active constraint combinations. This approach exacerbates gradient interference across diverse variants, restricting generalization. Third, prior training strategies (whether uniformly sampling single-constraint variants or simultaneous exposure to all variants), while effective on the single-depot setting, have shown limited generalization on the MDVRP. Lastly, the reliance on RL-based training in MTL settings limits convergence stability and solution quality, suffering from gradient interference and diminishing reward signals, as noted by \citet{Pan2025}. These limitations motivate our contributions, with our work diverging from previous efforts in multiple key aspects: we explicitly focus on the underexplored multi-depot setting, which has received limited attention in neural-based research; we introduce a lightweight FiLM mechanism to explicitly condition the learned representations on the active combination of problem constraints, showing that it effectively disentangles the latent space of node embeddings, leading to improved generalization; we design a dedicated CL training regimen for the multi-depot setting, which gradually increases the complexity of training by the number of active constraints in the variants sampled, outperforming prior training strategies; finally, we employ PO to fine-tune and train FiLMMeD model on single-depot variants, demonstrating that it significantly outperforms the conventional REINFORCE loss function, establishing it as a promising direction for future MTL works.

\section{Preliminaries}
\label{preliminaries}

\subsection{Problem definition}

\begin{figure*}[pos=h]
	\centering
	\includegraphics[width=\textwidth]{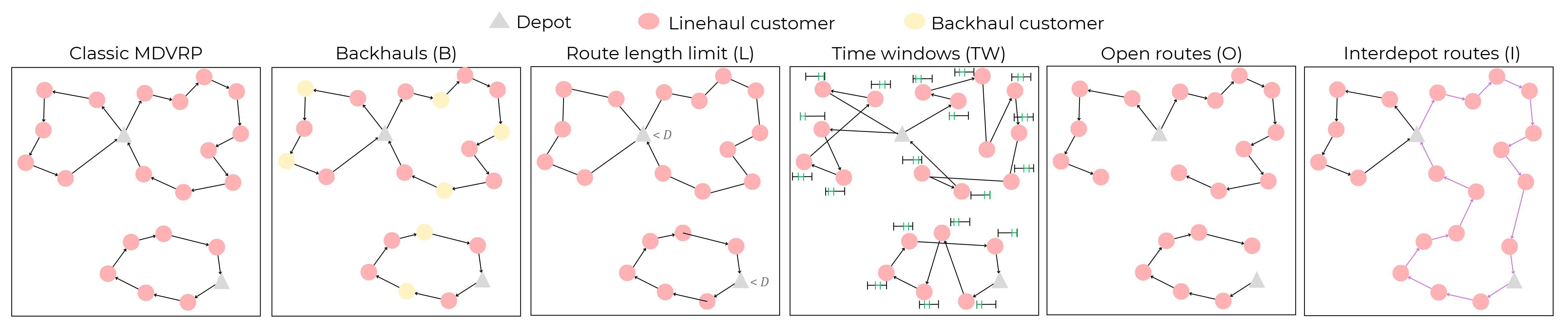}
	\caption{MDVRP constraints addressed in our work.}
	\label{mdvrp_variants}
\end{figure*}

In this study, we solve 24 different MDVRP variants, whose formulations we describe in this subsection. A classic MDVRP instance is defined on a graph $\mathcal{G} = \{\mathcal{V}, \mathcal{E}\}$. Here, $\mathcal{V} = \{v_0, ..., v_{m-1}, v_m, ..., v_{m+n-1}\}$ is the set of nodes in the problem, where $\mathcal{V}_d = \{v_0, ..., v_{m-1}\}$ denotes $m$ depots and $\mathcal{V}_c = \{v_m, ..., v_{m+n-1}\}$ denotes $n$ customers. The edge set $\mathcal{E} = \{e_{ij}: i, j \in \mathcal{V}, i \neq j, (i, j) \notin \mathcal{V}_d \times \mathcal{V}_d\}$ contains all connections between distinct nodes, excluding direct depot-to-depot edges. Each edge is associated with a travel cost $c_{ij}$. At each depot, a fleet of vehicles is available, where each vehicle has a capacity limit $C$. A fundamental constraint across all MDVRP variants is that every customer $v_i$ has a demand $\delta_i$, which must be satisfied by exactly one vehicle dispatched from one of the $m$ depots. The solution $\tau$ represents the sequence of nodes visited in the problem, and is composed of multiple individual routes, each corresponding to the path followed by a single vehicle. Once a vehicle completes its route, it must return to the depot from which it departed. A solution is feasible if all customer demands are satisfied and no vehicle exceeds the capacity limit on its assigned route. The objective is to find the optimal solution $\tau^*$ that minimizes the total travel distance across all vehicles.

Besides the classic MDVRP, we also solve other variants by adding constraints on top of the original problem formulation. In our work, we explore five additional constraints, four of which have been addressed in prior MTL works, but mostly focused on single-depot VRPs \citep{Liu2024,Zhou2024,Berto2025}. We also consider another constraint specific to MDVRPs, which are inter-depot routes \citep{Crevier2007,Ramos2020}, leading to 8 additional MDVRP variants previously unsolved in neural-based works. A brief description of the five additional constraints is provided below, and an illustration is provided in Figure~\ref{mdvrp_variants}.

\textbf{Backhaul (B):} In the standard MDVRP, a vehicle’s load decreases whenever it visits a customer. In the MDVRP with backhauls (MDVRPB), some customers require pickups instead of deliveries, causing the vehicle’s load to increase. These customers are called backhaul customers, whereas traditional delivery customers are linehaul customers. In this paper, we adopt two different backhaul settings. Across the experiments compared to \citet{Zhou2024}, we use the mixed backhaul, in which a single vehicle may alternate between serving linehaul and backhaul customers without any strict precedence order. For experiments compared to \citet{Berto2025}, a single vehicle must not alternate between back and linehaul customers, i.e., it may only visit backhaul customers after completing all its linehaul visits.

\textbf{Route length limit (L):} In the MDVRP with route length limits (MDVRPL), the length/duration of each vehicle’s route must not exceed a prescribed threshold $D$.

\textbf{Open route (O):} In the MDVRP with open routes (MDOVRP), vehicles are not required to return to their departure depot after completing their respective routes. Therefore, each route ends at the last visited customer.

\textbf{Time window (TW):} In the MDVRP with time windows (MDVRPTW), each node $v_i \in \mathcal{V}$ has an associated time window $[e_i, l_i]$ and a service time $s_i$. A vehicle visiting node $v_i$ must arrive no later than $l_i$. Service can only begin within the time window $[e_i,l_i]$, requiring a wait if the arrival is earlier than $e_i$. All depots are also assigned an ending time window, requiring vehicles to arrive at the depot before it.

\textbf{Inter-depot routes (I):} The MDVRP with inter-depot routes (MDVRPI) is a relatively less-studied variant of the MDVRP. In this setting, vehicles are allowed to stop at any depot during a route to reload their capacity $C$ (hence the term inter-depot routes). However, each vehicle must still return to its original departing depot at the end of its route. Consistent with the general problem definition, direct depot-to-depot edges are excluded. The original MDVRPI formulation also includes route length limit constraints \citep{Crevier2007,Ramos2020}. However, in our work, we relax this requirement to allow for a broader set of MDVRPI variants. Nonetheless, some of the MDVRPI variants we consider do include route length limits (see Table~\ref{mdvrp_constraint_combinations}). In total, we solve eight MDVRPI variants, whereas only four would be possible without relaxing the route length limit constraint.

The combination of the aforementioned constraints lead to different MDVRP variants, which we detail in Table~\ref{mdvrp_constraint_combinations}.

\begin{table}[h]
	\caption{24 MDVRP variants with five constraints.}
	\label{mdvrp_constraint_combinations}
	\vskip 0.1in
	\begin{center}
		\begin{small}
			\renewcommand\arraystretch{1.1}
			\resizebox{\columnwidth}{!}{ 
				\begin{tabular}{l|ccccc}
					\toprule
					& Open Route (O) & Backhaul (B) & Duration Limit (L) & Time Window (TW) & Inter-depot routes (I) \\
					\midrule
					MDVRP & & & & & \\
					MDOVRP & \checkmark & & & & \\
					MDVRPB & & \checkmark & & & \\
					MDVRPL & & & \checkmark & & \\
					MDVRPTW & & & & \checkmark & \\
					MDOVRPTW & \checkmark & & & \checkmark & \\
					MDOVRPB & \checkmark & \checkmark & & & \\
					MDOVRPL & \checkmark & & \checkmark & & \\
					MDVRPBL & & \checkmark & \checkmark & & \\
					MDVRPBTW & & \checkmark & & \checkmark & \\
					MDVRPLTW & & & \checkmark & \checkmark & \\
					MDOVRPBL & \checkmark & \checkmark & \checkmark & & \\
					MDOVRPBTW & \checkmark & \checkmark & & \checkmark & \\
					MDOVRPLTW & \checkmark & & \checkmark & \checkmark & \\
					MDVRPBLTW & & \checkmark & \checkmark & \checkmark & \\
					MDOVRPBLTW & \checkmark & \checkmark & \checkmark & \checkmark & \\
					MDVRPI & & & & & \checkmark \\
					MDVRPIB & & \checkmark & & & \checkmark \\
					MDVRPIL & & & \checkmark & & \checkmark \\
					MDVRPITW & & & & \checkmark & \checkmark \\
					MDVRPIBL & & \checkmark & \checkmark & & \checkmark \\
					MDVRPIBTW & & \checkmark & & \checkmark & \checkmark \\
					MDVRPILTW & & & \checkmark & \checkmark & \checkmark \\
					MDVRPIBLTW & & \checkmark & \checkmark & \checkmark & \checkmark \\
					\bottomrule
			\end{tabular}}
	\end{small}
\end{center}
\end{table}

\subsection{Learning to solve MDVRPs}

Most neural-based models generate solutions autoregressively by learning a policy $\pi_{\theta}$ through an attention-based neural network \citep{Kool2019,Kwon2020,Liu2024,Zhou2024}. We adopt this approach, utilizing an encoder-decoder framework, where a complete solution is built timestep by timestep by framing the MDVRP as a Markov decision process (MDP). The process begins with the encoder embedding an input instance $\mathcal{G}$ based on its underlying features, such as node coordinates, demands and time windows. A decoder then constructs the solution autoregressively. At each timestep $t$, it receives a state, from which it computes the probability of selecting each candidate node next to extend the partial solution.

Similar to prior neural MDVRP models \citep{Li2024mdmta,Correa2026}, FiLMMeD constructs solutions by building one route at a time. To encourage systematic exploration of the solution space, the initial action is chosen deterministically following the POMO framework \citep{Kwon2020}. For example, in an instance with 50 customers and 3 depots, there may be up to 150 possible customer–depot initializations, resulting in 150 distinct trajectories (i.e., solutions) evaluated in parallel.

After that, the model sequentially constructs the current route by selecting the next customer to be served, until it decides to end the route, which occurs when the agent selects the departure depot. Upon terminating the route, the agent selects the depot from where the next route will begin (which can be the same depot or not). The process repeats until a full solution is constructed. This decision-making sequence can be formalized as an MDP, whose individual components we describe below:

\textbf{State:} At each timestep $t$, the agent observes a state $s_t$, which represents the current partial MDVRP solution. The state comprises the embeddings of the previously selected node together with state variables carrying information about the partial solution.

\textbf{Action:} After observing the state $s_t$, the agent selects an action $a_t \in \mathcal{A}_t$, where $\mathcal{A}_t$ denotes the set of feasible actions at timestep $t$. Here, $a_t$ can be of two types, as discussed before: 1) a depot node, which is chosen to either start a new route or terminate the current one. In the case of variants with inter-depot routes, a depot node (one that is different from the starting depot) can be chosen throughout a route to replenish the vehicle’s capacity; 2) a customer node, which represents the next node to be visited by the vehicle in the current route.

\textbf{State transition:} After executing $a_t$, the environment transitions to a new state $s_{t+1}$ reflecting the updated partial solution. The vehicle’s capacity $C_t$, elapsed time $T_t$ and route length $D_t$ are updated accordingly, and its location is set to the coordinates of the selected node. The feasibility mask is then recomputed to exclude visited nodes, as well as any unvisited nodes that are no longer reachable due to the tighter constraints induced by the chosen action.

\textbf{Reward:} The reward $r(\tau, \mathcal{G})$ is computed after an entire solution $\tau$ is constructed for instance $\mathcal{G}$. It is defined as the negative sign of the total distance traveled by all vehicles, meaning that maximizing the expected reward corresponds to minimizing the total travel distance.

\textbf{Policy:} Throughout training, the agent learns a decision-making policy $\pi_\theta$ parameterized by an attention-based neural network. At each timestep $t$, given the current state $s_t$, the policy outputs a probability distribution over feasible next nodes, guiding the choice of action $a_t$. Then, the agent chooses the next action greedily or by stochastic sampling, according to this distribution.

\section{Methodology}
\label{methodology}

In this section, we present the architecture of our approach and important methodological details. We first present an overview of our model, including the FiLM mechanism and the rationale behind it. Then, we present the components of our training regimen, including the CL strategy and the PO algorithm.

\subsection{FiLMMeD architecture}

FiLMMeD follows the general encoder-decoder architecture of previous Transformer-based MTL models \citep{Liu2024,Zhou2024}. In this work, we apply FiLMMeD modifications into three base architectures: FiLMMeD-MTPOMO, FiLMMeD-MVMoE, and FiLMMeD-CaDA. To streamline the description, we primarily use MTPOMO and MVMoE as references since they share a similar model pipeline. For specific architectural details regarding the FiLMMeD-CaDA model, we refer readers to \citet{Li2025}. We note, however, that the FiLM mechanism is integrated into CaDA in the exact same manner described for the other models. Furthermore, our core contributions remain modular, i.e., they can be seamlessly adapted to a wide variety of other neural-based MTL architectures.

\subsubsection{Encoder}

The encoder starts by processing a given MDVRP instance $\mathcal{G}$. First, linear layers are used to project the static features of each node in the instance onto a $d$-dimensional embedding space. Each depot node $v_j \in \mathcal{V}_d, \forall j \in \{0, ..., m-1\}$ is characterized by its two-dimensional coordinates. The features of each customer node $v_i \in \mathcal{V}_c, \forall i \in \{m, ..., m+n-1\}$ include the two-dimensional coordinates, demand, early and late time windows, and service time. Let $h_{j}^{(d)}$ and $h_{i}^{(c)}$ denote the initial embeddings of depot and customers nodes, respectively. First, two separate linear layers are used to process each set of features from depots and customers, as follows:
\begin{equation}
	h_{j}^{(d)} = W_0 x_{v_j}^{(d)} + b_0, \ \forall v_j \in \mathcal{V}_d
\end{equation}
\begin{equation}
	\label{eq_linear_customer}
	h_{i}^{(c)} = W_1 x_{v_i}^{(c)} + b_1, \ \forall v_i \in \mathcal{V}_c
\end{equation}
where $x_{v_j}^{(d)} \in \mathbb{R}^{2}$ denotes the feature vector of depot $v_j$, $x_{v_i}^{(c)} \in \mathbb{R}^{6}$ denotes the feature vector of customer $v_i$, $W_0 \in \mathbb{R}^{d \times 2}$ and $W_1 \in \mathbb{R}^{d \times 6}$ are learnable weight matrices, and $b_0, b_1 \in \mathbb{R}^{d}$ are bias vectors.

After that, we modulate the customer embeddings using FiLM \citep{Perez2018}. FiLM is a conditioning technique that allows a neural network to conditionally modify an embedding based on external information. Here, each customer embedding $h_{i}^{(c)}$ is explicitly modulated according to a Boolean conditioning vector $z \in \mathbb{R}^5$, which indicates the active constraints in the input instance, that is, whether an instance contains B, L, O, TW or I attributes. More specifically, FiLM computes feature-wise scaling ($\gamma$) and feature-wise shifting ($\beta$) parameters based on the conditioning vector:
\begin{equation}
	\label{film_eq1}
	\gamma = f_{\gamma} (z) \in \mathbb{R}^d, \ \ \ \ \beta = f_{\beta} (z) \in \mathbb{R}^d
\end{equation}
where $f_{\gamma}$ and $f_{\beta}$ are learnable linear projections.

Each customer embedding is then modulated as an affine transformation:
\begin{equation}
	\label{film_eq2}
	FiLM (h_{i}^{(c)} | \gamma, \beta) = \gamma h_{i}^{(c)} + \beta.
\end{equation}

Here, $\gamma$ serves a scaling parameter that controls the importance of each node in response to the conditioning vector. For example, if time windows are active, $\gamma$ might emphasize the embedding channels related to early and late time windows, making the model more sensitive to time windows constraints during subsequent transformations. $\beta$ acts as a bias parameter that shifts each embedding by an offset determined by the active set of constraints. This allows the model to adapt the learned representations of each node based on its context, instead of treating all embeddings identically regardless of the variant solved.

After that, the result embeddings are concatenated with the previously processed depot embeddings, resulting in $h^{(0)}_{i}, \ \forall v_i \in \mathcal{V}$. This sequence is then processed by $L$ encoder layers. Each layer $\ell \in \{1, \dots, L\}$ consists of an MHA sub-layer followed by a node-wise Feed-Forward Network (FFN), with residual connections and Instance Normalization (IN) before and after the FFN.

In the MHA sub-layer with $A$ heads, for each node $v_i$ and head $a \in \{1, \dots, A\}$, we compute query ($Q$), key ($K$), and value ($V$) vectors: 
\begin{equation} Q^{(\ell,a)}_i = W^{(\ell,a)}_Q h^{(\ell-1)}_i
\end{equation} 
\begin{equation} K^{(\ell,a)}_i = W^{(\ell,a)}_K h^{(\ell-1)}_i
\end{equation} 
\begin{equation} V^{(\ell,a)}_i = W^{(\ell,a)}_V h^{(\ell-1)}_i 
\end{equation} 
where $W^{(\ell,a)}_Q, W^{(\ell,a)}_K, W^{(\ell,a)}V \in \mathbb{R}^{d_k \times d}$ are learnable projection matrices, and $d_k = d/A$.

The attention weight between nodes $v_i$ and $v_j$ is computed as: 
\begin{equation} w^{(\ell,a)}_{ij} = \text{Softmax} \left( \frac{(Q^{(\ell,a)}_i)^\top K^{(\ell,a)}_j}{\sqrt{d_k}} \right). 
\end{equation}

This weight represents the influence of node $v_j$ on node $v_i$ within the $a$-th head.  The output of head $a$ for node $v_i$ is a weighted sum of values: $o^{(\ell,a)}_{i} = \sum_{v_k \in \mathcal{V}} w^{(\ell,a)}_{ik} \, V^{(\ell,a)}_k$. The outputs from all heads are concatenated and projected: 
\begin{equation} \text{MHA}(h^{(\ell-1)}_{i}) = W_O \left[ o^{(\ell,1)}_i, \dots, o^{(\ell,A)}_i \right] 
\end{equation} 
where $W_O \in \mathbb{R}^{d \times d}$.

The MHA output is then processed by the first residual and normalization block: 
\begin{equation} \tilde{h}^{(\ell)}_i = \text{IN}\left( h^{(\ell-1)}_i + \text{MHA}(h^{(\ell-1)}_i) \right)
\end{equation} 

Finally, the standard FFN is applied, followed by a second residual and normalization block to produce the layer output: \begin{equation} 
	h^{(\ell)}_i = \text{IN}\left( \tilde{h}^{(\ell)}_i + \text{FFN}(\tilde{h}^{(\ell)}_i) \right) 
\end{equation} 
The FFN consists of two linear transformations with a ReLU activation in between. The description above details the encoder of FiLMMeD-MTPOMO. For FiLMMeD-MVMoE, the only change is that the standard FFN layer is replaced by a mixture-of-experts layer \citep{Zhou2024}.

\subsubsection{Decoder}

After the encoding process, the decoder operates on the final node embeddings $\{h^{(L)}_i\}_{i=0}^{m+n-1}$ produced by the encoder. Like prior MTL architectures, we adopt a heavy encoder and light decoder structure, consisting of 6 encoder layers and a single decoder layer \citep{Liu2024,Zhou2024,Berto2025}. Since the encoder runs only once per instance and the decoder relies on a single MHA module for repeated steps, this design ensures computational efficiency. 

First, the decoder utilizes its own set of learnable projections to compute keys and values based on the encoded node embeddings, which remain static throughout the decoding process: 
\begin{equation} 
	K_i^{dec} = W^{dec}_K h^{(L)}_i, \quad V_i^{dec} = W^{dec}_V h^{(L)}_i, \quad \forall v_i \in \mathcal{V} 
\end{equation} 
where $W^{dec}_K, W^{dec}_V \in \mathbb{R}^{d \times d}$.

At each decoding step $t$, a context feature vector is constructed from the embedding of the previously selected node and the current state variables. The state variables include the vehicle's remaining capacity, the elapsed time, current route length, and Boolean indicators on the presence of open and inter-depot routes. This vector is projected to form the dynamic query $Q_{(t)}^{dec}$: 
\begin{equation} 
	Q_{(t)}^{dec} = W^{dec}_Q \text{Concat}(h^{(L)}_{\tau_{t-1}}, C_t, T_t, D_t, o_t, id_t),
\end{equation}
where $h^{(L)}_{\tau_{t-1}}$ denotes the embedding of the previously selected node.

Then, an MHA mechanism is employed. The query $Q_{(t)}^{dec}$ attends to the static keys $K_i^{dec}$ to aggregate the values $V_i^{dec}$, producing an updated context vector $h_c^{'t}$. Any infeasible node $v_j$ has its attention score set to $-\infty$. Finally, we compute the compatibility score $u_{i}$ for each node $v_i$ using a single-head attention mechanism. This step uses the updated context vector $h_c^{'t}$ as the query and the original encoder embeddings $h_{i}^{(L)}$ as the keys:
\begin{equation}
	u_i =
	\begin{cases}
		\xi \cdot \tanh\left( \frac{(h_{c}^{'t})^T h_{i}^{(L)}}{\sqrt{d_k}} \right), & \text{if } v_i \in \mathcal{A}_t,\\[2mm]
		-\infty, & \text{otherwise.}
	\end{cases}
\end{equation}

Here, $d_k = d/A$ (with $A$ being the number of heads), and $\xi$ is a clipping parameter set to 10 \citep{Bello2017}.  Similar to the MHA step, we mask the compatibilities of infeasible nodes by setting $u_i = -\infty$. Finally, the probability of selecting node $v_i$ at step $t$ is computed via the Softmax function:
\begin{equation} 
	p_i = \text{Softmax}(u_i) = \frac{e^{u_i}}{\sum_{j} e^{u_j}}
\end{equation}

During inference, the nodes are always selected greedily (that is, $\operatorname{argmax}_i p_i$), while during training, they are selected by sampling from the generated distribution.

\subsection{Theoretical motivation behind FiLM}
\label{film_theory}

Next, we provide theoretical arguments for why FiLM conditioning may improve policy learning and generalization across heterogeneous variants. Our analysis centers across two key aspects: 1) the gradient conflicts problem in MTL; 2) compositional generalization.

\textbf{Gradient conflict mitigation:} When training a unified model across multiple (MD)VRP variants, the primary challenge lies in gradient interference \citep{Yu2020}. Essentially, different tasks (in our case, different constraints) induce distinct loss landscapes. For instance,the gradients for variants with time windows, backhauls and open routes generally exhibit a low cosine similarity. Without proper conditioning, gradients from variant $A$ may destructively interfere with those from variant $B$, leading to performance degradation on both, or one of them performing better in detriment of the other.

Formally, let $\mathcal{T}$ denote the set of variants and $\mathcal{L}_\chi(\theta)$ the loss for variant $\chi \in \mathcal{T}$. Standard MTL suffers from conflicting updates when $\langle \nabla_\theta \mathcal{L}_{\chi_1}, \nabla_\theta \mathcal{L}_{\chi_2} \rangle < 0$ for $\chi_1 \neq \chi_2$. FiLM can alleviate this issue by introducing task-specific affine modulations $(\gamma_\chi, \beta_\chi)$ applied to node' representations. These modulations condition the forward pass, allowing different variants to induce distinct feature transformations before gradients are computed. As a result, variant-specific signals can be partially absorbed by the FiLM parameters, reducing the pressure on shared weights to accommodate conflicting objectives.

While this mechanism does not eliminate gradient conflicts in the shared parameters, it provides a  lightweight pathway for task-specific adaptation, enhancing stability and generalization without fully separating the model into distinct experts, which is computationally prohibitive.

\textbf{Compositional generalization:} Another theoretical advantage of FiLM is its ability to generalize to unseen constraint combinations. By factorizing the conditioning vector $z$ into independent constraint indicators, the model learns constraint-specific transformations that compose via superposition. This is similar to what hypernetworks do \citep{Ha2017}, where the adaptation parameters for a new task are generated by a meta-network rather than learned directly. When encountering a novel combination for the first time, the model can approximate the optimal transformation by interpolating between the learned transformations for each active constraint.

\subsection{Training}

\subsubsection{Curriculum learning}
\label{sec_cl}

CL is a training strategy in which the model is exposed to tasks of progressively increasing complexity. By first learning to solve simpler problems, the model acquires fundamental patterns that can later be extended and refined as the task complexity increases. In the context of our work, complexity is determined by the number and combination of active constraints in each MDVRP instance.

We implement CL by progressively expanding the set of variants from which the model learns during training. In the beginning, the curriculum is dominated by single-constraint problems, allowing the model to learn the effect of each constraint in isolation. Concretely, during the first 30\% of epochs, each batch is sampled from one of the following variants: MDVRP, MDOVRP, MDVRPB, MDVRPL, MDVRPTW, MDVRPI, MDOVRPTW, MDVRPBTW, MDVRPITW.

While most of the variants available during the initial phase of the curriculum are single-constraint, we also include three two-constraint variants with time windows (MDOVRPTW, MDVRPBTW, and MDVRPITW). This choice is not arbitrary, but motivated by empirical observations specific to the multi-depot setting: learning solely from variants with a single constraint (like MTPOMO \citep{Liu2024}), led to poor convergence on several MDVRP variants. In particular, variants involving backhauls, open routes, or inter-depot routes showed a very slow convergence, and was further exacerbated when paired with time windows, which strongly interact with other constraints. While typical training strategies have proven effective for single-depot cases \citep{Zhou2024,Berto2025}, we found they achieved sub-par generalization for the MDVRP. More discussion on this issue is provided in Section~\ref{sec_cl_ablation}.

After this initial phase, we expand the curriculum, with all remaining two-constraint variants being introduced next. At 60\% of total training epochs, we add the full set of three-constraint variants, and finally, at 90\%, the remaining ones. As demonstrated in our ablation studies, this strategy proved to improve convergence, and ultimately, generalization across the tested MDVRP variants.

To train our model, we employ the REINFORCE algorithm with shared baselines \citep{Kwon2020}, an RL method designed specifically for CO problems. A description of this algorithm is provided next.

\subsubsection{REINFORCE with shared baselines}

To train the MDVRP models of FiLMMeD, like most neural-based VRP works, we employ the REINFORCE with shared baselines algorithm \citep{Kwon2020}. Each epoch comprises 100,000 instances, which are sampled in batches of size $B$. For each instance $\mathcal{G}_i, \forall i \in \{1, ..., B\}$, the decoder generates $N$ trajectories $\{\tau_i^1, \dots, \tau_i^N\}$. This is achieved by enforcing diverse starting nodes for each trajectory, allowing the model to broadly explore the solution space in a computationally efficient manner. 

By generating multiple parallel rollouts, a low-variance baseline $b_i$ can be calculated for each instance $\mathcal{G}_i$, as the average reward of all its $N$ generated trajectories:
\begin{equation}
	b_i = \frac{1}{N} \sum_{j=1}^{N} R(\tau_i^j, \mathcal{G}_i).
\end{equation}

Then, the policy gradient $\nabla_\theta J(\theta)$ is approximated using the advantage of each trajectory relative to this shared baseline:
\begin{equation}
	\nabla_\theta J(\theta) \approx \frac{1}{B N} \sum_{i=1}^{B} \sum_{j=1}^{N} \left( R(\tau_i^j, \mathcal{G}_i) - b_i \right) \nabla_\theta \log p_\theta(\tau_i^j | \mathcal{G}_i).
\end{equation}
Finally, the model parameters $\theta$ are updated via gradient ascent, with a learning rate $\eta$.

During training, we set the number of parallel trajectories to $N = m+n-1$ \citep{Li2024mdmta,Correa2026}. This results in $N=52$ for 50-node instances and $N=102$ for 100-node instances, since we consider 3 depots. The set of starting nodes includes the last two depots (i.e., depots with indexes 1 and 2) and all available customers. For trajectories starting at a customer node, the vehicle is assumed to depart from the first depot (depot with index 0). This approach allows us to cover a wide range of initial solutions without incurring the computational burden of enumerating all possible starting nodes ($3 \times 50 = 150$). We note that this choice does not impose any structural biases on the final solution, since the model can choose any other depot to anchor subsequent routes during the decoding process.

\subsubsection{Preference optimization}
\label{preference_optimization}

Although FiLMMeD is primarily designed for MDVRP variants, we do not wish to neglect single-depot VRPs, which constitute a large portion of practical routing problems and remain widely used benchmarks. Therefore, we fine-tune the MDVRP-pre-trained model on 16 single-depot VRPs previously explored in MTL research \citep{Zhou2024}. Importantly, we employ PO instead of RL for this fine-tuning stage. We also train additional single-depot models from scratch using PO, under the same experimental settings as \citet{Berto2025}.

Our choice for pre-training MDVRP models using REINFORCE rather than PO was to ensure a fairer assessment against established baselines -- namely, REINFORCE itself and the uniform single-constraint sampling, opposed to our CL strategy. This baseline comparison allowed us to validate FiLMMeD's architectural innovations, that is, FiLM and CL, without introducing confounding effects from a different training paradigm. Nevertheless, we did perform ablation studies on pre-training MDVRP models with PO, which significantly outperformed its REINFORCE counterpart (see Section ~\ref{ablation_po}).

Besides solving single-depot problems, this design choice serves a broader purpose. While PO has recently shown improved generalization and sample efficiency \citep{Pan2025}, its application has been largely limited to non-MTL domains. Our results suggest that PO is not merely a viable alternative to REINFORCE, but a more principled training paradigm for MTL, addressing several of its fundamental limitations. We also demonstrated its broad applicability across a variety of different model architectures, with PO outperforming RL on every occasion (see Section~\ref{po_algo_performance}).

In summary, PO replaces typical numeric rewards with pairwise preferences between candidate solutions. Instead of learning from scalar rewards, the model is optimized from qualitative comparisons, that is, whether one solution is better than another according to a given objective. This approach stabilizes learning, mitigating issues related to diminishing reward signals and promoting a better exploration of the solution search space.

Concretely, at each batch, we randomly select $B$ instances from one of the 16 single-depot VRP variants (see Table~\ref{vrp_constraint_combinations} for a full list of variants). For each instance $\mathcal{G}_i, \ \forall i \in \{1, ..., B\}$, we generate $N$ solutions (setting again $N ) m + n -1$) and compute their ground-truth rewards $r_{1}^{i}, ..., r_{N}^{i}$. These rewards induce conflict-free preference labels, that is, solution $j$ is preferred over solution $k$ whenever $r_j > r_k$. Within each instance, we compute all possible preference label pairs by comparing every solution against every other one. Here, $y_{j, k}^{i}$ denotes the preference label between solutions $j$ and $k$ which is equal to 1 if $r_{j}^{i} > r_{k}^{i}$, and 0 if not.

For each solution $j$, the log-probabilities of the selected actions across the trajectory are accumulated, giving $\text{log} \pi_{\theta} (\tau_{j}^{i})$. The difference of log-probabilities between solutions $j$ and $k$ is then mapped into a preference probability:
\begin{equation}
	p_{\theta} (\tau_{j}^{i} \succ \tau_{k}^{i}) = \sigma(\alpha(\text{log} \pi_{\theta} (\tau_{j}^{i}) - \text{log} \pi_{\theta} (\tau_{k}^{i})))
\end{equation}
where $\alpha$ is an entropy regularization parameter and $\sigma$ is a Sigmoid function. The PO loss is then defined as the negative log-likelihood of the observed preferences within all instances and pairs:
\begin{equation}
	\mathcal{L}_{\text{PO}} = - \frac{1}{B N^2} \sum_{i=1}^{B} \sum_{j=1, \ k=1}^{N} y_{j, k}^{i} \ \text{log} p_{\theta} (\tau_{j}^{i} \succ \tau_{k}^{i})
\end{equation}
Intuitively, this loss pushes the policy to assign greater probabilities to preferred trajectories relative to other solutions within the same instance. Our implementation follows the same algorithm proposed by \citet{Pan2025}, adapted for the MTL setting.

\begin{table}[h]
	\caption{16 VRP variants with four constraints.}
	\label{vrp_constraint_combinations}
	\vskip 0.1in
	\begin{center}
		\begin{small}
			\renewcommand\arraystretch{1.1}
			\resizebox{\columnwidth}{!}{ 
				\begin{tabular}{l|cccc}
					\toprule
					\midrule
					& Open Route (O) & Backhaul (B) & Duration Limit (L) & Time Window (TW) \\
					\midrule
					CVRP & & & & \\
					OVRP & \checkmark & & & \\
					VRPB & & \checkmark & & \\
					VRPL & & & \checkmark & \\
					VRPTW & & & & \checkmark \\
					OVRPTW & \checkmark & & & \checkmark \\
					OVRPB & \checkmark & \checkmark & & \\
					OVRPL & \checkmark & & \checkmark & \\
					VRPBL & & \checkmark & \checkmark & \\
					VRPBTW & & \checkmark & & \checkmark \\
					VRPLTW & & & \checkmark & \checkmark \\
					OVRPBL & \checkmark & \checkmark & \checkmark & \\
					OVRPBTW & \checkmark & \checkmark & & \checkmark \\
					OVRPLTW & \checkmark & & \checkmark & \checkmark \\
					VRPBLTW & & \checkmark & \checkmark & \checkmark \\
					OVRPBLTW & \checkmark & \checkmark & \checkmark & \checkmark \\
					\midrule
					\bottomrule
			\end{tabular}}
	\end{small}
\end{center}
\end{table}

\textbf{Why PO for Multi-task Learning?} One of the most crucial challenges for the application of MTL in CO lies in the gradient interference across different tasks. Simply put, different VRP variants induce conflicting parameter updates due to heterogeneous constraint structures. In this regard, we identify that PO holds several theoretical advantages over RL:

\paragraph{Variance reduction and scale invariance:} The REINFORCE with shared baselines algorithm, commonly used in neural-based CO works, relies on the gradient estimator $\nabla_\theta J = \mathbb{E}[(R(\tau, \mathcal{G}) - b) \nabla_\theta \log p_{\theta} (\tau, \mathcal{G})]$, where variance scales with the absolute variance of tour lengths across instances. In MTL, this is especially problematic, since different tasks typically have much different reward scales. For instance, the average costs for variants with time windows are generally much larger than variants without, causing high-cost instances to dominate gradients and implicitly prioritize certain tasks over others. In contrast, PO optimizes the Bradley-Terry objective $\mathcal{L}_{\text{PO}}$, which depends only on \emph{relative} tour rankings within a single instance. Since preferences $\mathbb{1}[R(\tau_j, \mathcal{G}) > R(\tau_k, \mathcal{G})]$ are strictly invariant to reward scaling, PO ensures a much more stable learning.

We verified this empirically by measuring gradients across multiple training batches. PO reduced per-parameter gradient variance by over $2000$ times (from $2.33 \times 10^{-6}$ to $1.09 \times 10^{-9}$) and overall gradient magnitude by $72$ times (from $~ 1.00$ to $1.39 \times 10^{-2}$), while maintaining comparable signal-to-noise ratios ($\sim 1478 \ \text{vs} \ 971$). PO's lower gradient magnitude may enable a more tolerant learning process with respect to gradient interference, as smaller updates mitigate conflicting tasks from violently overwriting each other's parameters.

\paragraph{$O(N^2)$ supervision density:} For each instance $\mathcal{G}_i$, REINFORCE extracts $N$ scalar signals $\{R(\tau_{i}^{j}, \mathcal{G}_i) - b_i\}_{j=1}^{N}$, discarding information when multiple trajectories are near-optimal and the policy converges (i.e., advantages $\approx$ 0). Conversely, PO computes $\binom{N}{2}$ pairwise comparisons from each instance, providing a much denser credit assignment than REINFORCE, allowing it to extract more information from the same trajectories, leading to a faster convergence and lower performance gaps.

\section{Experiments}
\label{experimental_results}

In this section, we validate the effectiveness of the proposed method through a series of experiments. We evaluate FiLMMeD on 24 MDVRP variants and 16 single-depot VRPs. We also conduct various ablation studies to measure the impact of all proposed contributions.

\textbf{Baselines:} \textit{MDVRP variants:} For traditional solvers, we use PyVRP \citep{Wouda2024}, a highly versatile hybrid genetic search (HGS) algorithm capable of solving all 24 MDVRP variants. We solve instances with 50 and 100 nodes, running PyVRP for 20 and 40 seconds per instance, respectively. Additionally, we report results using ten times the computational budget (200 and 400 seconds per instance). Following prior work, we parallelized PyVRP across 32 CPU cores \citep{Zhou2024}. All PyVRP default hyper-parameters were kept. For neural-based methods, we compare our method against MTPOMO \citep{Liu2024} and MVMoE \citep{Zhou2024}. We consider two versions of FiLMMeD, which incorporate our proposed contributions on top of MTPOMO and MVMoE, labeled as \textit{FiLMMeD-MTPOMO} and \textit{FiLMMeD-MVMoE} in the results. \textit{VRP variants:} For single-depot VRPs, we evaluate FiLMMeD against two different experimental configurations: 1) those defined by \citet{Zhou2024} (FiLMMeD-MTPOMO and FiLMMeD-MVMoE); 2) those defined by \citet{Berto2025} (labeled as \textit{FiLMMeD-CaDA}, incorporating our contributions on top of CaDA).

\textbf{Training:} Each neural-based model was trained for 300 epochs, with each epoch consisting of 100,000 MDVRP instances generated on the fly. Separate models were trained for 50-node and 100-node instances, with a batch size of 128 and 64 for each instance size. Each batch contains instances from a randomly selected MDVRP instance. MTPOMO and MVMoE were trained with the same hyper-parameters as in their original works, while the remaining hyper-parameters for FiLMMeD-MTPOMO and FiLMMeD-MVMoE are listed in Table~\ref{hyperparameters}. For the fine-tuning phase (applicable only to single-depot VRPs following the \citet{Zhou2024} experimentel setting), the FiLMMeD models pre-trained on MDVRP variants were further trained for 300 epochs using the PO algorithm described in Section~\ref{preference_optimization} (with a random single-depot VRP variant sampled in each batch). During this phase, we observed no significant difference in the computational time per epoch compared to the traditional RL-based training. In contrast, for the \citet{Berto2025} experimental setting, we trained FiLMMeD-CaDA from scratch for 300 epochs, using the same hyper-parameters and configurations as \citet{Li2025}. Unlike FiLMMeD-MTPOMO and FiLMMeD-MVMoE, FiLMMeD-CaDA was not fine-tuned from a pre-trained MDVRP model, since the models reported by \citet{Zhou2024} were trained on 100M instances, while the ones evaluated under the settings of \citet{Berto2025} were trained on only 30M instances. This ensures a fairer comparison. Had we fine-tuned FiLMMeD-CaDA from a pre-trained MDVRP model (which itself was trained over 30M instances), the total training exposure (60M instances) would have been double that of the baselines.

\textbf{Inference:} Across all neural-based models, we employed a greedy decoding combined with the multiple starting nodes strategy and the $\times$8 instance augmentation technique proposed by \citep{Kwon2020}. For each MDVRP variant, we generated two testing datasets, each containing 1,000 instances with 50 and 100 nodes and 3 depots (see Section~\ref{instance_gen} for the instance generation procedure). For single-depot VRPs in the \citet{Zhou2024} setting (Table~\ref{vrp_finetuningresults}), we used the same testing datasets as in \citep{Zhou2024}. For the \citet{Berto2025} setting (Table~\ref{vrp_berto}), we followed their respective experimental protocols. On single-depot problems, we used 50 and 100 starting nodes for the 50- and 100-node instances, respectively. On MDVRP variants, we increased this to 150 and 300 starting nodes to enumerate all possible initial depot-customer assignments (given that each instance contains 3 depots). We report the average objective values across each dataset (\textit{Obj.}), the average gap (\textit{Gap}) relative to the best non-neural baseline, and the computational time (\textit{Time}) required to solve all 1,000 instances. The best performing non-neural baseline is indicated with *.

\textbf{Hardware:} Non-neural baselines were run on a machine with 32 GB of RAM and an Intel Core i9-13900. Neural-based models were executed on a machine with 45GB of RAM, an Intel Xeon Gold 5315Y and an Nvidia RTX A4000. We note that we compare some results against those reported in prior works \citep{Zhou2024,Li2025}, which utilized faster GPUs. This hardware asymmetry explains some inference time discrepancies, and not architectural overhead introduced by our contributions (see Section~\ref{sec_computational_complexity} for more details).

\begin{table}[!ht]
	\centering
	\caption{Hyperparameters used by FiLMMeD-MTPOMO and FiLMMeD-MVMoE.}
	\label{hyperparameters}
	\begin{tabular}{p{3.3cm} p{4.3cm}}
		\hline
		\textbf{Hyperparameter} & \textbf{Value} \\
		\hline
		\multicolumn{2}{l}{\textbf{Model}} \\
		Embedding dimension $d_h$ & 128 \\
		Number of attention heads $A$ & 8 \\
		Number of encoder layers $L$ & 6 \\
		Feedforward hidden dimension $d_a$ & 512 \\
		Tanh clipping $\xi$ & 10.0 \\[3pt]
		\hline
		
		\multicolumn{2}{l}{\textbf{Training}} \\
		Batch size & 128 (for $n=50$) and 64 (for $n=100$) \\
		Instances per epoch & 100,000 \\
		Optimizer & Adam \\
		Learning rate (LR) & $1\times 10^{-4}$ \\
		Weight decay & $1\times 10^{-6}$ \\
		Training epochs & 300 \\[3pt]
		\hline
		
		\multicolumn{2}{l}{\textbf{Fine-tuning}} \\
		Batch size & 128 (for $n=50$) and 64 (for $n=100$) \\
		Learning rate & $3\times 10^{-4}$ \\
		Weight decay & $1\times 10^{-6}$ \\
		LR scheduler & MultiStepLR \\
		LR milestones & [270, 295] \\
		LR gamma & 0.1 \\
		Number of fine-tuning epochs & 300 \\
		Optimizer & Adam \\
		$\alpha$ & 0.03 \\[3pt]
		\hline
		
		\multicolumn{2}{l}{\textbf{Inference}} \\
		Number of POMO starting nodes (MDVRP) & 150 for $n=50$ and 300 for $n=100$ \\
		Number of POMO starting nodes (VRP) & 50 for $n=50$ and 100 for $n=100$ \\
		Instance augmentation & $\times8$, following Kwon et al. \cite{Kwon2020} \\
		\hline
	\end{tabular}
\end{table}

\subsection{Instance generation}
\label{instance_gen}

In this section, we describe the data generation process we followed for the MDVRP instances of each constraint type. Across all variants, the node coordinates for both customers and depots are randomly sampled from a uniform distribution $U(0, 1)$. We recall that for single-depot VRPs, we used the same datasets and instance generation processes as \citet{Zhou2024} and \citet{Berto2025}.

\textbf{Vehicle capacity (C):} Vehicle capacity is a constraint implicitly present in all variants considered in this work; therefore, we do not list it as a separate constraint type in Tables~\ref{mdvrp_constraint_combinations} and ~\ref{vrp_constraint_combinations}. For this constraint, we adopt the standard setup used in prior routing works \citep{Kool2019,Zhou2024}. The demand $\delta_i$ for each customer node $v_i \in \mathcal{V}_c$ is an integer sampled from the uniform distribution $U(1, 2, ..., 9)$. The vehicle capacity $C$ is set to 40, for instances with $n=50$, and to 50 for instances with $n=100$. Finally, the demand of each node is normalized by dividing it by $C$. The masking mechanism prevents the model from selecting nodes whose demand exceeds the vehicle’s remaining capacity and also masks out nodes that have already been served.

\textbf{Open route (O):} This constraint does not require any additional data to be generated. When active, it is represented simply as a Boolean flag $o_t$ in the context vector. For instances that simultaneously include both open routes and route length limit constraints, we follow the standard convention: because vehicles are not required to return to their departure depots, the final leg from the vehicle’s last visited customer back to the depot is excluded from the route-length calculation. Likewise, for instances with both open routes and time windows, the depot’s closing time $l_i$ imposes no restrictions to vehicles, since they are not required to return. We do not combine open routes with inter-depot routes in our MDVRP variants. The reason is that one would essentially contradict the other. Inter-depot routes allow vehicles to return to other depots to replenish their original capacity, whereas open routes explicitly remove the requirement to return to the depots at the end of service. For this reason, we treat these two as mutually exclusive.

\textbf{Route length limit (L):} To generate route length limits, we follow an approach similar to \citet{Berto2025}. We begin by computing the Euclidean distance between every depot-customer pair. The largest of these distances provides a natural lower bound on the maximum route length $D$, ensuring that any vehicle can reach any customer from its depot and return. Doubling this value produces a lower-bound estimate on the distance a vehicle would need to travel to visit that customer and return to its depot. We then sample $D$ from a uniform distribution between this value and 3.0. During decoding, the masking mechanism restricts vehicles to serving only those customers for which the current route length -- plus the cost of visiting the customer and returning to the depot -- remains below $D$. Again, when routes are open, the return distance to the depot is omitted from this calculation.

\textbf{Backhaul (B):} Following prior work \citep{Liu2024,Zhou2024}, we randomly designate 20\% of customer nodes as backhaul customers -- i.e., 10 nodes when $n=50$ and 20 nodes when $n=100$. During a route, the vehicle’s current capacity throughout a route must never exceed $C$ at any time. Consequently, during decoding, any backhaul customer whose $\delta_i$ would cause the vehicle to exceed its maximum capacity $C$ is masked out.

\textbf{Time window (TW):} For time windows, we follow the procedure of \citet{Berto2025}. If a vehicle arrives at a customer whose time window has not started yet, it must wait until $e_i$ to start service. Each depot also has an end time window, at which point all vehicles must have returned (when open routes are not present). During decoding, customers whose vehicles cannot complete service before the late time window $l_i$ are masked out.

\textbf{Inter-depot routes (I):} Like open routes, inter-depot routes do not require extra data generation. A Boolean $id_t$ in the context vector indicates the presence of this constraint. 

\subsection{Main MDVRP results}
\label{main_results}

\begin{sloppypar}
Our main results on the 24 MDVRP variants are shown in Table~\ref{mdvrp_mainresults}. Both FiLMMeD-MTPOMO and FiLMMeD-MVMoE consistently outperform its respective neural baselines (\textit{MTPOMO} and \textit{MVMoE}) across the majority of datasets, obtaining lower average solution costs and performance gaps. On 50-node instances, FiLMMeD-MTPOMO reached an average gap of 4.717\%, improving over its baseline by 2.888\%, while FiLMMeD-MVMoE achieved 4.740\%, outperforming MVMoE by 1.847\%. On 100-node instances, FiLMMeD-MTPOMO and FiLMMeD-MVMoE averaged 6.847\% and 6.433\%, outperforming their counterparts by 1.921\% and 5.288\%, respectively. These results indicate the effectiveness of the proposed FiLM mechanism in the encoder, as well as the CL training strategy.
\end{sloppypar}

\begin{table*}[!t]
	\vskip -0.05in
	\caption{Testing results on 1K test instances of 24 MDVRPs. The best MTL results are shown in gray, and the models improved by our approach are underlined.}
	\label{mdvrp_mainresults}
	\begin{center}
		\begin{scriptsize}
			\renewcommand\arraystretch{0.95}
			\resizebox{0.9\textwidth}{!}{ 
				\begin{tabular}{ll|cccccc|ll|cccccc}
					\midrule
					\multicolumn{2}{c|}{\multirow{2}{*}{Method}} & \multicolumn{3}{c}{\textbf{$n=50$}} & \multicolumn{3}{c|}{$n=100$} & \multicolumn{2}{c|}{\multirow{2}{*}{Method}} &
					\multicolumn{3}{c}{\textbf{$n=50$}} & \multicolumn{3}{c}{$n=100$} \\
					& & Obj. & Gap & Time & Obj. & Gap & Time & & & Obj. & Gap & Time & Obj. & Gap & Time \\
					\midrule
					\multirow{6}*{\rotatebox{90}{MDVRP}} & HGS-PyVRP & 7.995 & 0.013\% & 10.4m & 11.823 & 0.127\% & 20.8m & \multirow{6}*{\rotatebox{90}{MDOVRP}} & HGS-PyVRP & 5.338 & * & 10.4m & 7.960 & 0.026\% & 20.8m \\
					& HGS-PyVRP (10x) & 7.994 & * & 1.7h & 11.808 & * & 3.5h & & HGS-PyVRP (10x) & 5.338 & * & 1.7h & 7.958 & * & 3.5h \\
					& MTPOMO & 8.212 & 2.712\% & 13s & 12.294 & 4.117\% & 52s & & MTPOMO & 5.562 & 4.166\% & 13s & 8.404 & 5.573\% & 52s \\
					& FiLMMeD-MTPOMO & \underline{8.167} & \underline{2.163\%} & 13s & \underline{12.280} & \underline{4.002\%} & 49s & & FiLMMeD-MTPOMO & \underline{5.474} & \underline{2.522\%} & 13s & \underline{8.375} & \underline{5.207\%} & 52s \\
					& MVMoE/4E & 8.207 & 2.661\% & 18s & \cellcolor{gray!30}{12.129} & \cellcolor{gray!30}{2.726\%} & 68s & & MVMoE/4E & 5.503 & 3.071\% & 18s & 8.390 & 5.401\% & 69s \\
					& FiLMMeD-MVMoE/4E & \cellcolor{gray!30}\underline{8.152} & \cellcolor{gray!30}\underline{1.980\%} & 18s & 12.222 & 3.507\% & 69s & & FiLMMeD-MVMoE/4E & \cellcolor{gray!30}\underline{5.468} & \cellcolor{gray!30}\underline{2.411\%} & 19s & \cellcolor{gray!30}\underline{8.334} & \cellcolor{gray!30}\underline{4.696\%} & 74s \\
					\midrule
					\multirow{6}*{\rotatebox{90}{MDVRPB}} & HGS-PyVRP & 7.348 & 0.017\% & 10.4m & 10.645 & 0.169\% & 20.8m & \multirow{6}*{\rotatebox{90}{MDVRPL}} & HGS-PyVRP & 8.008 & 0.009\% & 10.4m & 11.820 & 0.140\% & 20.8m \\
					& HGS-PyVRP (10x) & 7.347 & * & 1.7h & 10.627 & * & 3.5h & & HGS-PyVRP (10x) & 8.007 & * & 1.7h & 11.803 & * & 3.5h \\
					& MTPOMO & 7.618 & 3.688\% & 12s & 11.203 & 5.436\% & 52s & & MTPOMO & 8.225 & 2.708\% & 14s & 12.288 & 4.103\% & 61s \\
					& FiLMMeD-MTPOMO & \underline{7.607} & \underline{3.545\%} & 12s & 11.248 & 5.864\% & 48s & & FiLMMeD-MTPOMO & \underline{8.184} & \underline{2.202\%} & 14s & \underline{12.280} & \underline{4.039\%} & 57s \\
					& MVMoE/4E & \cellcolor{gray!30}{7.588} & \cellcolor{gray!30}{3.280\%} & 17s & 11.225 & 5.647\% & 64s & & MVMoE/4E & 8.222 & 2.669\% & 20s & \cellcolor{gray!30}{12.126} & \cellcolor{gray!30}{2.735\%} & 76s \\
					& FiLMMeD-MVMoE/4E & 7.596 & 3.387\% & 17s & \cellcolor{gray!30}\underline{11.186} & \cellcolor{gray!30}\underline{5.268\%} & 64s & & FiLMMeD-MVMoE/4E & \cellcolor{gray!30}\underline{8.171} & \cellcolor{gray!30}\underline{2.040\%} & 19s & 12.220 & 3.535\% & 77s \\
					\midrule
					\multirow{6}*{\rotatebox{90}{MDVRPTW}} & HGS-PyVRP & 12.527 & 1.901\% & 10.4m & 21.540 & 8.746\% & 20.8m & \multirow{6}*{\rotatebox{90}{MDOVRPTW}} & HGS-PyVRP & 8.114 & 1.653\% & 10.4m & 13.477 & 2.407\% & 20.8m \\
					& HGS-PyVRP (10x) & 12.269 & * & 1.7h & 19.722 & * & 3.5h & & HGS-PyVRP (10x) & 7.981 & * & 1.7h & 13.143 & * & 3.5h \\
					& MTPOMO & 12.609 & 2.749\% & 17s & 20.610 & 4.485\% & 75s & & MTPOMO & 8.421 & 5.498\% & 17s & 14.212 & 8.117\% & 75s \\
					& FiLMMeD-MTPOMO & \cellcolor{gray!30}\underline{12.550} & \cellcolor{gray!30}\underline{2.277\%} & 17s & \underline{20.577} & \underline{4.323\%} & 71s & & FiLMMeD-MTPOMO & \cellcolor{gray!30}\underline{8.092} & \cellcolor{gray!30}\underline{1.363\%} & 19s & \underline{13.581} & \underline{3.300\%} & 77s \\
					& MVMoE/4E & 12.595 & 2.631\% & 23s & 22.020 & 11.629\% & 97s & & MVMoE/4E & 8.369 & 4.852\% & 24s & 14.401 & 9.510\% & 98s \\
					& FiLMMeD-MVMoE/4E & \underline{12.565} & \underline{2.401\%} & 24s & \cellcolor{gray!30}\underline{20.533} & \cellcolor{gray!30}\underline{4.101\%} & 94s & & FiLMMeD-MVMoE/4E & \underline{8.104} & \underline{1.515\%} & 26s & \cellcolor{gray!30}\underline{13.542} & \cellcolor{gray!30}\underline{3.005\%} & 102s \\
					\midrule
					\multirow{6}*{\rotatebox{90}{MDOVRPB}} & HGS-PyVRP & 5.194 & * & 10.4m & 7.644 & 0.020\% & 20.8m & \multirow{6}*{\rotatebox{90}{MDOVRPL}} & HGS-PyVRP & 5.334 & 0.001\% & 10.4m & 7.982 & 0.032\% & 20.8m \\
					& HGS-PyVRP (10x) & 5.194 & * & 1.7h & 7.642 & * & 3.5h & & HGS-PyVRP (10x) & 5.334 & * & 1.7h & 7.979 & * & 3.5h \\
					& MTPOMO & 5.774 & 11.149\% & 13s & 8.482 & 10.951\% & 52s & & MTPOMO & 5.561 & 4.225\% & 14s & 8.432 & 5.651\% & 59s \\
					& FiLMMeD-MTPOMO & \underline{5.456} & \underline{5.031\%} & 14s & \underline{8.233} & \underline{7.706\%} & 52s & & FiLMMeD-MTPOMO & \underline{5.469} & \underline{2.501\%} & 15s & \underline{8.399} & \underline{5.243\%} & 60s \\
					& MVMoE/4E & 5.605 & 7.888\% & 17s & 8.504 & 11.234\% & 73s & & MVMoE/4E & 5.501 & 3.112\% & 20s & 8.411 & 5.389\% & 80s \\
					& FiLMMeD-MVMoE/4E & \cellcolor{gray!30}\underline{5.450} & \cellcolor{gray!30}\underline{4.924\%} & 19s & \cellcolor{gray!30}\underline{8.183} & \cellcolor{gray!30}\underline{7.053\%} & 70s & & FiLMMeD-MVMoE/4E & \cellcolor{gray!30}\underline{5.467} & \cellcolor{gray!30}\underline{2.473\%} & 21s & \cellcolor{gray!30}\underline{8.358} & \cellcolor{gray!30}\underline{4.724\%} & 81s \\
					\midrule
					\multirow{6}*{\rotatebox{90}{MDVRPBL}} & HGS-PyVRP & 7.318 & 0.016\% & 10.4m & 10.591 & 0.174\% & 20.8m & \multirow{6}*{\rotatebox{90}{MDVRPBTW}} & HGS-PyVRP & 12.615 & 2.354\% & 10.4m & 21.160 & 7.175\% & 20.8m \\
					& HGS-PyVRP (10x) & 7.316 & * & 1.7h & 10.572 & * & 3.5h & & HGS-PyVRP (10x) & 12.312 & * & 1.7h & 19.680 & * & 3.5h \\
					& MTPOMO & 7.601 & 3.896\% & 14s & 11.146 & 5.430\% & 60s & & MTPOMO & 13.711 & 11.317\% & 17s & 22.430 & 13.925\% & 76s \\
					& FiLMMeD-MTPOMO & \underline{7.587} & \underline{3.709\%} & 14s & 11.202 & 5.956\% & 54s & & FiLMMeD-MTPOMO & \cellcolor{gray!30}\underline{13.556} & \cellcolor{gray!30}\underline{10.067\%} & 17s & \underline{22.101} & \underline{12.280\%} & 70s \\
					& MVMoE/4E & \cellcolor{gray!30}{7.574} & \cellcolor{gray!30}{3.514\%} & 18s & 11.170 & 5.665\% & 71s & & MVMoE/4E & 13.690 & 11.150\% & 23s & 24.408 & 23.983\% & 96s \\
					& FiLMMeD-MVMoE/4E & 7.579 & 3.593\% & 18s & \cellcolor{gray!30}\underline{11.126} & \cellcolor{gray!30}\underline{5.239\%} & 70s & & FiLMMeD-MVMoE/4E & \underline{13.577} & \underline{10.246\%} & 23s & \cellcolor{gray!30}\underline{22.051} & \cellcolor{gray!30}\underline{12.026\%} & 90s \\
					\midrule
					\multirow{6}*{\rotatebox{90}{MDVRPLTW}} & HGS-PyVRP & 12.492 & 2.473 & 10.4m & 21.083 & 7.192\% & 20.8m & \multirow{6}*{\rotatebox{90}{MDOVRPBL}} & HGS-PyVRP & 5.187 & * & 10.4m & 7.649 & 0.026\% & 20.8m \\
					& HGS-PyVRP (10x) & 12.167 & * & 1.7h & 19.614 & * & 3.5h & & HGS-PyVRP (10x) & 5.187 & * & 1.7h & 7.647 & * & 3.5h \\
					& MTPOMO & 12.491 & 2.640\% & 18s & 20.498 & 4.489\% & 85s & & MTPOMO & 5.767 & 11.169\% & 13s & 8.485 & 10.926\% & 58s \\
					& FiLMMeD-MTPOMO & \cellcolor{gray!30}\underline{12.443} & \cellcolor{gray!30}\underline{2.249\%} & 18s & \underline{20.479} & \underline{4.394\%} & 80s & & FiLMMeD-MTPOMO & \underline{5.445} & \underline{4.973\%} & 15s & \underline{8.236} & \underline{7.675\%} & 58s \\
					& MVMoE/4E & 12.494 & 2.664\% & 25s & 21.933 & 11.801\% & 106s & & MVMoE/4E & 5.594 & 7.840\% & 18s & 8.506 & 11.192\% & 77s \\
					& FiLMMeD-MVMoE/4E & \underline{12.461} & \underline{2.388\%} & 25s & \cellcolor{gray!30}\underline{20.441} & \cellcolor{gray!30}\underline{4.196\%} & 104s & & FiLMMeD-MVMoE/4E & \cellcolor{gray!30}\underline{5.444} & \cellcolor{gray!30}\underline{4.950\%} & 20s & \cellcolor{gray!30}\underline{8.192} & \cellcolor{gray!30}\underline{7.106\%} & 75s \\
					\midrule
					\multirow{6}*{\rotatebox{90}{MDOVRPBTW}} & HGS-PyVRP & 8.172 & 2.379\% & 10.4m & 13.274 & 1.445\% & 20.8m & \multirow{6}*{\rotatebox{90}{\scalebox{0.9}{MDOVRPLTW}}} & HGS-PyVRP & 8.017 & 0.741\% & 10.4m & 13.575 & 3.721\% & 20.8m \\
					& HGS-PyVRP (10x) & 7.975 & * & 1.7h & 13.073 & * & 3.5h & & HGS-PyVRP (10x) & 7.956 & * & 1.7h & 13.063 & * & 3.5h \\
					& MTPOMO & 8.962 & 12.348\% & 17s & 15.098 & 15.437\% & 77s & & MTPOMO & 8.402 & 5.585\% & 18s & 14.127 & 8.134\% & 83s \\
					& FiLMMeD-MTPOMO & \cellcolor{gray!30}\underline{8.627} & \cellcolor{gray!30}\underline{8.147\%} & 19s & \underline{14.420} & \underline{10.269\%} & 76s & & FiLMMeD-MTPOMO & \cellcolor{gray!30}\underline{8.063} & \cellcolor{gray!30}\underline{1.321\%} & 20s & \underline{13.499} & \underline{3.317\%} & 85s \\
					& MVMoE/4E & 8.885 & 11.392\% & 23s & 15.694 & 19.966\% & 103s & & MVMoE/4E & 8.343 & 4.845\% & 24s & 14.325 & 9.613\% & 111s \\
					& FiLMMeD-MVMoE/4E & \underline{8.635} & \underline{8.244\%} & 25s & \cellcolor{gray!30}\underline{14.384} & \cellcolor{gray!30}\underline{9.991\%} & 97s & & FiLMMeD-MVMoE/4E & \underline{8.071} & \underline{1.419\%} & 27s & \cellcolor{gray!30}\underline{13.473} & \cellcolor{gray!30}\underline{3.114\%} & 111s \\
					\midrule
					\multirow{6}*{\rotatebox{90}{MDVRPBLTW}} & HGS-PyVRP & 13.006 & 5.951\% & 10.4m & 20.904 & 6.073\% & 20.8m & \multirow{6}*{\rotatebox{90}{\scalebox{0.9}{MDOVRPBLTW}}} & HGS-PyVRP & 8.153 & 2.249\% & 10.4m & 13.506 & 3.749\% & 20.8m \\
					& HGS-PyVRP (10x) & 12.243 & * & 1.7h & 19.643 & * & 3.5h & & HGS-PyVRP (10x) & 7.968 & * & 1.7h & 13.010 & * & 3.5h \\
					& MTPOMO & 13.624 & 11.262\% & 20s & 22.389 & 13.949\% & 85s & & MTPOMO & 8.952 & 12.293\% & 18s & 15.035 & 15.543\% & 84s \\
					& FiLMMeD-MTPOMO & \cellcolor{gray!30}\underline{13.460} & \cellcolor{gray!30}\underline{9.935\%} & 19s & \underline{22.080} & \underline{12.393\%} & 78s & & FiLMMeD-MTPOMO & \cellcolor{gray!30}\underline{8.612} & \cellcolor{gray!30}\underline{8.052\%} & 20s & \underline{14.353} & \underline{10.312\%} & 83s \\
					& MVMoE/4E & 13.602 & 11.090\% & 24s & 24.342 & 23.904\% & 104s & & MVMoE/4E & 8.890 & 11.523\% & 24s & 15.585 & 19.746\% & 111s \\
					& FiLMMeD-MVMoE/4E & \underline{13.486} & \underline{10.151\%} & 24s & \cellcolor{gray!30}\underline{22.021} & \cellcolor{gray!30}\underline{12.097\%} & 99s & & FiLMMeD-MVMoE/4E & \underline{8.627} & \underline{8.230\%} & 26s & \cellcolor{gray!30}\underline{14.315} & \cellcolor{gray!30}\underline{10.020\%} & 105s \\
					\midrule
					\multirow{6}*{\rotatebox{90}{MDVRPI}} & HGS-PyVRP & 7.930 & 0.677\% & 10.4m & 11.895 & 2.159\% & 20.8m & \multirow{6}*{\rotatebox{90}{MDVRPIB}} & HGS-PyVRP & 7.234 & 0.598\% & 10.4m & 10.655 & 1.900\% & 20.8m \\
					& HGS-PyVRP (10x) & 7.876 & * & 1.7h & 11.643 & * & 3.5h & & HGS-PyVRP (10x) & 7.191 & * & 1.7h & 10.456 & * & 3.5h \\
					& MTPOMO & 8.178 & 3.866\% & 14s & 12.161 & 4.482\% & 55s & & MTPOMO & 7.634 & 6.195\% & 15s & 11.174 & 6.897\% & 56s \\
					& FiLMMeD-MTPOMO & \underline{8.173} & \underline{3.797\%} & 14s & 12.236 & 5.131\% & 52s & & FiLMMeD-MTPOMO & \underline{7.592} & \underline{5.612\%} & 13s & 11.193 & 7.085\% & 51s \\
					& MVMoE/4E & \cellcolor{gray!30}{8.141} & \cellcolor{gray!30}{3.387\%} & 19s & \cellcolor{gray!30}{12.111} & \cellcolor{gray!30}{4.048\%} & 73s & & MVMoE/4E & 7.607 & 5.816\% & 19s & 11.243 & 7.566\% & 71s \\
					& FiLMMeD-MVMoE/4E & 8.168 & 3.742\% & 19s & 12.181 & 4.652\% & 71s & & FiLMMeD-MVMoE/4E & \cellcolor{gray!30}\underline{7.579} & \cellcolor{gray!30}\underline{5.428\%} & 18s & \cellcolor{gray!30}\underline{11.127} & \cellcolor{gray!30}\underline{6.452\%} & 67s \\
					\midrule
					\multirow{6}*{\rotatebox{90}{MDVRPIL}} & HGS-PyVRP & 7.972 & 0.681\% & 10.4m & 11.938 & 2.234\% & 20.8m & \multirow{6}*{\rotatebox{90}{MDVRPITW}} & HGS-PyVRP & 13.129 & 6.410\% & 10.4m & 24.548 & 24.425\% & 20.8m \\
					& HGS-PyVRP (10x) & 7.918 & * & 1.7h & 11.676 & * & 3.5h & & HGS-PyVRP (10x) & 12.285 & * & 1.7h & 19.597 & * & 3.5h \\
					& MTPOMO & 8.214 & 3.758\% & 14s & 12.205 & 4.561\% & 62s & & MTPOMO & 13.347 & 8.740\% & 24s & 21.086 & 7.598\% & 97s \\
					& FiLMMeD-MTPOMO & \underline{8.211} & \underline{3.722\%} & 15s & 12.274 & 5.164\% & 60s & & FiLMMeD-MTPOMO & \cellcolor{gray!30}\underline{12.570} & \cellcolor{gray!30}\underline{2.317\%} & 18s & \underline{20.452} & \underline{4.347\%} & 75s \\
					& MVMoE/4E & \cellcolor{gray!30}{8.185} & \cellcolor{gray!30}{3.391\%} & 20s & \cellcolor{gray!30}{12.159} & \cellcolor{gray!30}{4.161\%} & 81s & & MVMoE/4E & 13.100 & 6.622\% & 29s & 22.197 & 13.260\% & 110s \\
					& FiLMMeD-MVMoE/4E & 8.203 & 3.627\% & 20s & 12.215 & 4.652\% & 79s & & FiLMMeD-MVMoE/4E & \underline{12.584} & \underline{2.421\%} & 24s & \cellcolor{gray!30}\underline{20.411} & \cellcolor{gray!30}\underline{4.137\%} & 97s \\
					\midrule
					\multirow{6}*{\rotatebox{90}{MDVRPIBL}} & HGS-PyVRP & 7.264 & 0.575\% & 10.4m & 10.712 & 1.934\% & 20.8m & \multirow{6}*{\rotatebox{90}{MDVRPIBTW}} & HGS-PyVRP & 13.174 & 7.318\% & 10.4m & 24.149 & 23.231\% & 20.8m \\
					& HGS-PyVRP (10x) & 7.222 & * & 1.7h & 10.507 & * & 3.5h & & HGS-PyVRP (10x) & 12.252 & * & 1.7h & 19.516 & * & 3.5h \\
					& MTPOMO & 7.643 & 5.864\% & 16s & 11.233 & 6.937\% & 62s & & MTPOMO & 14.614 & 19.416\% & 25s & 23.057 & 18.135\% & 93s \\
					& FiLMMeD-MTPOMO & \underline{7.598} & \underline{5.245\%} & 14s & 11.249 & 7.091\% & 56s & & FiLMMeD-MTPOMO & \cellcolor{gray!30}\underline{13.486} & \cellcolor{gray!30}\underline{10.091\%} & 18s & \underline{21.953} & \underline{12.470\%} & 72s \\
					& MVMoE/4E & 7.619 & 5.527\% & 21s & 11.297 & 7.549\% & 78s & & MVMoE/4E & 14.251 & 16.335\% & 28s & 24.524 & 25.622\% & 108s \\
					& FiLMMeD-MVMoE/4E & \cellcolor{gray!30}\underline{7.593} & \cellcolor{gray!30}\underline{5.166\%} & 18s & \cellcolor{gray!30}\underline{11.183} & \cellcolor{gray!30}\underline{6.463\%} & 73s & & FiLMMeD-MVMoE/4E & \underline{13.514} & \underline{10.324\%} & 23s & \cellcolor{gray!30}\underline{21.894} & \cellcolor{gray!30}\underline{12.168\%} & 93s \\
					\midrule
					\multirow{6}*{\rotatebox{90}{MDVRPILTW}} & HGS-PyVRP & 13.415 & 8.654\% & 10.4m & 23.440 & 18.695\% & 20.8m & \multirow{6}*{\rotatebox{90}{\scalebox{0.9}{MDVRPIBLTW}}} & HGS-PyVRP & 13.297 & 8.480\% & 10.4m & 23.194 & 17.481\% & 20.8m \\
					& HGS-PyVRP (10x) & 12.288 & * & 1.7h & 19.602 & * & 3.5h & & HGS-PyVRP (10x) & 12.213 & * & 1.7h & 19.603 & * & 3.5h \\
					& MTPOMO & 13.327 & 8.514\% & 26s & 21.084 & 7.555\% & 112s & & MTPOMO & 14.495 & 18.765\% & 26s & 23.133 & 17.996\% & 105s \\
					& FiLMMeD-MTPOMO & \cellcolor{gray!30}\underline{12.573} & \cellcolor{gray!30}\underline{2.297\%} & 19s & \underline{20.455} & \underline{4.337\%} & 83s & & FiLMMeD-MTPOMO & \cellcolor{gray!30}\underline{13.447} & \cellcolor{gray!30}\underline{10.068\%} & 19s & \underline{22.042} & \underline{12.421\%} & 80s \\
					& MVMoE/4E & 13.127 & 6.814\% & 31s & 22.221 & 13.347\% & 119s & & MVMoE/4E & 14.172 & 16.005\% & 30s & 24.627 & 25.620\% & 116s \\
					& FiLMMeD-MVMoE/4E & \underline{12.592} & \underline{2.451\%} & 25s & \cellcolor{gray!30}\underline{20.412} & \cellcolor{gray!30}\underline{4.115\%} & 107s & & FiLMMeD-MVMoE/4E & \underline{13.467} & \underline{10.237\%} & 24s & \cellcolor{gray!30}\underline{21.973} & \cellcolor{gray!30}\underline{12.076\%} & 102s \\
					\midrule
				\end{tabular}
			}
		\end{scriptsize}
	\end{center}
	\vskip -0.1in
\end{table*}

\subsection{Results on single-depot variants (\citet{Zhou2024} setting)}
\label{sd_results}

The main results on single-depot VRPs are shown in Table~\ref{vrp_finetuningresults}. Overall FiLMMeD achieves competitive performance relative to its baselines. While no single model consistently prevails across all variants, FiLMMeD remains highly competitive in all settings. Specifically, on 50-node instances FiLMMeD-MTPOMO achieves an average gap of 4.138\%, compared to 4.602\% for MTPOMO, while FiLMMeD-MVMoE reaches 4.081\% versus 4.352\% for MVMoE. On 100-node datasets, FiLMMeD-MTPOMO obtained an average gap of 4.099\% (versus 4.969\% from MTPOMO), and FiLMMeD-MVMoE 4.045\% (against 4.595\% from MVMoE).

Importantly, we recall that while the baselines MTPOMO and MVMoE were trained for 100M instances (5000 epochs with 20,000 instances each), FiLMMeD was fine-tuned on 30M instances (300 epochs with 100,000 instances each). Even when accounting the pre-training phase on the MDVRP, the total training represents 60\% of the baselines training (30M instances from the pre-training phase plus 30M from the fine-tuning phase).

\begin{table*}[!t]
	\vskip -0.05in
	\caption{Fine-tuning performance on 1K test instances of 16 single-depot VRPs, following the setting of \citet{Zhou2024}. The best MTL results are shown in gray, and the models improved by our approach are underlined.}
	\label{vrp_finetuningresults}
	\vskip 0.1in
	\begin{center}
		\begin{scriptsize}
			\renewcommand\arraystretch{0.95}
			\resizebox{0.95\textwidth}{!}{ 
				\begin{tabular}{ll|cccccc|ll|cccccc}
					\midrule
					\multicolumn{2}{c|}{\multirow{2}{*}{Method}} & \multicolumn{3}{c}{\textbf{$n=50$}} & \multicolumn{3}{c|}{$n=100$} & \multicolumn{2}{c|}{\multirow{2}{*}{Method}} &
					\multicolumn{3}{c}{\textbf{$n=50$}} & \multicolumn{3}{c}{$n=100$} \\
					& & Obj. & Gap & Time & Obj. & Gap & Time & & & Obj. & Gap & Time & Obj. & Gap & Time \\
					\midrule
					\multirow{8}*{\rotatebox{90}{CVRP}} & HGS & 10.334 & * & 4.6m & 15.504 & * & 9.1m & \multirow{8}*{\rotatebox{90}{VRPTW}} & HGS & 14.509 & * & 8.4m & 24.339 & * & 19.6m \\
					& LKH3 & 10.346 & 0.115\% & 9.9m & 15.590 & 0.556\% & 18.0m & & LKH3 & 14.607 & 0.664\% & 5.5m & 24.721 & 1.584\% & 7.8m \\
					& OR-Tools (x10) & 10.418 & 0.788\% & 1.7h & 15.935 & 2.751\% & 3.5h & & OR-Tools (x10) & 14.665 & 1.011\% & 1.7h & 25.212 & 3.482\% & 3.5h \\
					& POMO & 10.418 & 0.806\% & 3s & 15.734 & 1.488\% & 9s & & POMO & 14.940 & 2.990\% & 3s & 25.367 & 4.307\% & 11s \\
					& MTPOMO & 10.437 & 0.987\% & 3s & 15.790 & 1.846\% & 9s & & MTPOMO & 15.032 & 3.637\% & 3s & 25.610 & 5.313\% & 11s \\
					& FiLMMeD-MTPOMO & 10.457 & 1.185\% & 4s & 15.797 & 1.895\% & 15s & & FiLMMeD-MTPOMO & 15.064 & 3.869\% & 4s & 25.619 & 5.327\% & 20s \\
					& MVMoE/4E & \cellcolor{gray!30}{10.428} & \cellcolor{gray!30}{0.896\%} & 4s & \cellcolor{gray!30}{15.760} & \cellcolor{gray!30}{1.653\%} & 11s & & MVMoE/4E & \cellcolor{gray!30}{14.999} & \cellcolor{gray!30}{3.410\%} & 4s & \cellcolor{gray!30}{25.512} & \cellcolor{gray!30}{4.903\%} & 12s \\
					& FiLMMeD-MVMoE/4E & 10.446 & 1.082\% & 5s & 15.789 & 1.841\% & 20s & & FiLMMeD-MVMoE/4E & 15.046 & 3.736\% & 6s & 25.605 & 5.268\% & 26s \\
					\midrule
					\multirow{7}*{\rotatebox{90}{OVRP}} & LKH3 & 6.511 & 0.198\% & 4.5m & 9.828 & * & 5.3m & \multirow{7}*{\rotatebox{90}{VRPL}} & LKH3 & 10.571 & 0.790\% & 7.8m & 15.771 & * & 16.0m \\
					& OR-Tools (x10) & 6.498 & * & 1.7h & 9.842 & 0.122\% & 3.5h & & OR-Tools (x10) & 10.495 & * & 1.7h & 16.004 & 1.444\% & 3.5h \\
					& POMO & 6.609 & 1.685\% & 2s & 10.044 & 2.192\% & 8s & & POMO & 10.491 & -0.008\% & 2s & 15.785 & 0.093\% & 9s \\
					& MTPOMO & 6.671 & 2.634\% & 2s & 10.169 & 3.458\% & 8s & & MTPOMO & 10.513 & 0.201\% & 2s & 15.846 & 0.479\% & 9s \\
					& FiLMMeD-MTPOMO & \underline{6.665} & \underline{2.547\%} & 4s & \underline{10.113} & \underline{2.888\%} & 15s & & FiLMMeD-MTPOMO & 10.527 & 0.333\% & 4s & 15.849 & 0.502\% & 17s \\
					& MVMoE/4E & \cellcolor{gray!30}{6.655} & \cellcolor{gray!30}{2.402\%} & 3s & 10.138 & 3.136\% & 10s & & MVMoE/4E & \cellcolor{gray!30}{10.501} & \cellcolor{gray!30}{0.092\%} & 3s & \cellcolor{gray!30}{15.812} & \cellcolor{gray!30}{0.261\%} & 10s \\
					& FiLMMeD-MVMoE/4E & 6.662 & 2.505\% & 5s & \cellcolor{gray!30}\underline{10.111} & \cellcolor{gray!30}\underline{2.865\%} & 21s & & FiLMMeD-MVMoE/4E & 10.516 & 0.233\% & 5s & 15.845 & 0.471\% & 22s \\
					\midrule
					\multirow{6}*{\rotatebox{90}{VRPB}} & OR-Tools (x10) & 8.046 & * & 1.7h & 11.878 & * & 3.5h & \multirow{6}*{\rotatebox{90}{OVRPTW}} & OR-Tools (x10) & 8.683 & * & 1.7h & 14.380 & * & 3.5h \\
					& POMO & 8.149 & 1.276\% & 2s & 11.993 & 0.995\% & 7s & & POMO & 8.891 & 2.377\% & 3s & 14.728 & 2.467\% & 10s \\
					& MTPOMO & 8.182 & 1.684\% & 2s & 12.072 & 1.674\% & 7s & & MTPOMO & 8.987 & 3.470\% & 3s & 15.008 & 4.411\% & 10s \\
					& FiLMMeD-MTPOMO & 8.201 & 1.919\% & 3s & \underline{12.068} & \underline{1.636\%} & 13s & & FiLMMeD-MTPOMO & 8.989 & 3.499\% & 4s & \underline{14.956} & \underline{4.048\%} & 18s \\
					& MVMoE/4E & \cellcolor{gray!30}{8.170} & \cellcolor{gray!30}{1.540\%} & 3s & \cellcolor{gray!30}{12.027} & \cellcolor{gray!30}{1.285\%} & 9s & & MVMoE/4E & \cellcolor{gray!30}{8.964} & \cellcolor{gray!30}{3.210\%} & 4s & \cellcolor{gray!30}{14.927} & \cellcolor{gray!30}{3.852\%} & 11s \\
					& FiLMMeD-MVMoE/4E & 8.190 & 1.789\% & 4s & 12.059 & 1.554\% & 17s & & FiLMMeD-MVMoE/4E & 8.973 & 3.312\% & 6s & 14.939 & 3.929\% & 24s \\
					\midrule
					\multirow{5}*{\rotatebox{90}{OVRPB}} & OR-Tools (x10) & 5.745 & * & 1.7h & 8.365 & * & 3.5h & \multirow{5}*{\rotatebox{90}{OVRPL}} & OR-Tools (x10) & 6.490 & * & 1.7h & 9.790 & * & 3.5h \\
					& MTPOMO & 6.116 & 6.430\% & 2s & 8.979 & 7.335\% & 8s & & MTPOMO & 6.668 & 2.734\% & 2s & 10.126 & 3.441\% & 9s \\
					& FiLMMeD-MTPOMO & \cellcolor{gray!30}\underline{6.000} & \cellcolor{gray!30}\underline{4.397\%} & 4s & \cellcolor{gray!30}\underline{8.756} & \cellcolor{gray!30}\underline{4.667\%} & 14s & & FiLMMeD-MTPOMO & \underline{6.660} & \underline{2.615\%} & 4s & \cellcolor{gray!30}\underline{10.070} & \cellcolor{gray!30}\underline{2.868\%} & 17s \\
					& MVMoE/4E & 6.092 & 5.999\% & 3s & 8.959 & 7.088\% & 9s & & MVMoE/4E & \cellcolor{gray!30}{6.650} & \cellcolor{gray!30}{2.454\%} & 3s & 10.097 & 3.148\% & 10s \\
					& FiLMMeD-MVMoE/4E & \underline{5.999} & \underline{4.380\%} & 5s & \underline{8.759} & \underline{4.703\%} & 18s & & FiLMMeD-MVMoE/4E & 6.659 & 2.582\% & 6s & \underline{10.070} & \underline{2.869\%} & 23s \\
					\midrule
					\multirow{5}*{\rotatebox{90}{VRPBL}} & OR-Tools (x10) & 8.029 & * & 1.7h & 11.790 & * & 3.5h & \multirow{5}*{\rotatebox{90}{VRPBTW}} & OR-Tools (x10) & 14.771 & * & 1.7h & 25.496 & * & 3.5h \\
					& MTPOMO & 8.188 & 1.971\% & 2s & 11.998 & 1.793\% & 8s & & MTPOMO & 16.055 & 8.841\% & 3s & 27.319 & 7.413\% & 10s \\
					& FiLMMeD-MTPOMO & 8.194 & 2.057\% & 4s & \underline{11.986} & \underline{1.702\%} & 15s & & FiLMMeD-MTPOMO & \underline{15.886} & \underline{7.708\%} & 4s & \underline{26.912} & \underline{5.822\%} & 18s \\
					& MVMoE/4E & \cellcolor{gray!30}{8.172} & \cellcolor{gray!30}{1.776\%} & 3s & \cellcolor{gray!30}{11.945} & \cellcolor{gray!30}{1.346\%} & 9s & & MVMoE/4E & 16.022 & 8.600\% & 4s & 27.236 & 7.078\% & 11s \\
					& FiLMMeD-MVMoE/4E & 8.185 & 1.941\% & 5s & 11.975 & 1.603\% & 19s & & FiLMMeD-MVMoE/4E & \cellcolor{gray!30}\underline{15.858} & \cellcolor{gray!30}\underline{7.533\%} & 6s & \cellcolor{gray!30}\underline{26.887} & \cellcolor{gray!30}\underline{5.719\%} & 23s \\
					\midrule
					\multirow{5}*{\rotatebox{90}{VRPLTW}} & OR-Tools (x10) & 14.598 & * & 1.7h & 25.195 & * & 3.5h & \multirow{5}*{\rotatebox{90}{OVRPBL}} & OR-Tools (x10) & 5.739 & * & 1.7h & 8.348 & * & 3.5h \\
					& MTPOMO & 14.961 & 2.586\% & 3s & 25.619 & 1.920\% & 12s & & MTPOMO & 6.104 & 6.306\% & 2s & 8.961 & 7.343\% & 8s \\
					& FiLMMeD-MTPOMO & 14.999 & 2.835\% & 5s & 25.628 & 1.942\% & 22s & & FiLMMeD-MTPOMO & \underline{5.991} & \underline{4.364\%} & 4s & \underline{8.746} & \underline{4.768\%} & 15s \\
					& MVMoE/4E & \cellcolor{gray!30}{14.937} & \cellcolor{gray!30}{2.421\%} & 4s & \cellcolor{gray!30}{25.514} & \cellcolor{gray!30}{1.471\%} & 13s & & MVMoE/4E & 6.076 & 5.843\% & 3s & 8.942 & 7.115\% & 9s \\
					& FiLMMeD-MVMoE/4E & 14.978 & 2.696\% & 6s & 25.610 & 1.854\% & 28s & & FiLMMeD-MVMoE/4E & \cellcolor{gray!30}\underline{5.986} & \cellcolor{gray!30}\underline{4.260\%} & 5s & \cellcolor{gray!30}\underline{8.740} & \cellcolor{gray!30}\underline{4.705\%} & 19s \\
					\midrule
					\multirow{5}*{\rotatebox{90}{OVRPBTW}} & OR-Tools (x10) & 8.675 & * & 1.7h & 14.384 & * & 3.5h & \multirow{5}*{\rotatebox{90}{OVRPLTW}} & OR-Tools (x10) & 8.669 & * & 1.7h & 14.279 & * & 3.5h \\
					& MTPOMO & 9.514 & 9.628\% & 3s & 15.879 & 10.453\% & 10s & & MTPOMO & 8.987 & 3.633\% & 3s & 14.896 & 4.374\% & 11s \\
					& FiLMMeD-MTPOMO & \underline{9.415} & \underline{8.504\%} & 4s & \underline{15.619} & \underline{8.653\%} & 17s & & FiLMMeD-MTPOMO & 8.990 & 3.667\% & 5s & \underline{14.854} & \underline{4.080\%} & 20s \\
					& MVMoE/4E & 9.486 & 9.308\% & 4s & 15.808 & 9.948\% & 11s & & MVMoE/4E & \cellcolor{gray!30}{8.966} & \cellcolor{gray!30}{3.396\%} & 4s & \cellcolor{gray!30}{14.828} & \cellcolor{gray!30}{3.903\%} & 12s \\
					& FiLMMeD-MVMoE/4E & \cellcolor{gray!30}\underline{9.390} & \cellcolor{gray!30}\underline{8.216\%} & 6s & \cellcolor{gray!30}\underline{15.607} & \cellcolor{gray!30}\underline{8.572\%} & 22s & & FiLMMeD-MVMoE/4E & 8.971 & 3.440\% & 6s & 14.854 & 4.083\% & 26s \\
					\midrule
					\multirow{5}*{\rotatebox{90}{VRPBLTW}} & OR-Tools (x10) & 14.677 & * & 1.7h & 25.342 & * & 3.5h &  \multirow{5}*{\rotatebox{90}{\scalebox{0.9}{OVRPBLTW}}} & OR-Tools (x10) & 8.673 & * & 1.7h & 14.250 & * & 3.5h \\
					& MTPOMO & 15.980 & 9.035\% & 3s & 27.247 & 7.746\% & 11s & & MTPOMO & 9.532 & 9.851\% & 3s & 15.738 & 10.498\% & 10s \\
					& FiLMMeD-MTPOMO & \underline{15.834} & \underline{8.033\%} & 5s & \underline{26.818} & \underline{6.053\%} & 20s & & FiLMMeD-MTPOMO & \underline{9.430} & \underline{8.679\%} & 5s & \underline{15.486} & \underline{8.732\%} & 18s \\
					& MVMoE/4E & 15.945 & 8.775\% & 4s & 27.142 & 7.332\% & 12s & & MVMoE/4E & 9.503 & 9.516\% & 4s & 15.671 & 10.009\% & 11s \\
					& FiLMMeD-MVMoE/4E & \cellcolor{gray!30}\underline{15.794} & \cellcolor{gray!30}\underline{7.768\%} & 6s & \cellcolor{gray!30}\underline{26.801} & \cellcolor{gray!30}\underline{5.985\%} & 25s & & FiLMMeD-MVMoE/4E & \cellcolor{gray!30}\underline{9.415} & \cellcolor{gray!30}\underline{8.505\%} & 6s & \cellcolor{gray!30}\underline{15.481} & \cellcolor{gray!30}\underline{8.692\%} & 23s \\
					\midrule
			\end{tabular}}
	\end{scriptsize}
\end{center}
\vskip -0.1in
\end{table*}

\subsection{Ablation studies}
\label{ablation_studies}

In this section, we assess the contribution of each component of our approach through a series of ablation studies. All models in this section were trained on instances with 50 nodes, and inference was performed using a greedy decoding with $\times$8 instance augmentation.

\subsubsection{Curriculum learning}
\label{sec_cl_ablation}

To quantify the effectiveness of the CL regimen, we trained FiLMMeD-MTPOMO and FiLMMeD-MVMoE under three alternative strategies: 1) \textit{unified}, which consists in training on variants that contain only a single constraint (i.e., MDVRP, MDOVRP, MDVRPB, MDVRPL, MDVRPTW, MDVRPI), following \citet{Liu2024}; 2) \textit{full task set}, which consists in sampling uniformly across all available variants, following \citet{Berto2025}; 3) \textit{standard CL}, in which the initial training problem set contains only single-constraint variants. In Section~\ref{sec_cl}, we argue that including three additional variants in the initial CL training set (i.e., MDOVRPTW, MDVRPBTW and MDVRPITW) is beneficial to generalization. Hence, we include the \textit{standard CL} results as well, which does not feature these three variants in the initial training set.

\begin{table}[pos=!ht]
	\caption{Average objective function and gaps across 24 MDVRP variants for different training strategies.}
	\label{cl_ablation}
	\begin{center}
		\begin{small}
			\renewcommand\arraystretch{1.0}
			\begin{tabular}{l|cc}
				\toprule
				Model & Obj. & Gap \\
				\midrule
				FiLMMeD-MTPOMO - unified & 9.630 & 8.396\% \\
				FiLMMeD-MTPOMO - full task set & 9.320 & 5.295\% \\
				FiLMMeD-MTPOMO - standard CL & 9.306 & 5.134\% \\
				FiLMMeD-MTPOMO & \textbf{9.268} & \textbf{4.717\%} \\
				\midrule
				FiLMMeD-MVMoE - unified & 9.730 & 9.318\% \\
				FiLMMeD-MVMoE - full task set & 9.298 & 5.046\% \\
				FiLMMeD-MVMoE - standard CL & 9.299 & 5.020\% \\
				FiLMMeD-MVMoE & \textbf{9.273} & \textbf{4.740\%} \\
				\bottomrule
			\end{tabular}
		\end{small}
	\end{center}
\end{table}

All three baselines also include the FiLM mechanism. Table~\ref{cl_ablation} reports the average objective and gap for each model under the different training strategies. We observed that our CL training yields significant generalization improvements over the other three strategies on both FiLMMeD-MTPOMO and FiLMMeD-MVMoE. This confirms that, for MDVRP variants, a structured curriculum is essential for robust generalization. Simpler strategies like \textit{unified} sampling -- which relies on zero-shot composition -- struggled on the multi-depot setting. On variants with two, three and four constraints simultaneously, it struggled to learn how these different constraints interact with one another. In contrast to the single-depot VRP, learning constraints in isolation proved insufficient for the MDVRP. One possible reason for this stems from the fact that the multi-depot structure itself acts as an additional constraint, complicating constraint interactions significantly more than in single-depot VRPs. The standard CL strategy improves upon this problem by gradually increasing the complexity of variants sampled throughout training. However, the delayed exposure to complex variants led to stagnated learning. In contrast, our CL regimen, which exposes the model to more complex constraint interactions early on, led to a smoother convergence throughout training. 

\begin{figure}[pos=!ht]
	\centering
	\includegraphics[width=0.9\columnwidth]{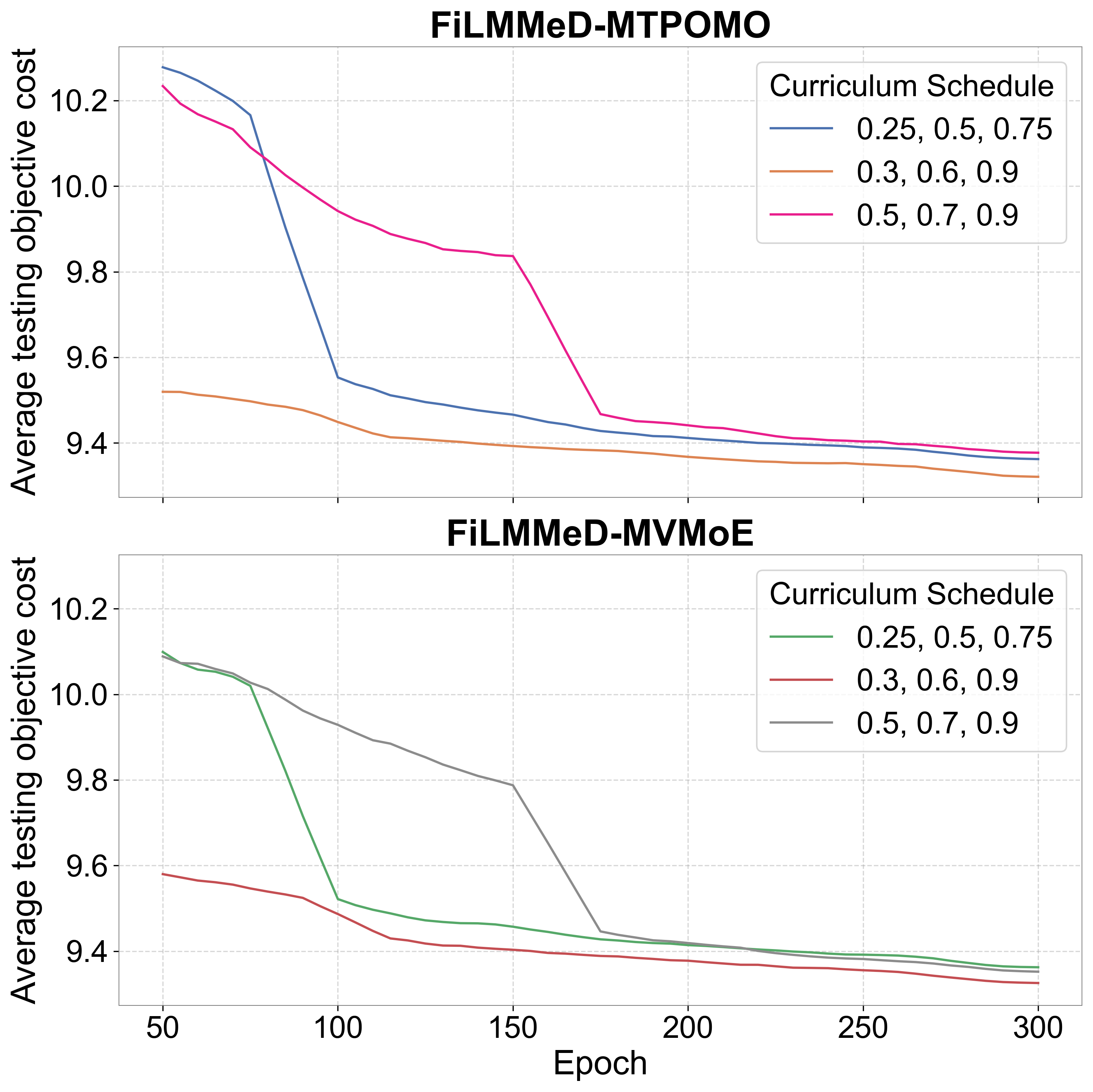}
	\caption{Comparison of different CL schedules: Evolution of the average testing objective cost across 24 MDVRP variants.}
	\label{cl_schedule}
\end{figure}

We also conducted sensitivity analyses on different CL schedules, i.e., when to add more complex variants into the training problem set. We examined two alternative curriculum schedules: 1) a faster progression (transitions at 25\%, 50\%, and 75\% of epochs); and 2) a slower progression (transitions at 50\%, 70\%, and 90\% of epochs). Figure~\ref{cl_schedule} compares the learning curves of these alternatives against our selected schedule (which has transitions at 30\%, 60\%, and 90\% of epochs). Here, we present the results from epoch 50 onward to improve visualization. Overall, our chosen schedule led to a much better generalization compared to the other two alternatives.

\subsubsection{Preference optimization}
\label{ablation_po}

\begin{figure}[pos=!ht]
	\centering
	\includegraphics[width=0.9\columnwidth]{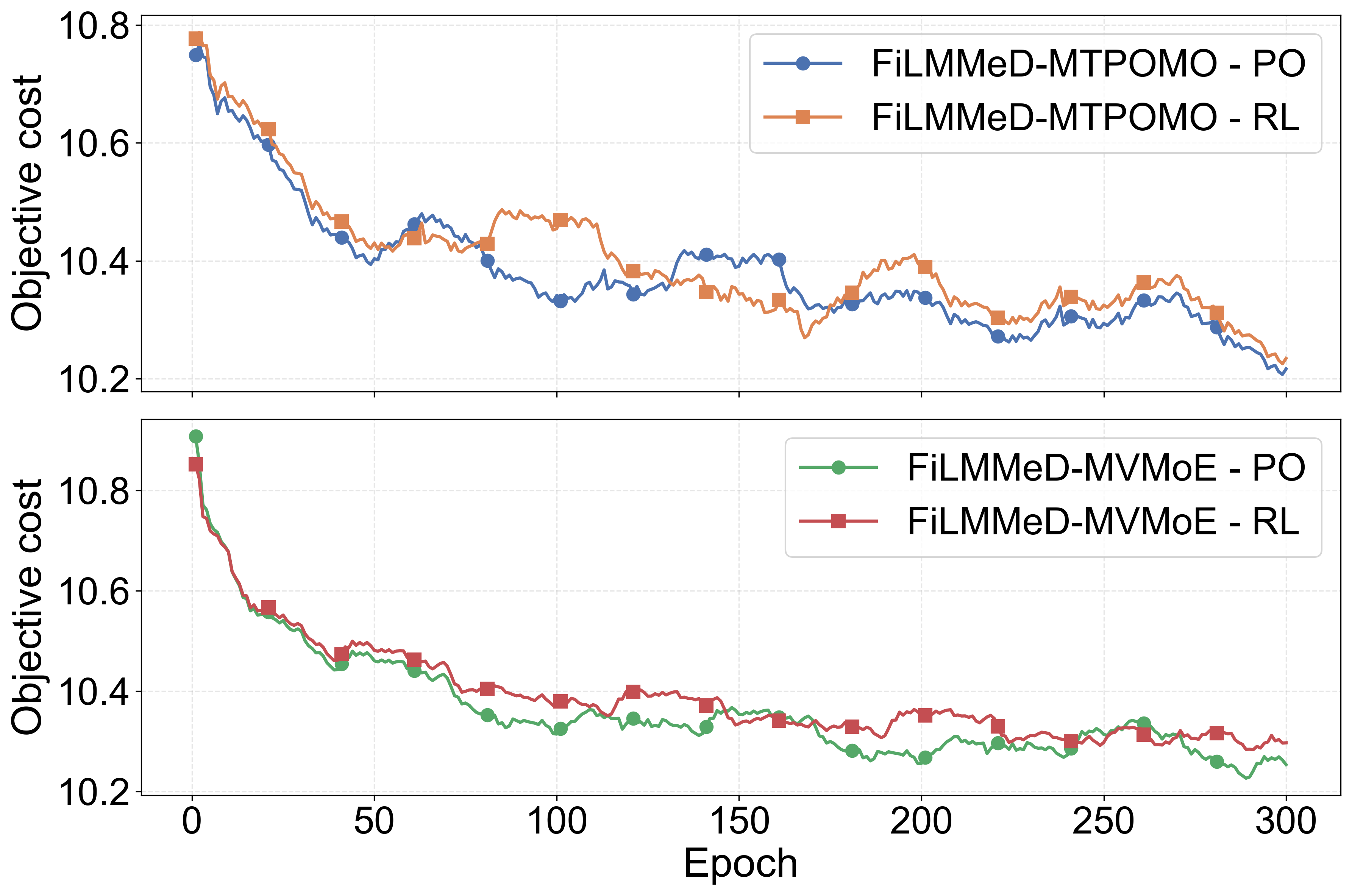}
	\caption{Convergence of FiLMMeD models during fine-tuning with PO and RL (smoothed over a moving average of 20 epochs for better visualization).}
	\label{fig_po_convergence}
\end{figure}

\begin{figure}[pos=!ht]
	\centering
	\includegraphics[width=0.9\columnwidth]{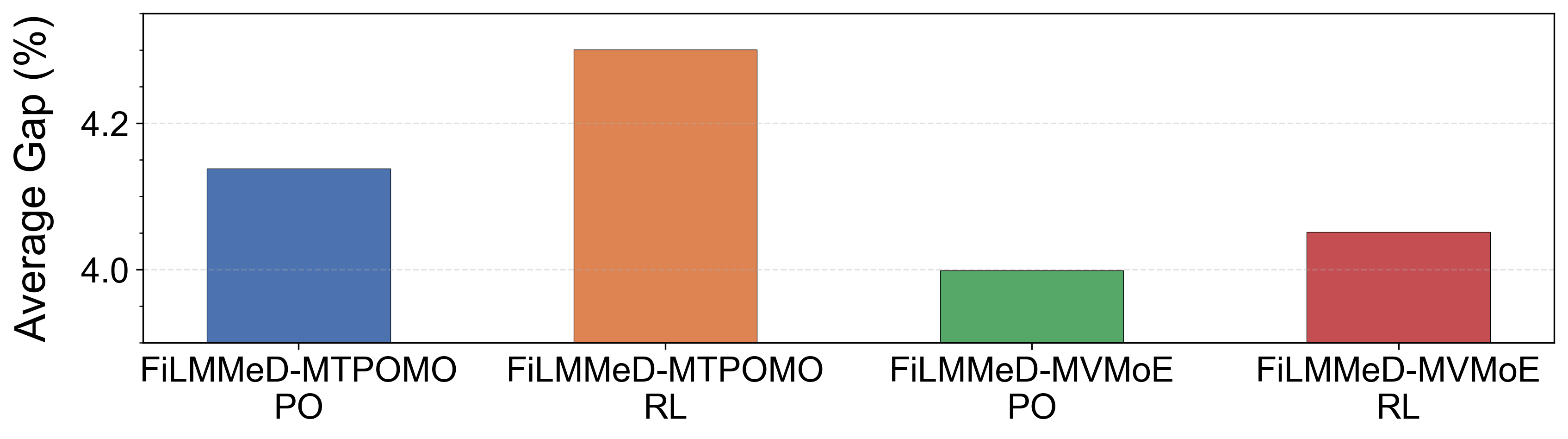}
	\caption{Average gap on 16 single-depot VRP variants of fine-tuned FiLMMeD models with PO and RL.}
	\label{fig_po_vs_rl_gap}
\end{figure}

To assess the impact of fine-tuning FiLMMeD on single-depot variants via PO, we also fine-tuned FiLMMeD for 300 additional epochs with the standard REINFORCE with shared baselines loss, keeping all configurations and hyperparameters identical. Figure~\ref{fig_po_convergence} reports the convergence of both approaches over these epochs. Both FiLMMeD-MTPOMO and FiLMMeD-MVMoE display a more favorable convergence profile when fine-tuned with PO rather than RL. We further compare the two strategies across the 16 single-depot VRP variants, reporting their average performance gaps during inference in Figure~\ref{fig_po_vs_rl_gap}. The results validate the convergence improvements observed during fine-tuning, with both models outperforming their respective RL variants.

As stated previously, we decided to pre-train MDVRP models with RL to better isolate the effects of the CL regimen and FiLM. However, for the sake of comparison, we also assessed the performance of PO when pre-training the MDVRP models.  Table~\ref{tab_po_mdvrp} shows that PO works just as well in the multi-depot setting, yielding consistent improvements over RL across both FiLMMeD-MTPOMO and FiLMMeD-MVMoE architectures. These results further support its adoption as a superior alternative to RL in future MTL models.

\begin{table}[pos=h]
	\centering
	\caption{Average gaps across all 24 MDVRP variants: RL vs. PO comparison}
	\label{tab_po_mdvrp}
	\begin{tabular}{lcc}
		\toprule
		Model & RL & PO \\
		\midrule
		FiLMMeD-MTPOMO & 4.717\% & \textbf{4.439\%} \\
		FiLMMeD-MVMoE & 4.740\% & \textbf{4.509\%} \\
		\bottomrule
	\end{tabular}
\end{table}

\subsubsection{FiLM}

Lastly, we evaluate the contribution of the FiLM mechanism. For this analysis, we fine-tuned MTPOMO and MVMoE, starting from the pre-trained models released by \citet{Zhou2024}\footnote{\url{https://github.com/RoyalSkye/Routing-MVMoE/tree/main/pretrained}}. Both models were originally trained on single-depot variants, and our goal was to extend their capabilities to also handle MDVRP variants. To extend their capabilities to multi-depot settings while testing compositional generalization, we fine-tuned both architectures with and without FiLM on a subset of 12 variants: 6 single-depot (CVRP, OVRP, VRPTW, VRPL, VRPB, OVRPTW) and 6 multi-depot (MDVRP, MDVRPB, MDVRPL, MDOVRP, MDVRPTW, MDVRPI), holding out the remaining 28 variants for zero-shot evaluation. 

\begin{figure}[pos=!h]
	\centering
	\includegraphics[width=0.9\columnwidth]{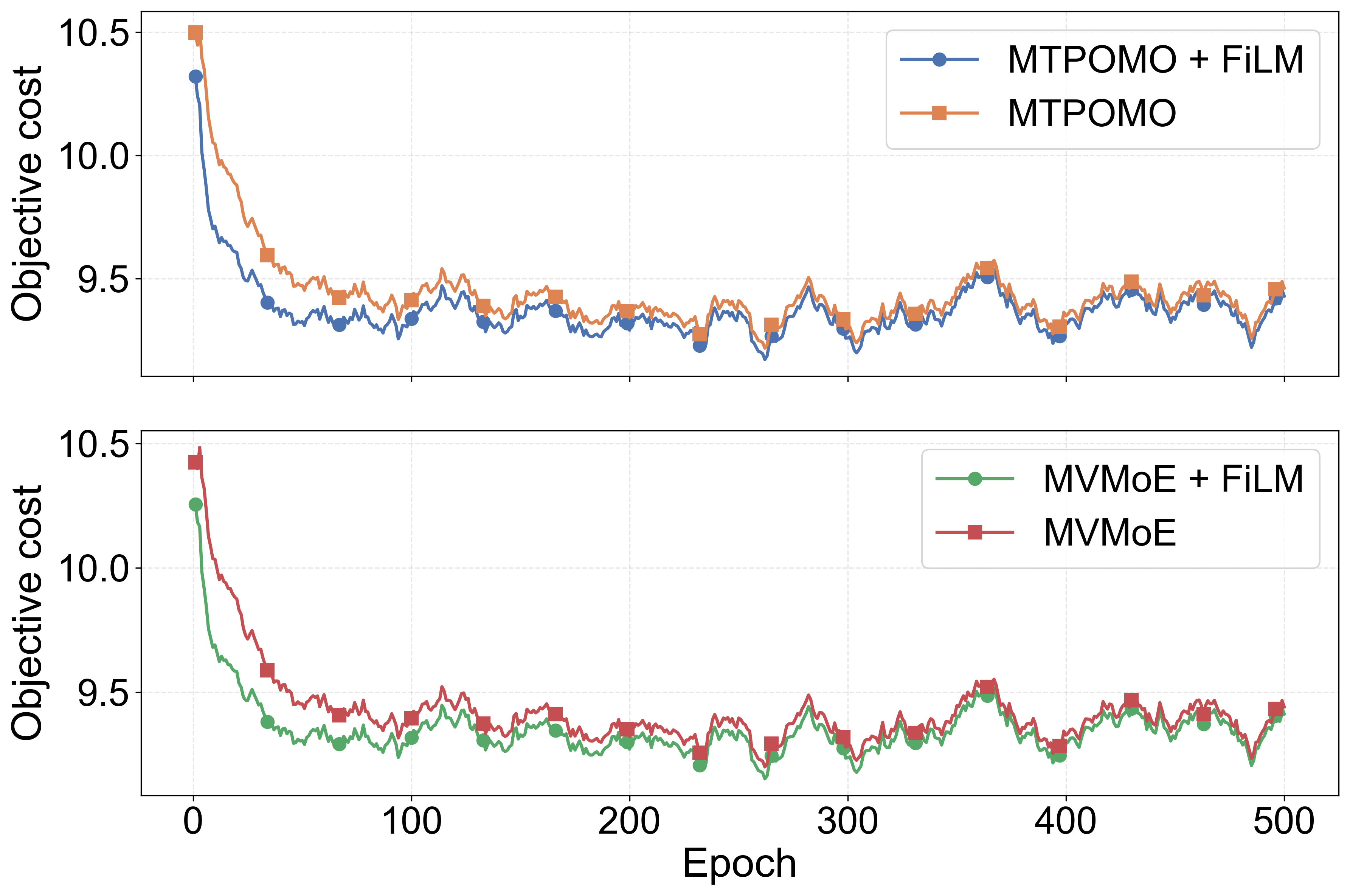}
	\caption{Convergence of MTPOMO and MVMoE fine-tuned with and without FiLM on both single- and multi-depot variants.}
	\label{fig_film_convergence}
\end{figure}

We incorporated FiLM by extending the conditioning vector $z$ to 6 dimensions, adding a Boolean feature indicating whether an instance contains one or multiple depots. To preserve the prior knowledge from the pre-trained models, we initialized the FiLM parameters as follows: $\gamma$ weights set to zero with bias set to one, and $\beta$ weights and bias set to zero. This ensures that, at initialization, the FiLM layers behave as identity transformations and do not alter the existing representations. We then fine-tuned both the FiLM-augmented and original (non-FiLM) models for 500 epochs (approximately 10\% of the original total training data), using 20,000 training instances per epoch. To isolate the architectural contribution of FiLM and ensure fair comparison with the original pre-trained models, both the FiLM-augmented and non-FiLM models were trained using the REINFORCE with shared baselines algorithm. The convergence across the 500 fine-tuning epochs is shown in Figure~\ref{fig_film_convergence}. Overall, the FiLM component resulted in a clearly smoother convergence on both models.

This translated into substantially better inference performance, particularly on zero-shot generalization to unseen constraint combinations. As reported in Table~\ref{tab_unseen_gen}, FiLM-augmented models consistently outperformed their non-FiLM counterparts across both seen and unseen variants. FiLM-augmented MTPOMO achieved an average gap of 5.437\% compared to 5.972\% for the baseline MTPOMO, with the most pronounced gains observed on the 28 held-out unseen variants (6.764\% vs. 7.386\%). Similarly, FiLMMeD-augmented MVMoE attained an average gap of 5.297\% versus 5.716\% for the standard MVMoE, with unseen variant performance improving from 7.098\% to 6.620\%. These results confirm our prior intuition that FiLM would enhance the performance on unseen constraint combinations.

\begin{table}[pos=!ht]
	\centering
	\caption{Generalization performance on seen vs. unseen variants. Models trained on 12 variants and evaluated zero-shot on 28 held-out variants.}
	\label{tab_unseen_gen}
	\begin{tabular}{lccc}
		\toprule
		Methods & Seen (12) & Unseen (28) & Average \\
		\midrule
		MTPOMO & 2.672\% & 7.386\% & 5.972\% \\
		MTPOMO + FiLM & \textbf{2.341\%} & \textbf{6.764\%} & \textbf{5.437\%} \\
		\midrule
		MVMoE & 2.493\% & 7.098\% & 5.716\% \\
		MVMoE + FiLM & \textbf{2.208\%} & \textbf{6.620\%} & \textbf{5.297\%} \\
		\bottomrule
	\end{tabular}
\end{table}

\begin{figure}[pos=!h]
	\centering
	\includegraphics[width=0.95\columnwidth]{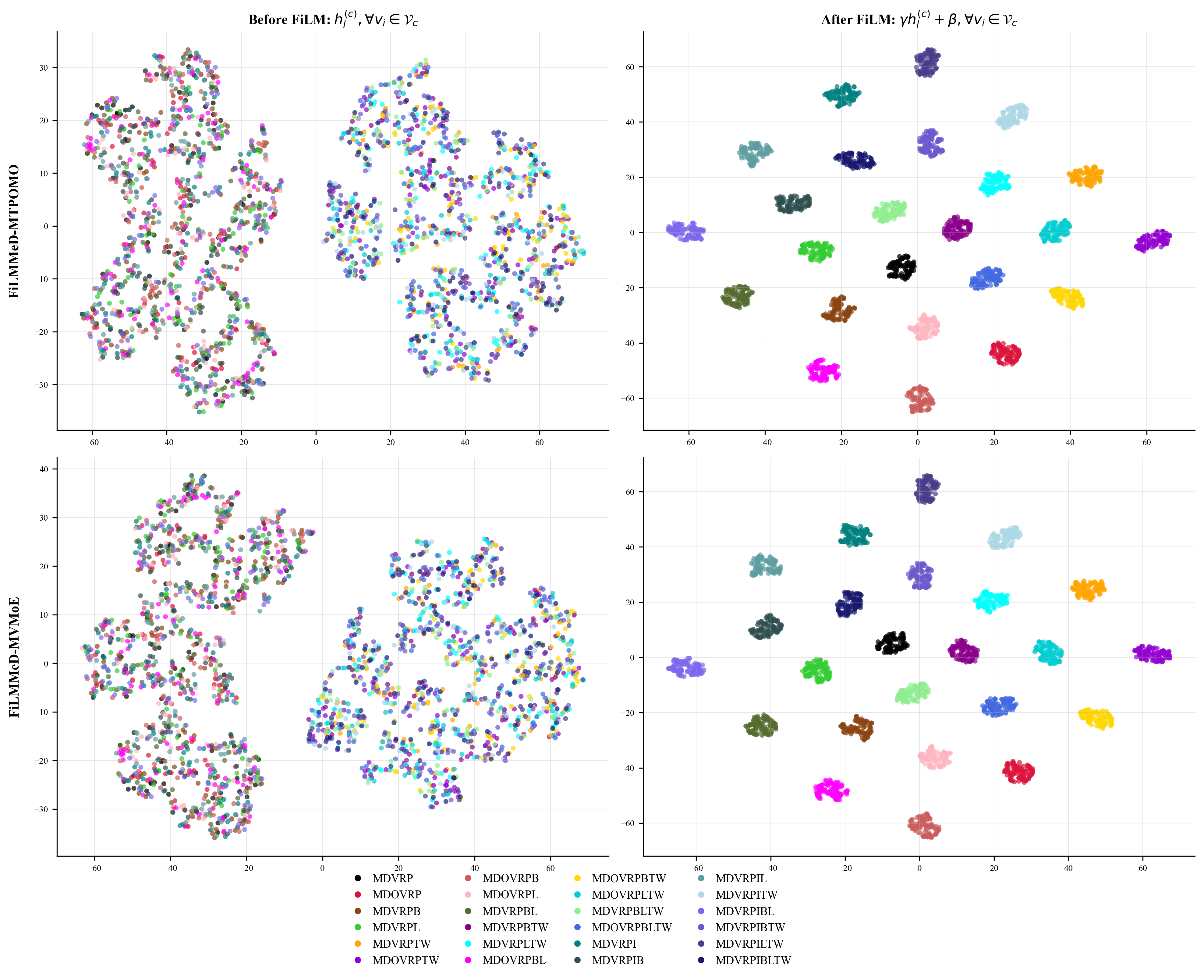}
	\caption{T-SNE visualization comparison for the learned customer embeddings before and after the FiLM mechanism.}
	\label{tsne_fig_film}
\end{figure}

\begin{figure}[pos=!h]
	\centering
	\includegraphics[width=0.95\columnwidth]{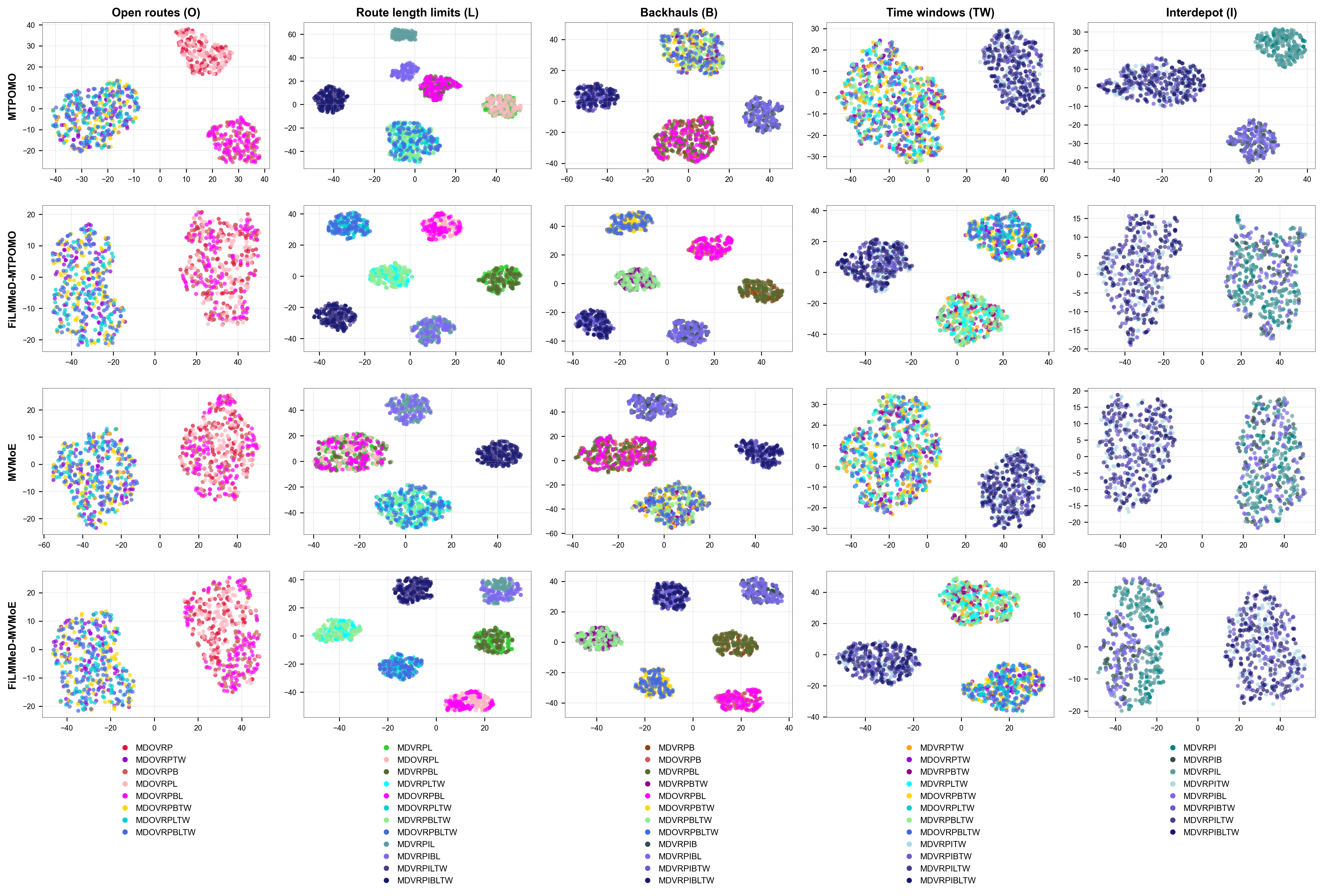}
	\caption{T-SNE visualization comparison for the last encoder layer of different models.}
	\label{tsne_fig}
\end{figure}

To provide more empirical evidence regarding the effectiveness of the FiLM mechanism, we analyzed the latent representations of all 24 MDVRP variants using the t-SNE technique \citep{vanDerMaaten2008}. First, we compared the learned customer embeddings directly before and after the FiLM transformation. The pre-modulation embeddings (on the left of  Figure~\ref{tsne_fig_film}) reveal a latent space structured primarily by input data topology. We can see that two large cluster emerge, distinguishing variants with and without time windows. In this case, variants that differ by other constraints (such as open routes and backhauls) remain indistinguishable and entangled within these clusters. This indicates that the linear layer (the same from equation ~\ref{eq_linear_customer}) learns a generally task-agnostic unified representation, capturing fundamental spatial and demand features shared across all variants.

\begin{table*}[pos=!h]
	\vskip -0.05in
	\caption{Performance on 1K test instances of 16 single-depot VRPs, following the setting of \citet{Berto2025}. The best MTL results (among neural models) are shown in gray, and the models improved by our approach (FiLMMeD vs. respective CaDA) are underlined.}
	\label{vrp_berto}
	\vskip 0.1in
	\begin{center}
		\begin{scriptsize}
			\renewcommand\arraystretch{0.95}
			\resizebox{0.95\textwidth}{!}{
				\begin{tabular}{ll|cccccc|ll|cccccc}
					\toprule
					\multicolumn{2}{c|}{\multirow{2}{*}{Method}} & \multicolumn{3}{c}{\textbf{$n=50$}} & \multicolumn{3}{c|}{$n=100$} & \multicolumn{2}{c|}{\multirow{2}{*}{Method}} &
					\multicolumn{3}{c}{\textbf{$n=50$}} & \multicolumn{3}{c}{$n=100$} \\
					\cmidrule(lr){3-5} \cmidrule(lr){6-8} \cmidrule(lr){11-13} \cmidrule(lr){14-16}
					& & Obj. & Gap & Time & Obj. & Gap & Time & & & Obj. & Gap & Time & Obj. & Gap & Time \\
					\midrule
					\multirow{9}*{\rotatebox{90}{CVRP}}
					& HGS-PyVRP & 10.372 & * & 10.4m & 15.628 & * & 20.8m & \multirow{9}*{\rotatebox{90}{VRPTW}} & HGS-PyVRP & 16.031 & * & 10.4m & 25.423 & * & 20.8m \\
					& OR-Tools & 10.572 & 1.907\% & 10.4m & 16.280 & 4.178\% & 20.8m & & OR-Tools & 16.089 & 0.347\% & 10.4m & 25.814 & 1.506\% & 20.8m \\
					& MTPOMO & 10.520 & 1.423\% & 2s & 15.941 & 2.030\% & 8s & & MTPOMO & 16.419 & 2.423\% & 2s & 26.433 & 3.962\% & 9s \\
					& MVMoE & 10.499 & 1.229\% & 3s & 15.888 & 1.693\% & 11s & & MVMoE & 16.400 & 2.298\% & 3s & 26.390 & 3.789\% & 11s \\
					& RouteFinder & 10.502 & 1.257\% & 2s & 15.860 & 1.524\% & 8s & & RouteFinder & 16.341 & 1.933\% & 2s & 26.228 & 3.154\% & 8s \\
					& CaDA & 10.494 & 1.182\% & 2s & 15.870 & 1.578\% & 8s & & CaDA & 16.278 & 1.536\% & 2s & 26.070 & 2.530\% & 8s \\
					& FiLMMeD-CaDA & \underline{10.468} & \underline{0.924\%} & 4s & \underline{15.793} & \underline{1.089\%} & 19s & & FiLMMeD-CaDA & \underline{16.266} & \underline{1.461\%} & 4s & \cellcolor{gray!30}\underline{26.029} & \cellcolor{gray!30}\underline{2.365\%} & 20s \\
					& CaDA† & 10.476 & 1.012\% & 5s & 15.800 & 1.139\% & 28s & & CaDA† & 16.291 & 1.616\% & 6s & 26.106 & 2.673\% & 31s \\
					& FiLMMeD-CaDA† & \cellcolor{gray!30}\underline{10.461} & \cellcolor{gray!30}\underline{0.865\%} & 5s & \cellcolor{gray!30}\underline{15.762} & \cellcolor{gray!30}\underline{0.894\%} & 28s & & FiLMMeD-CaDA† & \cellcolor{gray!30}\underline{16.251} & \cellcolor{gray!30}\underline{1.365\%} & 6s & \underline{26.062} & \underline{2.494\%} & 31s \\
					\midrule
					\multirow{9}*{\rotatebox{90}{OVRP}}
					& HGS-PyVRP & 6.507 & * & 10.4m & 9.725 & * & 20.8m & \multirow{9}*{\rotatebox{90}{VRPL}} & HGS-PyVRP & 10.587 & * & 10.4m & 15.766 & * & 20.8m \\
					& OR-Tools & 6.553 & 0.686\% & 10.4m & 9.995 & 2.732\% & 20.8m & & OR-Tools & 10.570 & 2.343\% & 10.4m & 16.466 & 5.302\% & 20.8m \\
					& MTPOMO & 6.717 & 3.194\% & 2s & 10.216 & 5.028\% & 8s & & MTPOMO & 10.775 & 1.733\% & 2s & 16.157 & 2.483\% & 8s \\
					& MVMoE & 6.705 & 3.003\% & 3s & 10.177 & 4.617\% & 11s & & MVMoE & 10.753 & 1.525\% & 3s & 16.099 & 2.113\% & 11s \\
					& RouteFinder & 6.682 & 2.658\% & 2s & 10.115 & 3.996\% & 8s & & RouteFinder & 10.747 & 1.485\% & 2s & 16.057 & 1.858\% & 8s \\
					& CaDA & 6.670 & 2.468\% & 2s & 10.121 & 4.045\% & 8s & & CaDA & 10.731 & 1.333\% & 2s & 16.057 & 1.847\% & 8s \\
					& FiLMMeD-CaDA & \underline{6.649} & \underline{2.148\%} & 4s & \underline{10.078} & \underline{3.605\%} & 19s & & FiLMMeD-CaDA & \underline{10.706} & \underline{1.101\%} & 4s & \underline{15.988} & \underline{1.419\%} & 19s \\
					& CaDA† & 6.658 & 2.293\% & 5s & 10.078 & 3.616\% & 29s & & CaDA† & 10.711 & 1.152\% & 5s & 15.987 & 1.424\% & 29s \\
					& FiLMMeD-CaDA† & \cellcolor{gray!30}\underline{6.647} & \cellcolor{gray!30}\underline{2.128\%} & 5s & \cellcolor{gray!30}\underline{10.030} & \cellcolor{gray!30}\underline{3.115\%} & 29s & & FiLMMeD-CaDA† & \cellcolor{gray!30}\underline{10.699} & \cellcolor{gray!30}\underline{1.044\%} & 5s & \cellcolor{gray!30}\underline{15.950} & \cellcolor{gray!30}\underline{1.181\%} & 29s \\
					\midrule
					\multirow{9}*{\rotatebox{90}{VRPB}}
					& HGS-PyVRP & 9.687 & * & 10.4m & 14.377 & * & 20.8m & \multirow{9}*{\rotatebox{90}{OVRPTW}} & HGS-PyVRP & 10.510 & * & 10.4m & 16.926 & * & 20.8m \\
					& OR-Tools & 9.802 & 1.159\% & 10.4m & 14.933 & 3.853\% & 20.8m & & OR-Tools & 10.519 & 0.078\% & 10.4m & 17.027 & 0.583\% & 20.8m \\
					& MTPOMO & 10.036 & 3.596\% & 2s & 15.102 & 5.052\% & 8s & & MTPOMO & 10.676 & 1.558\% & 2s & 17.442 & 3.022\% & 9s \\
					& MVMoE & 10.007 & 3.292\% & 3s & 15.023 & 4.505\% & 10s & & MVMoE & 10.674 & 1.541\% & 3s & 17.416 & 2.870\% & 12s \\
					& RouteFinder & 9.979 & 3.000\% & 2s & 14.935 & 3.906\% & 8s & & RouteFinder & 10.645 & 1.264\% & 2s & 17.328 & 2.352\% & 9s \\
					& CaDA & 9.960 & 2.800\% & 2s & 14.960 & 4.038\% & 8s & & CaDA & 10.613 & 0.957\% & 2s & 17.226 & 1.751\% & 9s \\
					& FiLMMeD-CaDA & \underline{9.925} & \underline{2.442\%} & 4s & \underline{14.859} & \underline{3.369\%} & 19s & & FiLMMeD-CaDA & \underline{10.611} & \underline{0.940\%} & 4s & \cellcolor{gray!30}\underline{17.214} & \cellcolor{gray!30}\underline{1.678\%} & 19s \\
					& CaDA† & 9.931 & 2.502\% & 5s & 14.852 & 3.319\% & 28s & & CaDA† & 10.623 & 1.058\% & 6s & 17.247 & 1.876\% & 32s \\
					& FiLMMeD-CaDA† & \cellcolor{gray!30}\underline{9.909} & \cellcolor{gray!30}\underline{2.278\%} & 5s & \cellcolor{gray!30}\underline{14.801} & \cellcolor{gray!30}\underline{2.964\%} & 28s & & FiLMMeD-CaDA† & \cellcolor{gray!30}\underline{10.597} & \cellcolor{gray!30}\underline{0.810\%} & 6s & \underline{17.232} & \underline{1.781\%} & 33s \\
					\midrule
					\multirow{9}*{\rotatebox{90}{VRPBL}}
					& HGS-PyVRP & 10.186 & * & 10.4m & 14.779 & * & 20.8m & \multirow{9}*{\rotatebox{90}{VRPBLTW}} & HGS-PyVRP & 18.361 & * & 10.4m & 29.026 & * & 20.8m \\
					& OR-Tools & 10.331 & 1.390\% & 10.4m & 15.426 & 4.338\% & 20.8m & & OR-Tools & 18.422 & 0.332\% & 10.4m & 29.830 & 2.770\% & 20.8m \\
					& MTPOMO & 10.679 & 4.760\% & 2s & 15.718 & 6.294\% & 8s & & MTPOMO & 19.001 & 2.199\% & 3s & 30.948 & 3.794\% & 9s \\
					& MVMoE & 10.639 & 4.384\% & 3s & 15.642 & 5.771\% & 11s & & MVMoE & 18.983 & 2.097\% & 3s & 30.892 & 3.609\% & 12s \\
					& RouteFinder & 10.569 & 3.713\% & 2s & 15.523 & 5.008\% & 8s & & RouteFinder & 18.910 & 1.713\% & 2s & 30.705 & 2.978\% & 9s \\
					& CaDA & 10.543 & 3.461\% & 2s & 15.525 & 5.001\% & 8s & & CaDA & 18.848 & 1.376\% & 2s & 30.520 & 2.359\% & 9s \\
					& FiLMMeD-CaDA & \underline{10.512} & \underline{3.152\%} & 4s & \underline{15.424} & \underline{4.318\%} & 19s & & FiLMMeD-CaDA & \underline{18.848} & \underline{1.369\%} & 4s & \cellcolor{gray!30}\underline{30.499} & \cellcolor{gray!30}\underline{2.287\%} & 19s \\
					& CaDA† & 10.511 & 3.143\% & 5s & 15.397 & 4.153\% & 29s & & CaDA† & 18.869 & 1.484\% & 5s & 30.556 & 2.483\% & 32s \\
					& FiLMMeD-CaDA† & \cellcolor{gray!30}\underline{10.497} & \cellcolor{gray!30}\underline{3.005\%} & 5s & \cellcolor{gray!30}\underline{15.342} & \cellcolor{gray!30}\underline{3.776\%} & 29s & & FiLMMeD-CaDA† & \cellcolor{gray!30}\underline{18.832} & \cellcolor{gray!30}\underline{1.288\%} & 6s & \underline{30.523} & \underline{2.371\%} & 32s \\
					\midrule
					\multirow{9}*{\rotatebox{90}{VRPBTW}}
					& HGS-PyVRP & 18.292 & * & 10.4m & 29.467 & * & 20.8m & \multirow{9}*{\rotatebox{90}{VRPLTW}} & HGS-PyVRP & 16.356 & * & 10.4m & 25.757 & * & 20.8m \\
					& OR-Tools & 18.366 & 0.383\% & 10.4m & 29.945 & 1.597\% & 20.8m & & OR-Tools & 16.441 & 0.499\% & 10.4m & 26.259 & 1.899\% & 20.8m \\
					& MTPOMO & 18.649 & 1.938\% & 2s & 30.478 & 3.426\% & 9s & & MTPOMO & 16.832 & 2.877\% & 2s & 26.913 & 4.455\% & 9s \\
					& MVMoE & 18.632 & 1.841\% & 3s & 30.437 & 3.284\% & 12s & & MVMoE & 16.817 & 2.783\% & 3s & 26.866 & 4.272\% & 12s \\
					& RouteFinder & 18.573 & 1.517\% & 2s & 30.249 & 2.641\% & 9s & & RouteFinder & 16.728 & 2.248\% & 2s & 26.706 & 3.645\% & 9s \\
					& CaDA & 18.500 & 1.117\% & 2s & 30.059 & 1.999\% & 9s & & CaDA & 16.669 & 1.879\% & 2s & 26.540 & 2.995\% & 9s \\
					& FiLMMeD-CaDA & \underline{18.497} & \underline{1.105\%} & 4s & \cellcolor{gray!30}\underline{30.042} & \cellcolor{gray!30}\underline{1.938\%} & 21s & & FiLMMeD-CaDA & \underline{16.661} & \underline{1.836\%} & 4s & \cellcolor{gray!30}\underline{26.498} & \cellcolor{gray!30}\underline{2.836\%} & 20s \\
					& CaDA† & 18.524 & 1.251\% & 6s & 30.115 & 2.188\% & 32s & & CaDA† & 16.684 & 1.984\% & 6s & 26.548 & 3.043\% & 31s \\
					& FiLMMeD-CaDA† & \cellcolor{gray!30}\underline{18.484} & \cellcolor{gray!30}\underline{1.034\%} & 6s & \underline{30.076} & \underline{2.059\%} & 31s & & FiLMMeD-CaDA† & \cellcolor{gray!30}\underline{16.636} & \cellcolor{gray!30}\underline{1.688\%} & 6s & \underline{26.510} & \underline{2.883\%} & 31s \\
					\midrule
					\multirow{9}*{\rotatebox{90}{OVRPB}}
					& HGS-PyVRP & 6.898 & * & 10.4m & 10.335 & * & 20.8m & \multirow{9}*{\rotatebox{90}{OVRPBL}} & HGS-PyVRP & 6.899 & * & 10.4m & 10.335 & * & 20.8m \\
					& OR-Tools & 6.928 & 0.412\% & 10.4m & 10.577 & 2.315\% & 20.8m & & OR-Tools & 6.927 & 0.386\% & 10.4m & 10.582 & 2.363\% & 20.8m \\
					& MTPOMO & 7.105 & 2.973\% & 2s & 10.882 & 5.264\% & 8s & & MTPOMO & 7.112 & 3.053\% & 2s & 10.888 & 5.318\% & 8s \\
					& MVMoE & 7.089 & 2.744\% & 3s & 10.841 & 4.869\% & 11s & & MVMoE & 7.094 & 2.799\% & 3s & 10.847 & 4.929\% & 11s \\
					& RouteFinder & 7.065 & 2.385\% & 2s & 10.774 & 4.233\% & 8s & & RouteFinder & 7.068 & 2.417\% & 2s & 10.778 & 4.266\% & 8s \\
					& CaDA & 7.049 & 2.159\% & 2s & 10.762 & 4.099\% & 8s & & CaDA & 7.051 & 2.166\% & 2s & 10.762 & 4.102\% & 8s \\
					& FiLMMeD-CaDA & \underline{7.027} & \underline{1.844\%} & 4s & \underline{10.695} & \underline{3.456\%} & 20s & & FiLMMeD-CaDA & \underline{7.028} & \underline{1.846\%} & 4s & \underline{10.695} & \underline{3.459\%} & 20s \\
					& CaDA† & 7.038 & 2.003\% & 6s & 10.708 & 3.592\% & 30s & & CaDA† & 7.040 & 2.011\% & 5s & 10.709 & 3.598\% & 29s \\
					& FiLMMeD-CaDA† & \cellcolor{gray!30}\underline{7.024} & \cellcolor{gray!30}\underline{1.802\%} & 6s & \cellcolor{gray!30}\underline{10.655} & \cellcolor{gray!30}\underline{3.071\%} & 29s & & FiLMMeD-CaDA† & \cellcolor{gray!30}\underline{7.025} & \cellcolor{gray!30}\underline{1.797\%} & 5s & \cellcolor{gray!30}\underline{10.655} & \cellcolor{gray!30}\underline{3.074\%} & 30s \\
					\midrule
					\multirow{9}*{\rotatebox{90}{OVRPBLTW}}
					& HGS-PyVRP & 11.668 & * & 10.4m & 19.156 & * & 20.8m & \multirow{9}*{\rotatebox{90}{OVRPBTW}} & HGS-PyVRP & 11.669 & * & 10.4m & 19.156 & * & 20.8m \\
					& OR-Tools & 11.681 & 0.106\% & 10.4m & 19.305 & 0.767\% & 20.8m & & OR-Tools & 11.682 & 0.109\% & 10.4m & 19.303 & 0.757\% & 20.8m \\
					& MTPOMO & 11.823 & 1.315\% & 3s & 19.658 & 2.602\% & 9s & & MTPOMO & 11.823 & 1.307\% & 3s & 19.656 & 2.592\% & 9s \\
					& MVMoE & 11.816 & 1.249\% & 4s & 19.640 & 2.514\% & 12s & & MVMoE & 11.816 & 1.245\% & 4s & 19.637 & 2.499\% & 13s \\
					& RouteFinder & 11.789 & 1.017\% & 2s & 19.554 & 2.061\% & 9s & & RouteFinder & 11.790 & 1.027\% & 2s & 19.555 & 2.062\% & 9s \\
					& CaDA & 11.760 & 0.771\% & 2s & \cellcolor{gray!30}{19.435} & \cellcolor{gray!30}{1.439\%} & 9s & & CaDA & 11.761 & 0.779\% & 2s & 19.436 & 1.441\% & 9s \\
					& FiLMMeD-CaDA & \underline{11.754} & \underline{0.721\%} & 4s & 19.437 & 1.441\% & 22s & & FiLMMeD-CaDA & \underline{11.754} & \underline{0.719\%} & 4s & \cellcolor{gray!30}\underline{19.435} & \cellcolor{gray!30}\underline{1.434\%} & 22s \\
					& CaDA† & 11.773 & 0.885\% & 7s & 19.479 & 1.666\% & 33s & & CaDA† & 11.774 & 0.892\% & 6s & 19.478 & 1.663\% & 34s \\
					& FiLMMeD-CaDA† & \cellcolor{gray!30}\underline{11.747} & \cellcolor{gray!30}\underline{0.665\%} & 6s & \underline{19.456} & \underline{1.546\%} & 34s & & FiLMMeD-CaDA† & \cellcolor{gray!30}\underline{11.747} & \cellcolor{gray!30}\underline{0.655\%} & 7s & \underline{19.454} & \underline{1.537\%} & 33s \\
					\midrule
					\multirow{9}*{\rotatebox{90}{OVRPL}}
					& HGS-PyVRP & 6.507 & * & 10.4m & 9.724 & * & 20.8m & \multirow{9}*{\rotatebox{90}{OVRPLTW}} & HGS-PyVRP & 10.510 & * & 10.4m & 16.926 & * & 20.8m \\
					& OR-Tools & 6.552 & 0.668\% & 10.4m & 10.001 & 2.791\% & 20.8m & & OR-Tools & 10.497 & 0.114\% & 10.4m & 17.023 & 0.728\% & 20.8m \\
					& MTPOMO & 6.720 & 3.248\% & 2s & 10.224 & 5.112\% & 8s & & MTPOMO & 10.677 & 1.572\% & 2s & 17.442 & 3.020\% & 9s \\
					& MVMoE & 6.706 & 3.028\% & 3s & 10.184 & 4.693\% & 11s & & MVMoE & 10.677 & 1.564\% & 3s & 17.418 & 2.880\% & 12s \\
					& RouteFinder & 6.683 & 2.680\% & 2s & 10.121 & 4.054\% & 8s & & RouteFinder & 10.646 & 1.267\% & 2s & 17.328 & 2.352\% & 9s \\
					& CaDA & 6.671 & 2.475\% & 2s & 10.122 & 4.052\% & 8s & & CaDA & 10.613 & 0.961\% & 2s & 17.226 & 1.752\% & 9s \\
					& FiLMMeD-CaDA & \underline{6.650} & \underline{2.157\%} & 4s & \underline{10.078} & \underline{3.604\%} & 20s & & FiLMMeD-CaDA & \underline{10.609} & \underline{0.926\%} & 4s & \cellcolor{gray!30}\underline{17.214} & \cellcolor{gray!30}\underline{1.678\%} & 21s \\
					& CaDA† & 6.658 & 2.295\% & 5s & 10.081 & 3.647\% & 29s & & CaDA† & 10.623 & 1.060\% & 6s & 17.245 & 1.870\% & 32s \\
					& FiLMMeD-CaDA† & \cellcolor{gray!30}\underline{6.647} & \cellcolor{gray!30}\underline{2.122\%} & 5s & \cellcolor{gray!30}\underline{10.027} & \cellcolor{gray!30}\underline{3.089\%} & 29s & & FiLMMeD-CaDA† & \cellcolor{gray!30}\underline{10.598} & \cellcolor{gray!30}\underline{0.819\%} & 6s & \underline{17.232} & \underline{1.785\%} & 32s \\
					\bottomrule
			\end{tabular}}
		\end{scriptsize}
	\end{center}
	\vskip -0.1in
\end{table*}

After applying FiLM, the latent space undergoes a severe reorganization (shown on the right of Figure~\ref{tsne_fig_film}). The embeddings are now organized into very distinct and separated clusters corresponding to each constraint combination, suggesting that the affine transformations performed by FiLM successfully disentangle task specific features from general ones.

\begin{sloppypar}
After, we compared the learned representations of both FiLMMeD models and its baselines by extracting the learned node embeddings from the last encoder layer, that is, $h_i^{(L)}, \allowbreak \ \forall v_i \in \mathcal{V}$. Figure~\ref{tsne_fig} presents the t-SNE projections of these embeddings for FiLMMeD-MTPOMO and FiLMMeD-MVMoE and their respective baselines. Overall, we can see that FiLMMeD exhibits more structured and well-defined clusters, particularly for variants involving route length limits, backhauls, and time windows. This suggests that the FiLM mechanism effectively helps the model modulate its internal representations, allowing it to better adapt towards diverse constraint combinations. 
\end{sloppypar}

\subsection{Results on single-depot variants (\citet{Berto2025} setting)}

\begin{table*}[pos=!h]
	\caption{Testing results on Cordeau MDVRP benchmark instances. The best neural-based method results are shown in gray, and the models improved by FiLMMeD are underlined.}
	\label{mdvrplib}
	\begin{center}
		\begin{small}
			\renewcommand\arraystretch{1.15}
			\resizebox{0.95\textwidth}{!}{
				\begin{tabular}{l|ccc|ccc|ccc|ccc|ccc}
					\toprule
					\multirow{2}{*}{Variant} & \multicolumn{3}{c|}{HGS-PyVRP} & \multicolumn{3}{c|}{MTPOMO} & \multicolumn{3}{c|}{FiLMMeD-MTPOMO} & \multicolumn{3}{c|}{MVMoE/4E} & \multicolumn{3}{c}{FiLMMeD-MVMoE/4E} \\
					& Obj. & Gap & Time & Obj. & Gap & Time & Obj. & Gap & Time & Obj. & Gap & Time & Obj. & Gap & Time \\
					\midrule
					MDVRP & 1851.19 & * & 5m & 2034.98 & 9.031\% & 7s & 2046.08 & 10.062\% & 7s & \cellcolor{gray!30}1993.30 & \cellcolor{gray!30}6.951\% & 9s & 2044.86 & 9.969\% & 9s \\
					MDOVRP & 1386.02 & * & 5m & 1722.21 & 22.915\% & 7s & \underline{\cellcolor{gray!30}1541.42} & \underline{\cellcolor{gray!30}9.852\%} & 8s & 1762.79 & 26.103\% & 9s & \underline{1538.02} & \underline{9.874\%} & 10s \\
					MDVRPB & 1756.56 & * & 5m & 1964.27 & 10.999\% & 8s & 1984.85 & 12.472\% & 8s & \cellcolor{gray!30}1940.64 & \cellcolor{gray!30}9.465\% & 10s & 1974.97 & 11.452\% & 10s \\
					MDVRPL & 1898.82 & * & 5m & 2035.81 & 6.704\% & 9s & 2040.60 & 7.327\% & 8s & \cellcolor{gray!30}1993.35 & \cellcolor{gray!30}4.629\% & 11s & 2037.33 & 7.180\% & 11s \\
					MDVRPTW & 3098.46 & * & 5m & 3476.98 & 11.196\% & 10s & \underline{3414.85} & \underline{9.426\%} & 10s & 3515.88 & 12.521\% & 13s & \underline{\cellcolor{gray!30}3389.29} & \underline{\cellcolor{gray!30}8.600\%} & 12s \\
					MDOVRPTW & 2153.43 & * & 5m & 2761.57 & 25.751\% & 10s & \underline{2394.81} & \underline{9.788\%} & 11s & 2789.77 & 27.329\% & 13s & \underline{\cellcolor{gray!30}2342.40} & \underline{\cellcolor{gray!30}7.757\%} & 13s \\
					MDOVRPB & 1374.08 & * & 5m & 1782.25 & 28.145\% & 8s & \underline{1573.51} & \underline{13.156\%} & 8s & 1809.60 & 30.541\% & 10s & \underline{\cellcolor{gray!30}1559.18} & \underline{\cellcolor{gray!30}12.140\%} & 10s \\
					MDOVRPL & 1385.97 & * & 5m & 1722.21 & 22.918\% & 8s & \underline{\cellcolor{gray!30}1540.99} & \underline{\cellcolor{gray!30}10.177\%} & 9s & 1762.79 & 26.106\% & 10s & \underline{1544.24} & \underline{10.244\%} & 11s \\
					MDVRPBL & 1760.30 & * & 5m & 1965.26 & 10.469\% & 9s & 1990.50 & 12.380\% & 9s & \cellcolor{gray!30}1953.89 & \cellcolor{gray!30}10.018\% & 10s & 1980.99 & 11.644\% & 10s \\
					MDVRPBTW & 3120.24 & * & 5m & 3715.97 & 18.304\% & 11s & \underline{3644.22} & \underline{15.843\%} & 11s & 3873.42 & 23.343\% & 14s & \underline{\cellcolor{gray!30}3632.57} & \underline{\cellcolor{gray!30}15.488\%} & 13s \\
					MDVRPLTW & 3086.58 & * & 5m & 3454.22 & 10.749\% & 11s & \underline{3394.27} & \underline{8.823\%} & 12s & 3490.28 & 11.739\% & 14s & \underline{\cellcolor{gray!30}3370.09} & \underline{\cellcolor{gray!30}8.230\%} & 14s \\
					MDOVRPBL & 1373.20 & * & 5m & 1798.56 & 29.549\% & 8s & \underline{1575.64} & \underline{13.698\%} & 9s & 1813.13 & 31.042\% & 10s & \underline{\cellcolor{gray!30}1570.14} & \underline{\cellcolor{gray!30}13.322\%} & 11s \\
					MDOVRPBTW & 2164.82 & * & 5m & 2917.39 & 32.298\% & 11s & \underline{2497.26} & \underline{14.340\%} & 12s & 2979.77 & 36.011\% & 13s & \underline{\cellcolor{gray!30}2479.57} & \underline{\cellcolor{gray!30}13.378\%} & 14s \\
					MDOVRPLTW & 2167.34 & * & 5m & 2775.88 & 25.798\% & 11s & \underline{2412.36} & \underline{10.179\%} & 12s & 2785.94 & 26.917\% & 14s & \underline{\cellcolor{gray!30}2378.96} & \underline{\cellcolor{gray!30}8.680\%} & 15s \\
					MDVRPBLTW & 3123.13 & * & 5m & 3735.69 & 18.527\% & 12s & \underline{3677.44} & \underline{17.156\%} & 12s & 3820.79 & 21.443\% & 15s & \underline{\cellcolor{gray!30}3638.68} & \underline{\cellcolor{gray!30}15.225\%} & 14s \\
					MDOVRPBLTW & 2156.32 & * & 5m & 2901.50 & 31.718\% & 12s & \underline{2520.55} & \underline{14.978\%} & 13s & 2928.39 & 33.111\% & 14s & \underline{\cellcolor{gray!30}2493.31} & \underline{\cellcolor{gray!30}13.845\%} & 15s \\
					MDVRPI & 1870.61 & * & 5m & 2048.13 & 9.479\% & 8s & 2073.85 & 10.965\% & 9s & \cellcolor{gray!30}2040.77 & \cellcolor{gray!30}8.963\% & 11s & 2075.04 & 11.000\% & 11s \\
					MDVRPIB & 1771.57 & * & 5m & \cellcolor{gray!30}2009.43 & \cellcolor{gray!30}12.889\% & 9s & 2026.46 & 14.129\% & 9s & 2033.86 & 13.889\% & 11s & 2023.05 & 14.125\% & 11s \\
					MDVRPIL & 1873.33 & * & 5m & 2052.86 & 9.624\% & 10s & 2073.89 & 10.805\% & 10s & \cellcolor{gray!30}2033.39 & \cellcolor{gray!30}8.586\% & 12s & 2068.03 & 10.385\% & 12s \\
					MDVRPITW & 3084.07 & * & 5m & 4437.59 & 44.428\% & 14s & \underline{3394.66} & \underline{8.963\%} & 11s & 3609.96 & 16.672\% & 14s & \underline{\cellcolor{gray!30}3359.50} & \underline{\cellcolor{gray!30}7.986\%} & 13s \\
					MDVRPIBL & 1776.03 & * & 5m & 2028.43 & 13.655\% & 10s & \underline{2011.70} & \underline{13.287\%} & 10s & 2035.19 & 14.109\% & 12s & \underline{\cellcolor{gray!30}2004.44} & \underline{\cellcolor{gray!30}13.279\%} & 12s \\
					MDVRPIBTW & 3097.93 & * & 5m & 4797.86 & 55.065\% & 15s & \underline{3647.92} & \underline{16.986\%} & 11s & 3931.74 & 26.356\% & 14s & \underline{\cellcolor{gray!30}3588.74} & \underline{\cellcolor{gray!30}15.070\%} & 14s \\
					MDVRPILTW & 3071.28 & * & 5m & 4496.17 & 44.037\% & 16s & \underline{3415.15} & \underline{10.023\%} & 12s & 3589.49 & 16.553\% & 15s & \underline{\cellcolor{gray!30}3369.62} & \underline{\cellcolor{gray!30}8.674\%} & 15s \\
					MDVRPIBLTW & 3069.37 & * & 5m & 4848.59 & 59.088\% & 16s & \underline{3578.70} & \underline{15.900\%} & 12s & 3869.80 & 26.123\% & 16s & \underline{\cellcolor{gray!30}3550.08} & \underline{\cellcolor{gray!30}14.561\%} & 15s \\
					\midrule
					Avg. & 2227.94 & * & 5m & 2811.83 & 23.472\% & 10s & \underline{2519.65} & \underline{12.113\%} & 10s & 2681.58 & 19.522\% & 12s & \underline{\cellcolor{gray!30}2500.55} & \underline{\cellcolor{gray!30}11.338\%} & 12s \\
					\bottomrule
				\end{tabular}
			}
		\end{small}
	\end{center}
\end{table*}

Like \citet{Liu2025curvature}, we evaluate our approach across two distinct standardized experimental settings from the literature. This section details the results for 16 single-depot VRP variants using the setting defined by \citet{Berto2025}. For these experiments, we trained a variation of CaDA \citep{Li2025} from scratch for 300 epochs. This version, denoted as \textit{FiLMMeD-CaDA}, incorporates the key contributions proposed in this work while retaining the core dual-attention architecture of CaDA. Specifically, we added the proposed FiLM mechanism to the encoder. Furthermore, instead of the traditional REINFORCE loss, we employed PO for training, setting $\alpha = 0.03$, following the recommendations of \citet{Pan2025}. Since our curriculum was designed for the MDVRP, we maintained the same mixed batch training strategy from the original work (where a single batch can contain instances from multiple different variants). All the key hyper-parameters of CaDA were kept the same for FiLMMeD-CaDA. We also include an additional FiLMMeD-CaDA† variation, which combines FiLMMeD-CaDA with ReLD \citep{Huang2025rethinking}. For neural baselines, we include results on MTPOMO, MVMoE, RouteFinder (in particular, their modern Transformer-based Encoder version), CaDA and CaDA†.

Table~\ref{vrp_berto} presents the main results on 16 single-depot VRP variants, of 50 and 100 nodes. Overall, both FiLMMeD-CaDA and FiLMMeD-CaDA† consistently outperformed their non-FiLMMeD counterparts across the majority of cases, further consolidating the effectiveness of our proposed contributions. Additionally, FiLMMeD-CaDA† displayed the best overall results across all other baselines, with an average performance gap of 1.46\% on 50-node instances, and 2.35\% on 100-node instances.

\begin{table*}[pos=!ht]
	\centering
	\tiny
	\caption{Results on CVRPLib instances from the Set-X~\citep{Uchoa2017}. The best individual results are shown in gray.}
	\label{setX_results}
	\resizebox{0.92\textwidth}{!}{\begin{tabular}{ll|cccccccccccccccc}
			\toprule
			\multicolumn{2}{c|}{Set-X} & \multicolumn{2}{c}{MTPOMO} & \multicolumn{2}{c}{MVMoE} & \multicolumn{2}{c}{RouteFinder} & \multicolumn{2}{c}{CaDA} & \multicolumn{2}{c}{FiLMMeD-CaDA} & \multicolumn{2}{c}{CaDA†} & \multicolumn{2}{c}{FiLMMeD-CaDA†} & \multicolumn{2}{c}{DBDA} \\
			\midrule
			Instance & Opt. & Cost & Gap & Cost & Gap & Cost & Gap & Cost & Gap & Cost & Gap & Cost & Gap & Cost & Gap & Cost & Gap \\
			\midrule
			X-n101-k25 & 27591 & 29470 & 6.810\% & 29076 & 5.382\% & 29048 & 5.281\% & 28944 & 4.904\% & 29095 & 5.451\% & \cellcolor{gray!30}28840 & \cellcolor{gray!30}4.527\% & 28848 & 4.556\% & 28965 & 4.980\% \\
			X-n106-k14 & 26362 & 28029 & 6.323\% & 27443 & 4.101\% & 27159 & 3.023\% & 27042 & 2.579\% & 26826 & 1.760\% & 26815 & 1.718\% & \cellcolor{gray!30}26733 & \cellcolor{gray!30}1.407\% & 26740 & 1.434\% \\
			X-n110-k13 & 14971 & 15100 & 0.862\% & 15327 & 2.378\% & 15314 & 2.291\% & 15229 & 1.723\% & 15203 & 1.549\% & 15343 & 2.485\% & 15197 & 1.510\% & \cellcolor{gray!30}15083 & \cellcolor{gray!30}0.748\% \\
			X-n115-k10 & 12747 & 13433 & 5.382\% & 13475 & 5.711\% & 13338 & 4.636\% & 13060 & 2.455\% & 13244 & 3.899\% & 13210 & 3.632\% & \cellcolor{gray!30}13080 & \cellcolor{gray!30}2.612\% & 13149 & 3.154\% \\
			X-n120-k6 & 13332 & 14051 & 5.393\% & 13782 & 3.375\% & 13765 & 3.248\% & 13678 & 2.595\% & \cellcolor{gray!30}13552 & \cellcolor{gray!30}1.650\% & 13771 & 3.293\% & 13628 & 2.220\% & 13575 & 1.823\% \\
			X-n125-k30 & 55539 & 59015 & 6.259\% & 58200 & 4.791\% & 58570 & 5.457\% & \cellcolor{gray!30}57748 & \cellcolor{gray!30}3.977\% & 57815 & 4.098\% & 57800 & 4.071\% & 57816 & 4.100\% & 57833 & 4.130\% \\
			X-n129-k18 & 28940 & 30176 & 4.271\% & 29334 & 1.361\% & 29457 & 1.786\% & 29500 & 1.935\% & 29397 & 1.579\% & 29399 & 1.586\% & \cellcolor{gray!30}29261 & \cellcolor{gray!30}1.109\% & 29460 & 1.797\% \\
			X-n134-k13 & 10916 & 11707 & 7.246\% & 11462 & 5.002\% & 11624 & 6.486\% & 11652 & 6.742\% & 11367 & 4.132\% & 11231 & 2.886\% & \cellcolor{gray!30}11224 & \cellcolor{gray!30}2.822\% & 11483 & 5.194\% \\
			X-n139-k10 & 13590 & 14058 & 3.444\% & 14099 & 3.745\% & \cellcolor{gray!30}13812 & \cellcolor{gray!30}1.634\% & 13940 & 2.575\% & 13929 & 2.494\% & 14004 & 3.046\% & 13856 & 1.957\% & 13944 & 2.605\% \\
			X-n143-k7 & 15700 & 16626 & 5.898\% & 16349 & 4.134\% & 16257 & 3.548\% & 16189 & 3.115\% & 16107 & 2.592\% & 16222 & 3.325\% & \cellcolor{gray!30}16053 & \cellcolor{gray!30}2.248\% & 16056 & 2.268\% \\
			X-n148-k46 & 43448 & 46648 & 7.365\% & 45893 & 5.627\% & \cellcolor{gray!30}45026 & \cellcolor{gray!30}3.632\% & 45606 & 4.967\% & 45704 & 5.193\% & 45226 & 4.092\% & 45824 & 5.469\% & 46689 & 7.459\% \\
			X-n153-k22 & 21220 & 23514 & 10.811\% & 23661 & 11.503\% & 23478 & 10.641\% & 23142 & 9.057\% & 23019 & 8.478\% & \cellcolor{gray!30}22750 & \cellcolor{gray!30}7.210\% & 23026 & 8.511\% & 22783 & 7.366\% \\
			X-n157-k13 & 16876 & 17922 & 6.198\% & 17439 & 3.336\% & 17315 & 2.601\% & 17295 & 2.483\% & 17270 & 2.335\% & 17205 & 1.950\% & 17193 & 1.878\% & \cellcolor{gray!30}17120 & \cellcolor{gray!30}1.446\% \\
			X-n162-k11 & 14138 & 14616 & 3.381\% & 14705 & 4.010\% & 14664 & 3.720\% & 14704 & 4.003\% & 14713 & 4.067\% & 14494 & 2.518\% & \cellcolor{gray!30}14477 & \cellcolor{gray!30}2.398\% & 14630 & 3.480\% \\
			X-n167-k10 & 20557 & 21662 & 5.375\% & 21504 & 4.607\% & 21425 & 4.222\% & \cellcolor{gray!30}21078 & \cellcolor{gray!30}2.534\% & 21180 & 3.029\% & 21330 & 3.760\% & 21249 & 3.366\% & 21209 & 3.172\% \\
			X-n172-k51 & 45607 & 48960 & 7.352\% & 47883 & 4.990\% & 48162 & 5.602\% & 48198 & 5.681\% & 47543 & 4.245\% & 47688 & 4.563\% & \cellcolor{gray!30}47559 & \cellcolor{gray!30}4.280\% & 48513 & 6.372\% \\
			X-n176-k26 & 47812 & 51989 & 8.736\% & 52117 & 9.004\% & 51501 & 7.716\% & 51120 & 6.919\% & 51711 & 8.155\% & 51381 & 7.465\% & \cellcolor{gray!30}51024 & \cellcolor{gray!30}6.718\% & 53069 & 10.995\% \\
			X-n181-k23 & 25569 & 26572 & 3.923\% & 26456 & 3.469\% & 26097 & 2.065\% & 26262 & 2.710\% & 26024 & 1.780\% & 25968 & 1.560\% & \cellcolor{gray!30}25953 & \cellcolor{gray!30}1.502\% & 26088 & 2.030\% \\
			X-n186-k15 & 24145 & 25236 & 4.519\% & 25151 & 4.166\% & 25153 & 4.175\% & 25345 & 4.970\% & 25197 & 4.357\% & 24974 & 3.433\% & \cellcolor{gray!30}24868 & \cellcolor{gray!30}2.994\% & 25218 & 4.444\% \\
			X-n190-k8 & 16980 & 18369 & 8.180\% & 19078 & 12.356\% & 17871 & 5.247\% & 17882 & 5.312\% & 17812 & 4.900\% & 17834 & 5.029\% & 17646 & 3.922\% & \cellcolor{gray!30}17379 & \cellcolor{gray!30}2.350\% \\
			X-n195-k51 & 44225 & 48310 & 9.237\% & 46974 & 6.216\% & 47396 & 7.170\% & 46723 & 5.648\% & 46400 & 4.919\% & \cellcolor{gray!30}46124 & \cellcolor{gray!30}4.294\% & 46316 & 4.728\% & 47612 & 7.659\% \\
			X-n200-k36 & 58578 & 62041 & 5.912\% & 61627 & 5.205\% & 61139 & 4.372\% & 61010 & 4.152\% & 60833 & 3.850\% & 60958 & 4.063\% & \cellcolor{gray!30}60773 & \cellcolor{gray!30}3.747\% & 60926 & 4.008\% \\
			X-n204-k19 & 19565 & 20652 & 5.556\% & 20584 & 5.208\% & 20531 & 4.937\% & 20735 & 5.980\% & 20491 & 4.733\% & \cellcolor{gray!30}20342 & \cellcolor{gray!30}3.971\% & 20356 & 4.043\% & 20671 & 5.653\% \\
			X-n209-k16 & 30656 & 32333 & 5.470\% & 32358 & 5.552\% & 31876 & 3.980\% & 32184 & 4.984\% & 32248 & 5.193\% & 31890 & 4.025\% & \cellcolor{gray!30}31735 & \cellcolor{gray!30}3.520\% & 31939 & 4.185\% \\
			X-n214-k11 & 10856 & 11699 & 7.765\% & 11597 & 6.826\% & 11668 & 7.480\% & 11748 & 8.217\% & 11659 & 7.397\% & 11487 & 5.812\% & 11469 & 5.647\% & \cellcolor{gray!30}11411 & \cellcolor{gray!30}5.112\% \\
			X-n219-k73 & 117595 & 121980 & 3.729\% & 124434 & 5.816\% & 120344 & 2.338\% & 120011 & 2.055\% & 122996 & 4.593\% & 120304 & 2.304\% & \cellcolor{gray!30}119142 & \cellcolor{gray!30}1.316\% & 119959 & 2.010\% \\
			X-n223-k34 & 40437 & 43381 & 7.280\% & 42694 & 5.582\% & 42251 & 4.486\% & 42273 & 4.540\% & 42178 & 4.305\% & 42200 & 4.360\% & \cellcolor{gray!30}41926 & \cellcolor{gray!30}3.682\% & 42075 & 4.051\% \\
			X-n228-k23 & 25742 & 28523 & 10.803\% & 28033 & 8.900\% & 28699 & 11.487\% & 27821 & 8.076\% & 27523 & 6.919\% & \cellcolor{gray!30}27547 & \cellcolor{gray!30}7.012\% & 27724 & 7.699\% & 28169 & 9.428\% \\
			X-n233-k16 & 19230 & 20644 & 7.353\% & 20656 & 7.415\% & 20761 & 7.962\% & 20285 & 5.486\% & 20538 & 6.801\% & \cellcolor{gray!30}20118 & \cellcolor{gray!30}4.618\% & 20307 & 5.601\% & 20424 & 6.209\% \\
			X-n237-k14 & 27042 & 30047 & 11.112\% & 29772 & 10.095\% & 29595 & 9.441\% & 30282 & 11.981\% & 29952 & 10.762\% & 29566 & 9.334\% & 29419 & 8.790\% & \cellcolor{gray!30}28179 & \cellcolor{gray!30}4.205\% \\
			X-n242-k48 & 82751 & 88179 & 6.559\% & 87497 & 5.735\% & 85704 & 3.569\% & 85813 & 3.700\% & 85576 & 3.414\% & 85775 & 3.654\% & \cellcolor{gray!30}85636 & \cellcolor{gray!30}3.486\% & 85665 & 3.521\% \\
			X-n247-k50 & 37274 & 41610 & 11.633\% & 40973 & 9.924\% & 40642 & 9.036\% & 39918 & 7.093\% & 40444 & 8.504\% & 40359 & 8.277\% & \cellcolor{gray!30}39666 & \cellcolor{gray!30}6.417\% & 41103 & 10.273\% \\
			\midrule
			\multicolumn{2}{c|}{Avg. Gap ($n<251$)} & \multicolumn{2}{c}{6.566\%} & \multicolumn{2}{c}{5.829\%} & \multicolumn{2}{c}{5.061\%} & \multicolumn{2}{c}{4.772\%} & \multicolumn{2}{c}{\underline{4.399\%}} & \multicolumn{2}{c}{4.309\%} & \multicolumn{2}{c}{\cellcolor{gray!30}\underline{3.758\%}} & \multicolumn{2}{c}{4.486\%} \\
			\midrule
			X-n251-k28 & 38684 & 41211 & 6.532\% & 41330 & 6.840\% & 40127 & 3.730\% & 40359 & 4.330\% & 40247 & 4.040\% & 40003 & 3.410\% & \cellcolor{gray!30}39912 & \cellcolor{gray!30}3.174\% & 40220 & 3.971\% \\
			X-n256-k16 & 18839 & 20400 & 8.286\% & 20559 & 9.130\% & 19994 & 6.131\% & 20372 & 8.137\% & 20086 & 6.619\% & 20061 & 6.487\% & \cellcolor{gray!30}19838 & \cellcolor{gray!30}5.303\% & 20087 & 6.625\% \\
			X-n261-k13 & 26558 & 28741 & 8.220\% & 28524 & 7.403\% & 28510 & 7.350\% & 28833 & 8.566\% & 28631 & 7.805\% & 28256 & 6.394\% & \cellcolor{gray!30}27962 & \cellcolor{gray!30}5.287\% & 28310 & 6.597\% \\
			X-n266-k58 & 75478 & 84617 & 12.108\% & 82048 & 8.705\% & 79832 & 5.769\% & 80115 & 6.144\% & 80247 & 6.318\% & 79480 & 5.302\% & 79578 & 5.432\% & \cellcolor{gray!30}79363 & \cellcolor{gray!30}5.147\% \\
			X-n270-k35 & 35291 & 38146 & 8.090\% & 38333 & 8.620\% & 37382 & 5.925\% & 37674 & 6.752\% & 37360 & 5.862\% & 36927 & 4.636\% & \cellcolor{gray!30}36798 & \cellcolor{gray!30}4.270\% & 37138 & 5.234\% \\
			X-n275-k28 & 21245 & 24688 & 16.206\% & 25021 & 17.774\% & 24187 & 13.848\% & 24482 & 15.237\% & 24347 & 14.601\% & 24127 & 13.566\% & 24185 & 13.839\% & \cellcolor{gray!30}22405 & \cellcolor{gray!30}5.460\% \\
			X-n280-k17 & 33503 & 36677 & 9.474\% & 36636 & 9.351\% & 36653 & 9.402\% & 36081 & 7.695\% & 35844 & 6.987\% & \cellcolor{gray!30}35370 & \cellcolor{gray!30}5.573\% & 35545 & 6.095\% & 35769 & 6.764\% \\
			X-n284-k15 & 20226 & 22474 & 11.114\% & 22583 & 11.653\% & 22154 & 9.532\% & 22295 & 10.229\% & 21916 & 8.355\% & 21619 & 6.887\% & 21697 & 7.273\% & \cellcolor{gray!30}21605 & \cellcolor{gray!30}6.818\% \\
			X-n289-k60 & 95151 & 104159 & 9.467\% & 102202 & 7.410\% & 100418 & 5.535\% & 99739 & 4.822\% & 100808 & 5.945\% & 100451 & 5.570\% & \cellcolor{gray!30}98873 & \cellcolor{gray!30}3.912\% & 99962 & 5.056\% \\
			X-n294-k50 & 47161 & 52769 & 11.891\% & 50886 & 7.898\% & 50637 & 7.370\% & 49929 & 5.869\% & 49980 & 5.977\% & \cellcolor{gray!30}49318 & \cellcolor{gray!30}4.574\% & 49614 & 5.201\% & 50028 & 6.079\% \\
			X-n298-k31 & 34231 & 37652 & 9.994\% & 37344 & 9.094\% & 37163 & 8.565\% & 36993 & 8.069\% & 36497 & 6.620\% & 36189 & 5.720\% & \cellcolor{gray!30}36084 & \cellcolor{gray!30}5.413\% & 36601 & 6.924\% \\
			X-n303-k21 & 21736 & 23556 & 8.373\% & 23263 & 7.025\% & 23442 & 7.849\% & 23748 & 9.257\% & 23295 & 7.174\% & 22937 & 5.525\% & \cellcolor{gray!30}22879 & \cellcolor{gray!30}5.259\% & 23312 & 7.251\% \\
			X-n308-k13 & 25859 & 28736 & 11.126\% & 28518 & 10.283\% & 28326 & 9.540\% & 28913 & 11.810\% & 28219 & 9.125\% & 27714 & 7.174\% & \cellcolor{gray!30}27586 & \cellcolor{gray!30}6.679\% & 28394 & 9.803\% \\
			X-n313-k71 & 94043 & 102253 & 8.730\% & 100620 & 6.994\% & 99564 & 5.871\% & 98899 & 5.164\% & 99511 & 5.814\% & 98667 & 4.917\% & 97780 & 3.974\% & \cellcolor{gray!30}98535 & \cellcolor{gray!30}4.777\% \\
			X-n317-k53 & 78355 & 82587 & 5.401\% & 83632 & 6.735\% & 80690 & 2.980\% & 80542 & 2.791\% & 81804 & 4.402\% & 80206 & 2.362\% & 80138 & 2.276\% & \cellcolor{gray!30}79907 & \cellcolor{gray!30}1.981\% \\
			X-n322-k28 & 29834 & 32593 & 9.248\% & 33497 & 12.278\% & 32658 & 9.466\% & 33206 & 11.303\% & 32959 & 10.474\% & \cellcolor{gray!30}32047 & \cellcolor{gray!30}7.418\% & 32089 & 7.558\% & 32746 & 9.761\% \\
			X-n327-k20 & 27532 & 30646 & 11.310\% & 30603 & 11.154\% & 29784 & 8.180\% & 30953 & 12.426\% & 30280 & 9.980\% & 29763 & 8.103\% & \cellcolor{gray!30}29522 & \cellcolor{gray!30}7.228\% & 30349 & 10.232\% \\
			X-n331-k15 & 31102 & 34734 & 11.678\% & 33636 & 8.147\% & 34048 & 9.472\% & 34578 & 11.176\% & 34200 & 9.960\% & 33203 & 6.755\% & \cellcolor{gray!30}33068 & \cellcolor{gray!30}6.321\% & 33794 & 8.655\% \\
			X-n336-k84 & 139111 & 152846 & 9.873\% & 149229 & 7.273\% & 146620 & 5.398\% & 146707 & 5.460\% & 149982 & 7.815\% & 146345 & 5.200\% & \cellcolor{gray!30}144870 & \cellcolor{gray!30}4.140\% & 147474 & 6.012\% \\
			X-n344-k43 & 42050 & 46619 & 10.866\% & 46947 & 11.646\% & 44914 & 6.811\% & 45571 & 8.373\% & 45044 & 7.121\% & 44563 & 5.976\% & \cellcolor{gray!30}44438 & \cellcolor{gray!30}5.679\% & 44850 & 6.659\% \\
			X-n351-k40 & 25896 & 29243 & 12.925\% & 28373 & 9.565\% & 28236 & 9.036\% & 28059 & 8.353\% & 28101 & 8.515\% & 27495 & 6.175\% & 27585 & 6.522\% & \cellcolor{gray!30}27676 & \cellcolor{gray!30}6.874\% \\
			X-n359-k29 & 51505 & 55778 & 8.296\% & 56165 & 9.048\% & 55122 & 7.023\% & 55183 & 7.141\% & 54884 & 6.561\% & 54125 & 5.087\% & 53857 & 4.567\% & \cellcolor{gray!30}53808 & \cellcolor{gray!30}4.471\% \\
			X-n367-k17 & 22814 & 26132 & 14.544\% & 25588 & 12.159\% & 25522 & 11.870\% & 25534 & 11.923\% & 25094 & 9.994\% & \cellcolor{gray!30}24323 & \cellcolor{gray!30}6.614\% & 24443 & 7.140\% & 24893 & 9.113\% \\
			X-n376-k94 & 147713 & 156857 & 6.190\% & 156546 & 5.980\% & 151975 & 2.885\% & 151390 & 2.489\% & 152539 & 3.267\% & 151129 & 2.313\% & 149853 & 1.449\% & \cellcolor{gray!30}150506 & \cellcolor{gray!30}1.891\% \\
			X-n384-k52 & 65940 & 73705 & 11.776\% & 73570 & 11.571\% & 70471 & 6.871\% & 70611 & 7.084\% & 70044 & 6.223\% & \cellcolor{gray!30}68838 & \cellcolor{gray!30}4.395\% & 69524 & 5.435\% & 70002 & 6.160\% \\
			X-n393-k38 & 38260 & 43533 & 13.782\% & 44638 & 16.670\% & 41552 & 8.604\% & 42934 & 12.216\% & 41929 & 9.590\% & 41394 & 8.191\% & \cellcolor{gray!30}40981 & \cellcolor{gray!30}7.112\% & 42105 & 10.050\% \\
			X-n401-k29 & 66154 & 71565 & 8.179\% & 71787 & 8.515\% & 69430 & 4.952\% & 69875 & 5.625\% & 69596 & 5.203\% & 68591 & 3.684\% & \cellcolor{gray!30}68410 & \cellcolor{gray!30}3.410\% & 69872 & 5.620\% \\
			X-n411-k19 & 19712 & 23869 & 21.089\% & 23139 & 17.385\% & 22849 & 15.914\% & 23521 & 19.323\% & 23198 & 17.685\% & \cellcolor{gray!30}21621 & \cellcolor{gray!30}9.684\% & 21851 & 10.851\% & 22508 & 14.184\% \\
			X-n420-k130 & 107798 & 122761 & 13.881\% & 116362 & 7.944\% & 117418 & 8.924\% & 115012 & 6.692\% & 115837 & 7.457\% & 114105 & 5.851\% & \cellcolor{gray!30}113290 & \cellcolor{gray!30}5.095\% & 116654 & 8.215\% \\
			X-n429-k61 & 65449 & 74261 & 13.464\% & 74158 & 13.307\% & 70164 & 7.204\% & 70969 & 8.434\% & 70546 & 7.792\% & \cellcolor{gray!30}69282 & \cellcolor{gray!30}5.856\% & 69491 & 6.176\% & 70129 & 7.151\% \\
			X-n439-k37 & 36391 & 41165 & 13.119\% & 42161 & 15.856\% & 39752 & 9.236\% & 41149 & 13.075\% & 39887 & 9.606\% & 39175 & 7.650\% & \cellcolor{gray!30}38809 & \cellcolor{gray!30}6.645\% & 40381 & 10.964\% \\
			X-n449-k29 & 55233 & 60162 & 8.924\% & 60015 & 8.658\% & 60634 & 9.779\% & 61144 & 10.702\% & 60812 & 10.101\% & 59136 & 7.066\% & \cellcolor{gray!30}59033 & \cellcolor{gray!30}6.880\% & 60116 & 8.841\% \\
			X-n459-k26 & 24139 & 29543 & 22.387\% & 29100 & 20.552\% & 27347 & 13.290\% & 28267 & 17.101\% & 27435 & 13.654\% & \cellcolor{gray!30}26334 & \cellcolor{gray!30}9.093\% & 26554 & 10.005\% & 26712 & 10.659\% \\
			X-n469-k138 & 221824 & 252031 & 13.618\% & 245581 & 10.710\% & 238904 & 7.700\% & 237548 & 7.089\% & 243469 & 9.758\% & 238868 & 7.684\% & \cellcolor{gray!30}234044 & \cellcolor{gray!30}5.509\% & 237471 & 7.054\% \\
			X-n480-k70 & 89449 & 101314 & 13.265\% & 100121 & 11.931\% & 95032 & 6.242\% & 95466 & 6.727\% & 95167 & 6.392\% & 94341 & 5.469\% & \cellcolor{gray!30}93986 & \cellcolor{gray!30}5.072\% & 94618 & 5.779\% \\
			X-n491-k59 & 66483 & 77536 & 16.625\% & 75226 & 13.151\% & 72618 & 9.228\% & 71702 & 7.850\% & 72031 & 8.345\% & 70501 & 6.044\% & \cellcolor{gray!30}70411 & \cellcolor{gray!30}5.908\% & 71793 & 7.987\% \\
			\midrule
			\multicolumn{2}{c|}{Avg. Gap ($251 \leq n<501$)} & \multicolumn{2}{c}{11.529\%} & \multicolumn{2}{c}{10.616\%} & \multicolumn{2}{c}{8.107\%} & \multicolumn{2}{c}{8.889\%} & \multicolumn{2}{c}{\underline{8.237\%}} & \multicolumn{2}{c}{6.295\%} & \multicolumn{2}{c}{\cellcolor{gray!30}\underline{5.913\%}} & \multicolumn{2}{c}{7.079\%} \\
			\midrule
			X-n502-k39 & 69226 & 75711 & 9.368\% & 77033 & 11.278\% & 71908 & 3.874\% & 72655 & 4.953\% & 72108 & 4.163\% & 71726 & 3.611\% & \cellcolor{gray!30}71271 & \cellcolor{gray!30}2.954\% & 71599 & 3.428\% \\
			X-n513-k21 & 24201 & 34910 & 44.250\% & 32858 & 35.771\% & 28542 & 17.937\% & 29422 & 21.573\% & 28844 & 19.185\% & \cellcolor{gray!30}26767 & \cellcolor{gray!30}10.603\% & 27202 & 12.400\% & 28136 & 16.260\% \\
			X-n524-k153 & 154593 & 176491 & 14.165\% & 171734 & 11.088\% & 174150 & 12.651\% & 168181 & 8.790\% & \cellcolor{gray!30}165183 & \cellcolor{gray!30}6.849\% & 168238 & 8.826\% & 166429 & 7.656\% & 172677 & 11.698\% \\
			X-n536-k96 & 94846 & 109897 & 15.869\% & 106031 & 11.793\% & 103242 & 8.852\% & 102355 & 7.917\% & 102551 & 8.123\% & 100967 & 6.454\% & \cellcolor{gray!30}100532 & \cellcolor{gray!30}5.995\% & 102177 & 7.729\% \\
			X-n548-k50 & 86700 & 110984 & 28.009\% & 104240 & 20.231\% & 100850 & 16.321\% & 102318 & 18.014\% & 101100 & 16.609\% & 98583 & 13.706\% & 99247 & 14.472\% & \cellcolor{gray!30}94022 & \cellcolor{gray!30}8.445\% \\
			X-n561-k42 & 42717 & 55936 & 30.946\% & 53110 & 24.330\% & 49133 & 15.020\% & 50287 & 17.721\% & 48537 & 13.624\% & 47301 & 10.731\% & \cellcolor{gray!30}47134 & \cellcolor{gray!30}10.340\% & 49390 & 15.621\% \\
			X-n573-k30 & 50673 & 60884 & 20.151\% & 62033 & 22.418\% & 56048 & 10.607\% & 55353 & 9.236\% & 54571 & 7.692\% & 53624 & 5.824\% & 53974 & 6.514\% & \cellcolor{gray!30}53807 & \cellcolor{gray!30}6.185\% \\
			X-n586-k159 & 190316 & 226245 & 18.879\% & 212545 & 11.680\% & 205654 & 8.059\% & 204649 & 7.531\% & 208415 & 9.509\% & 204726 & 7.572\% & \cellcolor{gray!30}201846 & \cellcolor{gray!30}6.058\% & 204828 & 7.625\% \\
			X-n599-k92 & 108451 & 131035 & 20.824\% & 126654 & 16.785\% & 116840 & 7.735\% & 117784 & 8.606\% & 117440 & 8.289\% & \cellcolor{gray!30}115132 & \cellcolor{gray!30}6.160\% & 115224 & 6.245\% & 116627 & 7.539\% \\
			X-n613-k62 & 59535 & 77555 & 30.268\% & 73633 & 23.680\% & 67545 & 13.454\% & 69069 & 16.014\% & 67296 & 13.035\% & \cellcolor{gray!30}65123 & \cellcolor{gray!30}9.386\% & 65851 & 10.609\% & 67346 & 13.120\% \\
			X-n627-k43 & 62164 & 76776 & 23.506\% & 70744 & 13.802\% & 67523 & 8.621\% & 69361 & 11.577\% & 69915 & 12.468\% & 67301 & 8.264\% & \cellcolor{gray!30}66603 & \cellcolor{gray!30}7.141\% & 77841 & 25.219\% \\
			X-n641-k35 & 63684 & 83138 & 30.548\% & 71986 & 13.036\% & 70631 & 10.909\% & 73624 & 15.608\% & 72517 & 13.870\% & \cellcolor{gray!30}69308 & \cellcolor{gray!30}8.831\% & 69601 & 9.291\% & 70952 & 11.413\% \\
			X-n655-k131 & 106780 & 120771 & 13.103\% & 118758 & 11.217\% & 112289 & 5.159\% & 110657 & 3.631\% & 112598 & 5.449\% & 109726 & 2.759\% & 109505 & 2.552\% & \cellcolor{gray!30}109546 & \cellcolor{gray!30}2.590\% \\
			X-n670-k130 & 146332 & 183183 & 25.183\% & 168210 & 14.951\% & 168829 & 15.374\% & 161571 & 10.414\% & 160473 & 9.664\% & 161354 & 10.266\% & \cellcolor{gray!30}158554 & \cellcolor{gray!30}8.352\% & 164643 & 12.513\% \\
			X-n685-k75 & 68205 & 92701 & 35.915\% & 82607 & 21.116\% & 77890 & 14.200\% & 78473 & 15.055\% & 76817 & 12.626\% & 74689 & 9.507\% & \cellcolor{gray!30}74523 & \cellcolor{gray!30}9.263\% & 77686 & 13.901\% \\
			X-n701-k44 & 81923 & 92723 & 13.183\% & 89704 & 9.498\% & 90580 & 10.567\% & 92198 & 12.542\% & 91600 & 11.812\% & 88246 & 7.718\% & \cellcolor{gray!30}87910 & \cellcolor{gray!30}7.308\% & 91921 & 12.204\% \\
			X-n716-k35 & 43373 & 59383 & 36.912\% & 52170 & 20.282\% & 49480 & 14.080\% & 50605 & 16.674\% & 50073 & 15.446\% & 47683 & 9.937\% & \cellcolor{gray!30}47292 & \cellcolor{gray!30}9.036\% & 51944 & 19.761\% \\
			X-n733-k159 & 136187 & 175848 & 29.122\% & 156268 & 14.745\% & 148581 & 9.101\% & 146080 & 7.264\% & 146443 & 7.531\% & 144836 & 6.351\% & \cellcolor{gray!30}142941 & \cellcolor{gray!30}4.959\% & 145364 & 6.739\% \\
			X-n749-k98 & 77269 & 102208 & 32.276\% & 92403 & 19.586\% & 85046 & 10.065\% & 85325 & 10.426\% & 85320 & 10.419\% & \cellcolor{gray!30}82650 & \cellcolor{gray!30}6.964\% & 83014 & 7.435\% & 84975 & 9.973\% \\
			X-n766-k71 & 114417 & 132968 & 16.213\% & 130101 & 13.708\% & 129866 & 13.502\% & 127752 & 11.655\% & 126515 & 10.573\% & 124163 & 8.518\% & \cellcolor{gray!30}123989 & \cellcolor{gray!30}8.366\% & 130645 & 14.183\% \\
			X-n783-k48 & 72386 & 108577 & 49.997\% & 96432 & 33.219\% & 82839 & 14.441\% & 87562 & 20.965\% & 87027 & 20.226\% & \cellcolor{gray!30}80255 & \cellcolor{gray!30}10.871\% & 80501 & 11.211\% & 87629 & 21.058\% \\
			X-n801-k40 & 73311 & 92125 & 25.663\% & 87187 & 18.928\% & 86121 & 17.474\% & 94076 & 28.325\% & 90039 & 22.818\% & 84979 & 15.916\% & 84397 & 15.122\% & \cellcolor{gray!30}80746 & \cellcolor{gray!30}10.142\% \\
			X-n819-k171 & 158121 & 192102 & 21.491\% & 178856 & 13.113\% & 174446 & 10.324\% & 172387 & 9.022\% & 172578 & 9.143\% & 170347 & 7.732\% & \cellcolor{gray!30}168486 & \cellcolor{gray!30}6.555\% & 170573 & 7.875\% \\
			X-n837-k142 & 193737 & 231002 & 19.235\% & 230226 & 18.834\% & 208669 & 7.707\% & 209540 & 8.157\% & 208974 & 7.865\% & 205416 & 6.028\% & \cellcolor{gray!30}205231 & \cellcolor{gray!30}5.933\% & 206980 & 6.836\% \\
			X-n856-k95 & 88965 & 117243 & 31.786\% & 105763 & 18.882\% & 98164 & 10.340\% & 102312 & 15.003\% & 100818 & 13.324\% & \cellcolor{gray!30}96653 & \cellcolor{gray!30}8.642\% & 98082 & 10.248\% & 100106 & 12.523\% \\
			X-n876-k59 & 99299 & 114212 & 15.018\% & 114175 & 14.981\% & 107477 & 8.236\% & 109693 & 10.467\% & 108742 & 9.509\% & 105570 & 6.315\% & \cellcolor{gray!30}105116 & \cellcolor{gray!30}5.858\% & 113121 & 13.920\% \\
			X-n895-k37 & 53860 & 106062 & 96.922\% & 70363 & 30.641\% & 64225 & 19.244\% & 73280 & 36.056\% & 70673 & 31.216\% & 62387 & 15.832\% & \cellcolor{gray!30}61973 & \cellcolor{gray!30}15.063\% & 74729 & 38.747\% \\
			X-n916-k207 & 329179 & 387367 & 17.677\% & 374899 & 13.889\% & 353039 & 7.248\% & 351887 & 6.898\% & 358336 & 8.857\% & 350863 & 6.587\% & \cellcolor{gray!30}347313 & \cellcolor{gray!30}5.509\% & 352460 & 7.072\% \\
			X-n936-k151 & 132715 & 200816 & 51.314\% & 161700 & 21.840\% & 162903 & 22.746\% & 154847 & 16.676\% & 153747 & 15.847\% & 151238 & 13.957\% & \cellcolor{gray!30}145801 & \cellcolor{gray!30}9.860\% & 153139 & 15.389\% \\
			X-n957-k87 & 85465 & 126220 & 47.686\% & 124190 & 45.311\% & \cellcolor{gray!30}103089 & \cellcolor{gray!30}20.621\% & 108664 & 27.144\% & 105588 & 23.522\% & 100635 & 17.750\% & 101187 & 18.396\% & 110763 & 29.600\% \\
			X-n979-k58 & 118976 & 138987 & 16.819\% & 132651 & 11.494\% & 129633 & 8.957\% & 133201 & 11.956\% & 137728 & 15.761\% & 127697 & 7.330\% & \cellcolor{gray!30}126903 & \cellcolor{gray!30}6.663\% & 134360 & 12.930\% \\
			X-n1001-k43 & 72355 & 132976 & 83.783\% & 89175 & 23.246\% & 85852 & 18.654\% & 92974 & 28.497\% & 90653 & 25.287\% & 82613 & 14.177\% & \cellcolor{gray!30}82012 & \cellcolor{gray!30}13.347\% & 95172 & 31.535\% \\
			\midrule
			\multicolumn{2}{c|}{Avg. Gap ($501 < n \leq 1001$)} & \multicolumn{2}{c}{30.190\%} & \multicolumn{2}{c}{18.918\%} & \multicolumn{2}{c}{12.253\%} & \multicolumn{2}{c}{14.199\%} & \multicolumn{2}{c}{\underline{13.344\%}} & \multicolumn{2}{c}{9.191\%} & \multicolumn{2}{c}{\cellcolor{gray!30}\underline{8.571\%}} & \multicolumn{2}{c}{13.555\%} \\
			\midrule
			\multicolumn{2}{c|}{Avg. Gap} & \multicolumn{2}{c}{15.863\%} & \multicolumn{2}{c}{11.693\%} & \multicolumn{2}{c}{8.428\%} & \multicolumn{2}{c}{9.230\%} & \multicolumn{2}{c}{\underline{8.616\%}} & \multicolumn{2}{c}{6.586\%} & \multicolumn{2}{c}{\cellcolor{gray!30}\underline{6.074\%}} & \multicolumn{2}{c}{8.322\%} \\
			\bottomrule
	\end{tabular}}
\end{table*}

\subsection{Results on benchmark datasets}
\label{appendix_cvrplib}

We also evaluate the performance of previous models on benchmark datasets. For the MDVRP, we generated new datasets derived from the 10 instances of \citet{Cordeau1997}. While often cited as MDVRP problems, these instances originally imposed route length limits (MDVRPL). We utilized their original node coordinates and demands to create 10 corresponding instances for each of the 23 remaining variants, totaling 240 benchmark instances. Missing attributes were generated following Section~\ref{instance_gen}: 20\% of customers were randomly assigned as backhauls, and time windows were generated using the standard protocol. All models were evaluated using a greedy decoding with $\times$8 instance augmentation. For each instance, we ran HGS-PyVRP with a 5-minute runtime limit. Table~\ref{mdvrplib} presents the results, averaged across the 10 instances for each variant. Once again, both FiLMMeD models showed a significantly better performance than their respective baselines across most variants.

For single-depot VRPs, we utilized the models from Table~\ref{vrp_berto}, that is, FiLMMeD-CaDA and FiLMMeD-CaDA†. Here, we considered the Set-X benchmark \citep{Uchoa2017} from CVRPLib, which contains instances ranging from 100 to 1000 nodes. Similar to the previous MDVRP experiments, a greedy decoding was used with $\times$8 instance augmentation. The results, summarized in Table~\ref{setX_results}, show once again that FiLMMeD variations outperform their non-FiLMMeD counterparts across tha majority of instances. Furthermore, FiLMMeD-CaDA† displays the best average performance, surpassing even the Dual-Branch Distance-Aware (DBDA) method from \citet{Jiang2026}, which was specifically designed for cross-distribution generalization. These experiments indicate that our method has great cross-size and cross-distribution capabilities, despite being primarily designed for cross-problem generalization.

\subsection{Results on unseen tasks}

\begin{table*}[pos=!h]
	\centering
	\caption{Zero-shot generalization performance on unseen VRP variants. The best MTL results (among neural models) are shown in gray, and the models improved by our approach are underlined.}
	\label{tab_zeroshot}
	\resizebox{0.95\textwidth}{!}{%
		\large
		\renewcommand{\arraystretch}{1.05}
		\begin{tabular}{l|cc|cc|cc|cc|cc|cc|cc|cc|cc}
			\toprule
			& \multicolumn{2}{c|}{HGS-PyVRP} & \multicolumn{2}{c|}{OR-Tools} & \multicolumn{2}{c|}{MTPOMO} & \multicolumn{2}{c|}{MVMoE} & \multicolumn{2}{c|}{RouteFinder} & \multicolumn{2}{c|}{CaDA} & \multicolumn{2}{c|}{FiLMMeD-CaDA} & \multicolumn{2}{c|}{CaDA†} & \multicolumn{2}{c}{FiLMMeD-CaDA†} \\
			\cmidrule(lr){2-3} \cmidrule(lr){4-5} \cmidrule(lr){6-7} \cmidrule(lr){8-9} \cmidrule(lr){10-11} \cmidrule(lr){12-13} \cmidrule(lr){14-15} \cmidrule(lr){16-17} \cmidrule(lr){18-19}
			Variant & Obj. & Gap & Obj. & Gap & Obj. & Gap & Obj. & Gap & Obj. & Gap & Obj. & Gap & Obj. & Gap & Obj. & Gap & Obj. & Gap \\
			\midrule
			VRPMB & 13.54 & * & 14.93 & 10.27\% & 15.04 & 11.32\% & 14.99 & 10.94\% & 14.88 & 10.13\% & 14.83 & 9.73\% & \underline{14.65} & \underline{8.37\%} & 14.62 & 8.15\% & \cellcolor{gray!25}\underline{14.60} & \cellcolor{gray!25}\underline{7.99\%} \\
			OVRPMB & 9.01 & * & 10.59 & 17.54\% & 10.87 & 20.65\% & 10.85 & 20.42\% & 10.72 & 19.02\% & 10.58 & 17.38\% & \underline{10.51} & \underline{16.71\%} & 10.47 & 16.24\% & \cellcolor{gray!25}\underline{10.39} & \cellcolor{gray!25}\underline{15.34\%} \\
			VRPMBL & 13.78 & * & 15.42 & 11.90\% & 15.41 & 11.97\% & 15.33 & 11.37\% & 15.18 & 10.32\% & 15.05 & 9.27\% & \underline{14.97} & \underline{8.71\%} & \cellcolor{gray!25}14.93 & \cellcolor{gray!25}8.40\% & 14.95 & 8.53\% \\
			VRPMBTW & 25.51 & * & 29.97 & 17.48\% & 28.31 & 11.06\% & 28.32 & 11.10\% & 28.29 & 10.87\% & 28.40 & 11.42\% & \underline{28.28} & \underline{10.92\%} & 28.20 & 10.62\% & \cellcolor{gray!25}\underline{28.16} & \cellcolor{gray!25}\underline{10.45\%} \\
			OVRPMBL & 9.01 & * & 10.59 & 17.54\% & 10.85 & 20.43\% & 10.82 & 20.14\% & 10.72 & 19.01\% & 10.58 & 17.41\% & \underline{10.51} & \underline{16.72\%} & 10.47 & 16.24\% & \cellcolor{gray!25}\underline{10.39} & \cellcolor{gray!25}\underline{15.37\%} \\
			OVRPMBTW & 16.97 & * & 19.31 & 13.78\% & 18.51 & 9.08\% & 18.55 & 9.33\% & 18.45 & 8.68\% & 18.57 & 9.44\% & \underline{18.56} & \underline{9.38\%} & \cellcolor{gray!25}18.40 & \cellcolor{gray!25}8.42\% & 18.48 & 8.90\% \\
			VRPMBLTW & 25.85 & * & 30.44 & 17.76\% & 28.73 & 11.27\% & 28.70 & 11.16\% & 28.65 & 10.82\% & 28.75 & 11.38\% & \underline{28.62} & \underline{10.86\%} & 28.51 & 10.43\% & \cellcolor{gray!25}\underline{28.48} & \cellcolor{gray!25}\underline{10.32\%} \\
			OVRPMBLTW & 16.97 & * & 19.31 & 13.78\% & 18.51 & 9.12\% & 18.55 & 9.30\% & 18.45 & 8.69\% & 18.57 & 9.43\% & \underline{18.55} & \underline{9.32\%} & \cellcolor{gray!25}18.40 & \cellcolor{gray!25}8.41\% & 18.48 & 8.91\% \\
			MDVRP & 11.89 & * & 12.52 & 5.27\% & 16.07 & 35.74\% & 16.02 & 35.35\% & 15.98 & 35.02\% & 17.16 & 45.33\% & \underline{14.30} & \underline{20.69\%} & \cellcolor{gray!25}13.99 & \cellcolor{gray!25}18.06\% & 14.03 & 18.43\% \\
			MDOVRP & 7.97 & * & 8.16 & 2.33\% & 10.28 & 29.06\% & 10.24 & 28.59\% & 10.18 & 27.82\% & \cellcolor{gray!25}9.74 & \cellcolor{gray!25}22.31\% & 9.84 & 23.70\% & 10.52 & 32.18\% & \underline{10.25} & \underline{28.73\%} \\
			MDVRPB & 11.64 & * & 12.22 & 5.01\% & 15.18 & 30.66\% & 15.12 & 30.13\% & 15.05 & 29.53\% & 17.86 & 54.31\% & \underline{14.44} & \underline{24.38\%} & \cellcolor{gray!25}13.99 & \cellcolor{gray!25}20.48\% & 14.09 & 21.34\% \\
			MDVRPL & 11.90 & * & 12.52 & 5.24\% & 16.30 & 37.58\% & 16.25 & 37.17\% & 16.20 & 36.76\% & 17.14 & 45.04\% & \underline{14.37} & \underline{21.27\%} & \cellcolor{gray!25}14.07 & \cellcolor{gray!25}18.64\% & 14.08 & 18.70\% \\
			MDVRPTW & 19.33 & * & 19.62 & 1.55\% & 26.68 & 38.56\% & 26.67 & 38.51\% & 26.51 & 37.64\% & 27.18 & 41.22\% & \underline{24.12} & \underline{25.25\%} & 23.28 & 20.75\% & \cellcolor{gray!25}\underline{22.83} & \cellcolor{gray!25}\underline{18.38\%} \\
			MDOVRPTW & 13.00 & * & 13.09 & 0.74\% & 17.57 & 35.67\% & 17.57 & 35.68\% & 17.48 & 34.96\% & 17.23 & 33.02\% & \underline{15.93} & \underline{23.03\%} & 15.96 & 23.17\% & \cellcolor{gray!25}\underline{15.14} & \cellcolor{gray!25}\underline{16.76\%} \\
			MDOVRPB & 8.69 & * & 8.87 & 2.15\% & 10.94 & 26.08\% & 10.89 & 25.56\% & 10.82 & 24.74\% & 10.93 & 26.06\% & \cellcolor{gray!25}\underline{10.53} & \cellcolor{gray!25}\underline{21.43\%} & 11.21 & 29.30\% & \underline{11.01} & \underline{26.93\%} \\
			MDOVRPL & 7.97 & * & 8.16 & 2.33\% & 10.28 & 29.07\% & 10.24 & 28.60\% & 10.18 & 27.84\% & \cellcolor{gray!25}9.74 & \cellcolor{gray!25}22.26\% & 9.93 & 24.73\% & 10.51 & 32.01\% & \underline{10.24} & \underline{28.63\%} \\
			MDVRPBL & 11.68 & * & 12.22 & 4.66\% & 15.80 & 35.54\% & 15.73 & 34.95\% & 15.62 & 33.98\% & 17.50 & 50.52\% & \underline{14.47} & \underline{24.15\%} & \cellcolor{gray!25}14.21 & \cellcolor{gray!25}21.86\% & 14.22 & 21.96\% \\
			MDVRPBTW & 22.03 & * & 22.40 & 1.69\% & 30.55 & 39.23\% & 30.55 & 39.22\% & 30.36 & 38.36\% & 30.88 & 40.84\% & \underline{27.54} & \underline{25.59\%} & 26.60 & 21.12\% & \cellcolor{gray!25}\underline{25.68} & \cellcolor{gray!25}\underline{16.85\%} \\
			MDVRPLTW & 19.35 & * & 19.66 & 1.58\% & 27.13 & 40.71\% & 27.12 & 40.67\% & 26.93 & 39.69\% & 27.63 & 43.44\% & \underline{24.24} & \underline{25.73\%} & 23.32 & 20.82\% & \cellcolor{gray!25}\underline{22.86} & \cellcolor{gray!25}\underline{18.38\%} \\
			MDOVRPBL & 8.69 & * & 8.87 & 2.13\% & 10.94 & 26.11\% & 10.90 & 25.66\% & 10.82 & 24.78\% & 10.93 & 25.99\% & \cellcolor{gray!25}\underline{10.65} & \cellcolor{gray!25}\underline{22.85\%} & 11.20 & 29.13\% & \underline{11.01} & \underline{26.90\%} \\
			MDOVRPBTW & 14.37 & * & 14.49 & 0.87\% & 19.69 & 37.62\% & 19.69 & 37.62\% & 19.59 & 36.95\% & 19.38 & 35.44\% & \underline{17.75} & \underline{24.09\%} & 17.77 & 24.10\% & \cellcolor{gray!25}\underline{16.59} & \cellcolor{gray!25}\underline{15.71\%} \\
			MDOVRPLTW & 13.00 & * & 13.09 & 0.70\% & 17.58 & 35.70\% & 17.58 & 35.74\% & 17.48 & 34.95\% & 17.22 & 32.99\% & \underline{15.93} & \underline{22.98\%} & 15.97 & 23.27\% & \cellcolor{gray!25}\underline{15.14} & \cellcolor{gray!25}\underline{16.76\%} \\
			MDVRPBLTW & 22.06 & * & 22.43 & 1.70\% & 31.09 & 41.52\% & 31.06 & 41.39\% & 30.86 & 40.49\% & 31.57 & 43.84\% & \underline{27.62} & \underline{25.79\%} & 26.62 & 21.04\% & \cellcolor{gray!25}\underline{25.75} & \cellcolor{gray!25}\underline{17.04\%} \\
			MDOVRPBLTW & 14.37 & * & 14.49 & 0.86\% & 19.69 & 37.64\% & 19.69 & 37.61\% & 19.60 & 36.96\% & 19.38 & 35.44\% & \underline{17.78} & \underline{24.23\%} & 17.77 & 24.08\% & \cellcolor{gray!25}\underline{16.59} & \cellcolor{gray!25}\underline{15.72\%} \\
			MDVRPMB & 10.68 & * & 12.22 & 14.37\% & 15.14 & 42.22\% & 15.08 & 41.67\% & 14.99 & 40.80\% & 17.51 & 65.10\% & \underline{14.24} & \underline{33.86\%} & \cellcolor{gray!25}13.78 & \cellcolor{gray!25}29.41\% & 13.89 & 30.52\% \\
			MDOVRPMB & 7.66 & * & 8.88 & 15.83\% & 10.91 & 42.57\% & 10.90 & 42.41\% & 10.77 & 40.67\% & 10.78 & 40.93\% & \cellcolor{gray!25}\underline{10.42} & \cellcolor{gray!25}\underline{36.20\%} & 10.97 & 43.35\% & \underline{10.78} & \underline{40.90\%} \\
			MDVRPMBL & 10.71 & * & 12.23 & 14.23\% & 15.49 & 45.23\% & 15.40 & 44.37\% & 15.28 & 43.27\% & 16.86 & 58.45\% & \underline{14.17} & \underline{32.82\%} & \cellcolor{gray!25}13.85 & \cellcolor{gray!25}29.77\% & 13.96 & 30.83\% \\
			MDVRPMBTW & 19.29 & * & 22.39 & 16.12\% & 28.44 & 48.01\% & 28.46 & 48.12\% & 28.43 & 47.93\% & 29.38 & 52.98\% & \underline{26.24} & \underline{36.64\%} & 24.69 & 28.37\% & \cellcolor{gray!25}\underline{24.08} & \cellcolor{gray!25}\underline{25.14\%} \\
			MDOVRPMBL & 7.66 & * & 8.87 & 15.73\% & 10.90 & 42.45\% & 10.88 & 42.13\% & 10.76 & 40.62\% & 10.78 & 40.91\% & \cellcolor{gray!25}\underline{10.52} & \cellcolor{gray!25}\underline{37.47\%} & 10.97 & 43.42\% & \underline{10.78} & \underline{40.83\%} \\
			MDOVRPMBTW & 12.96 & * & 14.49 & 11.79\% & 18.56 & 43.63\% & 18.61 & 44.04\% & 18.49 & 43.14\% & 18.70 & 44.76\% & \underline{17.19} & \underline{33.09\%} & 16.66 & 28.89\% & \cellcolor{gray!25}\underline{15.86} & \cellcolor{gray!25}\underline{22.62\%} \\
			MDVRPMBLTW & 19.31 & * & 22.43 & 16.16\% & 28.93 & 50.36\% & 28.89 & 50.19\% & 28.80 & 49.69\% & 29.95 & 55.78\% & \underline{26.26} & \underline{36.57\%} & 24.64 & 27.92\% & \cellcolor{gray!25}\underline{24.02} & \cellcolor{gray!25}\underline{24.68\%} \\
			MDOVRPMBLTW & 12.96 & * & 14.49 & 11.79\% & 18.56 & 43.65\% & 18.60 & 43.95\% & 18.50 & 43.17\% & 18.70 & 44.78\% & \underline{17.20} & \underline{33.16\%} & 16.66 & 28.89\% & \cellcolor{gray!25}\underline{15.86} & \cellcolor{gray!25}\underline{22.55\%} \\
			\midrule
			Avg. Gap & \multicolumn{2}{c|}{*} & \multicolumn{2}{c|}{9.40\%} & \multicolumn{2}{c|}{32.18\%} & \multicolumn{2}{c|}{31.97\%} & \multicolumn{2}{c|}{31.49\%} & \multicolumn{2}{c|}{36.36\%} & \multicolumn{2}{c|}{\underline{23.46\%}} & \multicolumn{2}{c|}{23.41\%} & \multicolumn{2}{c}{\cellcolor{gray!25}\underline{21.01\%}} \\
			\bottomrule
		\end{tabular}
	}
\end{table*}

To evaluate the zero-shot generalization capabilities of FiLMMeD-CaDA and FiLMMeD-CaDA†, we conducted further experiments across 32 unseen variants. Following the experimental settings of \citet{Berto2025} and \citet{Li2025}, these evaluations were performed on 100-node instances. The test set includes mixed backhauls (MB) and multi-depots (MD), variants that appeared in the settings of \citet{Zhou2024}, but remain unseen to the training protocol established by \citet{Berto2025}. The results are summarized in Table~\ref{tab_zeroshot}. Overall, FiLMMeD-CaDA and FiLMMeD-CaDA† consistently improved compared to their counterparts, obtaining a lower average gap. Furthermore, FiLMMeD-CaDA† displayed the best performance across all evaluated neural baselines, consolidating the zero-shot generalization of our method.

\subsection{Performance of the PO algorithm on other models}
\label{po_algo_performance}

To further evaluate the effectiveness of the PO algorithm in the MTL domain, we retrained MTPOMO, MVMoE and RouteFinder (once again, the modern Transformer-based Encoder variant of RouteFinder) using PO and RL. The RL-based models were trained via the standard REINFORCE with shared baselines algorithm. Following the experimental protocol of \citet{Berto2025}, the average performance gaps are reported in Table~\ref{table_po_ablation}. Across all three baselines, the PO algorithm exhibited superior generalization, underscoring even more its viability for future MTL VRP research. Moreover, its consistent performance improvements across different model architectures display great modularity, suggesting the application of PO for MTL is architecture-agnostic rather than tied to a specific model.

\begin{table}[pos=!h]
	\centering
	\caption{Average gaps across 16 VRP variants: RL vs. PO comparison.}
	\label{table_po_ablation}
	\begin{tabular}{lcc}
		\toprule
		Model & RL & PO \\
		\midrule
		MTPOMO & 2.354\% & \textbf{2.141\%} \\
		MVMoE & 2.169\% & \textbf{1.938\%} \\
		RouteFinder & 2.049\% & \textbf{1.807\%} \\
		\bottomrule
	\end{tabular}
\end{table}

\subsection{Complexity analysis}
\label{sec_computational_complexity}

In this section, we report the computational overhead of all FiLMMeD variations. The results, reported in Table~\ref{tab_computational_complexity}, show that the FiLM mechanism adds fewer than 0.01M parameters, an increase of less than 1\%. Similarly, the combined GPU memory overhead introduced by FiLM and PO remains negligible across all configurations.

Additionally, the training times per epoch are nearly identical between FiLMMeD and its counterparts, indicating that PO introduces minimal overhead compared to REINFORCE despite its $O(N^2)$ complexity in pairwise comparisons. This efficiency stems from the fact that the preference computation is lightweight relative to the autoregressive decoder's forward pass.

\begin{table}[pos=!h]
\centering
\caption{Computational complexity analysis.}
\label{tab_computational_complexity}
\resizebox{\columnwidth}{!}{%
	\scriptsize
	\renewcommand{\arraystretch}{1.05}
	\begin{tabular}{l | c c c}
		\toprule
		Model & Params (M) & \begin{tabular}[c]{@{}c@{}}Avg. GPU\\Memory (GB)\end{tabular} & \begin{tabular}[c]{@{}c@{}}Avg. Training\\Epoch Time\end{tabular} \\
		\midrule
		MTPOMO & 1.25 & 6.55 & 2m 5s \\
		FiLMMeD-MTPOMO & 1.26 & 6.81 & 2m 9s \\
		\midrule
		MVMoE & 3.68 & 9.00 & 3m 20s \\
		FiLMMeD-MVMoE & 3.68 & 9.49 & 3m 21s \\
		\midrule
		CaDA & 3.43 & 5.47 & 2m 50s \\
		FiLMMeD-CaDA & 3.43 & 5.97 & 2m 49s \\
		CaDA† & 3.56 & 6.92 & 3m 34s \\
		FiLMMeD-CaDA† & 3.56 & 7.74 & 3m 38s \\
		\bottomrule
	\end{tabular}
}
\end{table}

\section{Conclusion}
\label{conclusion}

This paper introduces FiLMMeD, a novel MTL model focused on solving MDVRP variants. FiLMMeD augments the standard encoder architecture used in MTL works with a FiLM-based conditioning mechanism that adapts the model’s internal representations to the specific set of active constraints in each instance. We demonstrate that this mechanism effectively learns a more discriminative latent space of node embeddings, leading to a better generalization. We also propose a CL training strategy that progressively exposes the model to increasingly more complex MDVRP variants, based on the number of constraints. Compared to established training strategies and sampling regimens, our curriculum yielded substantially improved generalization across the 24 tested MDVRP variants. Finally, we explored the use of PO as an alternative to RL to train FiLMMeD on single-depot variants. Across different experimental settings, our results show that PO provides a faster convergence and better generalization, offering an early but compelling signal that PO may be the preferred training paradigm for future MTL routing works. We have confirmed the effectiveness of FiLMMeD through extensive experiments on both multi- and single-depot problems. Additional ablation studies confirm the isolated impact of each individual contribution. For future work, we envision augmenting the PO algorithm to further improve the generalization of existing methods in the MTL setting.

\printcredits

\section*{Acknowledgements}

\begin{sloppypar}
This research is sponsored by national funds through FCT - Fundação para a Ciência e a Tecnologia (FCT), Portugal, through doctoral grant 2025.00622.BD and projects UID/00285/2025, and LA/P/0112/2020. This work was also supported by the European Union - NextGenerationEU, through the Portuguese Republic's Recovery and Resilience Plan Partnership Agreement [project C645808870-00000067], within the scope of the project PRODUTECH R3 – ‘‘Agenda Mobilizadora da Fileira das Tecnologias de Produção para a Reindustrialização’’, Total project investment: 166.988.013,71 Euros; Total Grant : 97.111.730,27 Euros; and by the European Regional Development Fund (ERDF) through the Operational Program for Competitiveness and Internationalization (COMPETE 2020) under the project POCI-01-0247-FEDER-046102 (PRODUTECH4S\&C).
\end{sloppypar}

\bibliographystyle{cas-model2-names}

\bibliography{cas-refs}

@article{Wang2019,
   author = {Zheng Wang and Jiuh-Biing Sheu},
   doi = {10.1016/j.trb.2019.03.005},
   issn = {0191-2615},
   journal = {Transportation Research Part B: Methodological},
   pages = {350-364},
   title = {Vehicle routing problem with drones},
   volume = {122},
   url = {https://www.sciencedirect.com/science/article/pii/S0191261518307884},
   year = {2019}
}

@article{Cattaruzza2017,
   author = {Diego Cattaruzza and Nabil Absi and Dominique Feillet and Jesús González-Feliu},
   doi = {10.1007/s13676-014-0074-0},
   issn = {21924384},
   issue = {1},
   journal = {EURO Journal on Transportation and Logistics},
   month = {3},
   pages = {51-79},
   publisher = {Springer Verlag},
   title = {Vehicle routing problems for city logistics},
   volume = {6},
   year = {2017}
}

@article{Li2019,
   author = {Yantong Li and Feng Chu and Chenpeng Feng and Chengbin Chu and Meng Chu Zhou},
   doi = {10.1109/TITS.2018.2835145},
   issn = {15249050},
   issue = {3},
   journal = {IEEE Transactions on Intelligent Transportation Systems},
   month = {3},
   pages = {867-878},
   publisher = {Institute of Electrical and Electronics Engineers Inc.},
   title = {Integrated Production Inventory Routing Planning for Intelligent Food Logistics Systems},
   volume = {20},
   year = {2019}
}

@article{Pessoa2020,
   author = {Artur Pessoa and Ruslan Sadykov and Eduardo Uchoa and François Vanderbeck},
   doi = {10.1007/s10107-020-01523-z},
   issn = {14364646},
   issue = {1-2},
   journal = {Mathematical Programming},
   month = {9},
   pages = {483-523},
   publisher = {Springer},
   title = {A generic exact solver for vehicle routing and related problems},
   volume = {183},
   year = {2020}
}

@article{Bogyrbayeva2024,
author = {Bogyrbayeva, Aigerim and Meraliyev, Meraryslan and Mustakhov, Taukekhan and Dauletbayev, Bissenbay},
title = {Machine Learning to Solve Vehicle Routing Problems: A Survey},
year = {2024},
issue_date = {June 2024},
publisher = {IEEE Press},
volume = {25},
number = {6},
issn = {1524-9050},
url = {https://doi.org/10.1109/TITS.2023.3334976},
doi = {10.1109/TITS.2023.3334976},
journal = {Trans. Intell. Transport. Sys.},
month = jun,
pages = {4754–4772},
numpages = {19}
}

@article{Bengio2021,
   author = {Yoshua Bengio and Andrea Lodi and Antoine Prouvost},
   doi = {10.1016/j.ejor.2020.07.063},
   issn = {0377-2217},
   issue = {2},
   journal = {European Journal of Operational Research},
   pages = {405-421},
   title = {Machine learning for combinatorial optimization: A methodological tour d’horizon},
   volume = {290},
   url = {https://www.sciencedirect.com/science/article/pii/S0377221720306895},
   year = {2021}
}

@inproceedings{Kool2019,
   author = {Wouter Kool and Herke van Hoof and Max Welling},
   booktitle = {7th International Conference on Learning Representations, ICLR 2019},
   title = {Attention, learn to solve routing problems!},
   year = {2019}
}

@inproceedings{Kwon2020,
   author = {Yeong-Dae Kwon and Jinho Choo and Byoungjip Kim and Iljoo Yoon and Youngjune Gwon and Seungjai Min},
   booktitle = {Advances in Neural Information Processing Systems},
   title = {POMO: Policy optimization with multiple optima for reinforcement learning},
   year = {2020}
}

@inproceedings{Zhou2023,
   author = {Jianan Zhou and Yaoxin Wu and Wen Song and Zhiguang Cao and Jie Zhang},
   booktitle = {Proceedings of Machine Learning Research},
   pages = {42769-42789},
   title = {Towards Omni-generalizable Neural Methods for Vehicle Routing Problems},
   year = {2023}
}

@article{Elatar2023,
   author = {Said Elatar and Karim Abouelmehdi and Mohammed Essaid Riffi},
   doi = {10.1016/j.procs.2023.03.051},
   issn = {1877-0509},
   journal = {Procedia Computer Science},
   note = {The 14th International Conference on Ambient Systems, Networks and Technologies Networks (ANT) and The 6th International Conference on Emerging Data and Industry 4.0 (EDI40)},
   pages = {398-404},
   title = {The vehicle routing problem in the last decade: variants, taxonomy and metaheuristics},
   volume = {220},
   url = {https://www.sciencedirect.com/science/article/pii/S1877050923005860},
   year = {2023}
}

@article{Zhou2024,
   author = {Jianan Zhou and Zhiguang Cao and Yaoxin Wu and Wen Song and Yining Ma and Jie Zhang and Chi Xu},
   journal = {Proceedings of Machine Learning Research},
   pages = {61804-61824},
   title = {MVMoE: Multi-Task Vehicle Routing Solver with Mixture-of-Experts},
   year = {2024}
}

@inproceedings{Liu2024,
   author = {Fei Liu and Xi Lin and Z. Wang and Q. Zhang and T. Xialiang and M. Yuan},
   doi = {10.1145/3637528.3672040},
   isbn = {9798400704901},
   booktitle = {Proceedings of the ACM SIGKDD International Conference on Knowledge Discovery and Data Mining},
   pages = {1898-1908},
   title = {Multi-Task Learning for Routing Problem with Cross-Problem Zero-Shot Generalization},
   year = {2024}
}

@article{
Berto2025,
title={{RouteFinder: Towards Foundation Models for Vehicle Routing Problems}},
author={Federico Berto and Chuanbo Hua and Nayeli Gast Zepeda and Andr{\'e} Hottung and Niels Wouda and Leon Lan and Junyoung Park and Kevin Tierney and Jinkyoo Park},
journal={Transactions on Machine Learning Research},
issn={2835-8856},
year={2025},
url={https://openreview.net/forum?id=QzGLoaOPiY},
}

@inproceedings{Perez2018,
author = {Perez, Ethan and Strub, Florian and de Vries, Harm and Dumoulin, Vincent and Courville, Aaron},
title = {FiLM: visual reasoning with a general conditioning layer},
year = {2018},
isbn = {978-1-57735-800-8},
publisher = {AAAI Press},
booktitle = {Proceedings of the Thirty-Second AAAI Conference on Artificial Intelligence and Thirtieth Innovative Applications of Artificial Intelligence Conference and Eighth AAAI Symposium on Educational Advances in Artificial Intelligence},
articleno = {483},
numpages = {10},
location = {New Orleans, Louisiana, USA},
series = {AAAI'18/IAAI'18/EAAI'18}
}

@INPROCEEDINGS{Lyu2023,
  author={Lyu, Wenjun and Wang, Haotian and Hong, Zhiqing and Wang, Guang and Yang, Yu and Liu, Yunhuai and Zhang, Desheng},
  booktitle={2023 IEEE 39th International Conference on Data Engineering (ICDE)}, 
  title={REDE: Exploring Relay Transportation for Efficient Last-mile Delivery}, 
  year={2023},
  volume={},
  number={},
  pages={3003-3016},
  doi={10.1109/ICDE55515.2023.00230}}

@inproceedings{
Li2025,
title={Ca{DA}: Cross-Problem Routing Solver with Constraint-Aware Dual-Attention},
author={Han Li and Fei Liu and Zhi Zheng and Yu Zhang and Zhenkun Wang},
booktitle={Forty-second International Conference on Machine Learning},
year={2025},
url={https://openreview.net/forum?id=CS4RyQuTig}
}

@misc{Mazyavkina2021,
   author = {Nina Mazyavkina and Sergey Sviridov and Sergei Ivanov and Evgeny Burnaev},
   doi = {10.1016/j.cor.2021.105400},
   issn = {03050548},
   journal = {Computers and Operations Research},
   month = {10},
   publisher = {Elsevier Ltd},
   title = {Reinforcement learning for combinatorial optimization: A survey},
   volume = {134},
   year = {2021}
}

@inproceedings{Vinyals2015,
   author = {Oriol Vinyals and Meire Fortunato and Navdeep Jaitly},
   booktitle = {Advances in Neural Information Processing Systems},
   pages = {2692-2700},
   title = {Pointer Networks},
   year = {2015}
}

@inproceedings{
Pan2025,
title={Preference Optimization for Combinatorial Optimization Problems},
author={Mingjun Pan and Guanquan Lin and You-Wei Luo and Bin Zhu and Zhien Dai and Lijun Sun and Chun Yuan},
booktitle={Forty-second International Conference on Machine Learning},
year={2025},
url={https://openreview.net/forum?id=Jwe5FJ8QGx}
}

@inproceedings{
Liao2025,
title={{BOPO}: Neural Combinatorial Optimization via Best-anchored and Objective-guided Preference Optimization},
author={Zijun Liao and Jinbiao Chen and Debing Wang and Zizhen Zhang and Jiahai Wang},
booktitle={Forty-second International Conference on Machine Learning},
year={2025},
url={https://openreview.net/forum?id=FLy6yXdrlW}
}

@inproceedings{Bello2017,
   author = {Irwan Bello and Hieu Pham and Quoc V Le and Mohammad Norouzi and Samy Bengio},
   booktitle = {5th International Conference on Learning Representations, ICLR 2017 - Workshop Track Proceedings},
   title = {Neural combinatorial optimization with reinforcement learning},
   year = {2017}
}

@inproceedings{Nazari2018,
   author = {Mohammadreza Nazari and Afshin Oroojlooy and Martin Takáč and Lawrence V Snyder},
   booktitle = {Advances in Neural Information Processing Systems},
   pages = {9839-9849},
   title = {Reinforcement learning for solving the vehicle routing problem},
   year = {2018}
}

@article{Correa2026,
title = {TuneNSearch: A hybrid transfer learning and local search approach for solving vehicle routing problems},
journal = {Computers \& Operations Research},
volume = {190},
pages = {107433},
year = {2026},
issn = {0305-0548},
doi = {https://doi.org/10.1016/j.cor.2026.107433},
url = {https://www.sciencedirect.com/science/article/pii/S0305054826000511},
author = {Arthur Corrêa and Cristóvão Silva and Liming Xu and Alexandra Brintrup and Samuel Moniz}
}

@inproceedings{Bi2025,
   author = {Jieyi Bi and Yining Ma and Jianan Zhou and Wen Song and Zhiguang Cao and Yaoxin Wu and Jie Zhang},
   booktitle = {Advances in Neural Information Processing Systems},
   title = {Learning to handle complex constraints for vehicle routing problems},
   year = {2025}
}

@inproceedings{Chalumeau2023,
   author = {Felix Chalumeau and Shikha Surana and Clément Bonnet and Nathan Grinsztajn and Arnu Pretorius and Alexandre Laterre and Thomas D Barrett},
   booktitle = {Advances in Neural Information Processing Systems},
   title = {Combinatorial Optimization with Policy Adaptation using Latent Space Search},
   year = {2023}
}

@inproceedings{Grinsztajn2023,
   author = {Nathan Grinsztajn and Daniel Furelos-Blanco and Shikha Surana and Clément Bonnet and Thomas D Barrett},
   booktitle = {Advances in Neural Information Processing Systems},
   title = {Winner Takes It All: Training Performant RL Populations for Combinatorial Optimization},
   year = {2023}
}

@inproceedings{Zhou2024b,
   author = {Jianan Zhou and Yaoxin Wu and Zhiguang Cao and Wen Song and Jie Zhang and Zhiqi Shen},
   booktitle = {Advances in Neural Information Processing Systems},
   pages = {121731-121764},
   title = {Collaboration! Towards Robust Neural Methods for Routing Problems},
   volume = {37},
   year = {2024}
}

@inproceedings{Kim2022,
   author = {Minsu Kim and Junyoung Park and Jinkyoo Park},
   isbn = {9781713871088},
   booktitle = {Advances in Neural Information Processing Systems},
   title = {Sym-NCO: Leveraging Symmetricity for Neural Combinatorial Optimization},
   year = {2022}
}

@inproceedings{
Drakulic2023,
title={{BQ}-{NCO}: Bisimulation Quotienting for Efficient Neural Combinatorial Optimization},
author={Darko Drakulic and Sofia Michel and Florian Mai and Arnaud Sors and Jean-Marc Andreoli},
booktitle={Thirty-seventh Conference on Neural Information Processing Systems},
year={2023},
url={https://openreview.net/forum?id=BRqlkTDvvm}
}

@article{Zhang2022,
   author = {Yu Zhang and Qiang Yang},
   doi = {10.1109/TKDE.2021.3070203},
   issue = {12},
   journal = {IEEE Transactions on Knowledge and Data Engineering},
   pages = {5586-5609},
   title = {A Survey on Multi-Task Learning},
   volume = {34},
   year = {2022}
}

@inproceedings{Vaswani2017,
   author = {Ashish Vaswani and Noam M Shazeer and Niki Parmar and Jakob Uszkoreit and Llion Jones and Aidan N Gomez and Lukasz Kaiser and Illia Polosukhin},
   booktitle = {Advances in Neural Information Processing Systems},
   pages = {5999-6009},
   title = {Attention is All you Need},
   year = {2017}
}

@misc{Shazeer2020,
      title={GLU Variants Improve Transformer}, 
      author={Noam Shazeer},
      year={2020},
      eprint={2002.05202},
      archivePrefix={arXiv},
      primaryClass={cs.LG},
      url={https://arxiv.org/abs/2002.05202}, 
}

@inbook{Zhang2019rms,
author = {Zhang, Biao and Sennrich, Rico},
title = {Root mean square layer normalization},
year = {2019},
publisher = {Curran Associates Inc.},
address = {Red Hook, NY, USA},
booktitle = {Proceedings of the 33rd International Conference on Neural Information Processing Systems},
articleno = {1110},
numpages = {12}
}

@ARTICLE{Wang2025spsm,
  author={Wang, Yang and Jia, Ya-Hui and Chen, Wei-Neng and Mei, Yi},
  journal={IEEE Transactions on Artificial Intelligence}, 
  title={Soft Parameter Sharing Model for Cross-Problem Generalization in Vehicle Routing Problems}, 
  year={2025},
  volume={},
  number={},
  pages={1-15},
  doi={10.1109/TAI.2025.3576336}}

@inproceedings{
Huang2025rethinking,
title={Rethinking Light Decoder-based Solvers for Vehicle Routing Problems},
author={Ziwei Huang and Jianan Zhou and Zhiguang Cao and Yixin XU},
booktitle={The Thirteenth International Conference on Learning Representations},
year={2025},
url={https://openreview.net/forum?id=4pRwkYpa2u}
}

@inproceedings{
Goh2025shield,
title={{SHIELD}: Multi-task Multi-distribution Vehicle Routing Solver with Sparsity and Hierarchy},
author={Yong Liang Goh and Zhiguang Cao and Yining Ma and Jianan Zhou and Mohammed Haroon Dupty and Wee Sun Lee},
booktitle={Forty-second International Conference on Machine Learning},
year={2025},
url={https://openreview.net/forum?id=6DJEaz1cCj}
}

@inproceedings{
Pan2025decomposable,
title={Multi-Task Vehicle Routing Solver via Mixture of Specialized Experts under State-Decomposable {MDP}},
author={Yuxin Pan and Zhiguang Cao and Chengyang GU and Liu Liu and Peilin Zhao and Yize Chen and Fangzhen Lin},
booktitle={The Thirty-ninth Annual Conference on Neural Information Processing Systems},
year={2025},
url={https://openreview.net/forum?id=ezSyZM6Lp7}
}

@inproceedings{
Liu2025curvature,
title={A Mixed-Curvature based Pre-training Paradigm for Multi-Task Vehicle Routing Solver},
author={Suyu Liu and Zhiguang Cao and Shanshan Feng and Yew-Soon Ong},
booktitle={Forty-second International Conference on Machine Learning},
year={2025},
url={https://openreview.net/forum?id=JsPyLqCgks}
}

@inproceedings{
Zheng2025mtlkd,
title={{MTL}-{KD}: Multi-Task Learning Via Knowledge Distillation for Generalizable Neural Vehicle Routing Solver},
author={Yuepeng Zheng and Fu Luo and Zhenkun Wang and Yaoxin Wu and Yu Zhou},
booktitle={The Thirty-ninth Annual Conference on Neural Information Processing Systems},
year={2025},
url={https://openreview.net/forum?id=rlH3e7VlY8}
}

@inproceedings{Chen2019neurewriter,
   author = {Xinyun Chen and Yuandong Tian},
   booktitle = {Advances in Neural Information Processing Systems},
   title = {Learning to perform local rewriting for combinatorial optimization},
   year = {2019}
}

@inproceedings{Hottung2020,
   author = {André Hottung and Kevin Tierney},
   doi = {10.3233/FAIA200124},
   isbn = {9781643681009},
   booktitle = {European Conference on Artificial Intelligence},
   pages = {443-450},
   title = {Neural large neighborhood search for the capacitated vehicle routing problem},
   year = {2020}
}

@inproceedings{Ma2021dual,
   author = {Yining Ma and Jingwen Li and Zhiguang Cao and Wen Song and Le Zhang and Zhenghua Chen and Jing Tang},
   isbn = {9781713845393},
   booktitle = {Advances in Neural Information Processing Systems},
   pages = {11096-11107},
   title = {Learning to Iteratively Solve Routing Problems with Dual-Aspect Collaborative Transformer},
   year = {2021}
}

@article{Wu2022improvement,
   author = {Yaoxin Wu and Wen Song and Zhiguang Cao and Jie Zhang and Andrew Lim},
   doi = {10.1109/TNNLS.2021.3068828},
   issn = {21622388},
   issue = {9},
   journal = {IEEE Transactions on Neural Networks and Learning Systems},
   month = {9},
   pages = {5057-5069},
   pmid = {33793405},
   publisher = {Institute of Electrical and Electronics Engineers Inc.},
   title = {Learning Improvement Heuristics for Solving Routing Problems},
   volume = {33},
   year = {2022}
}

@inproceedings{Hudson2022,
   author = {Benjamin Hudson and Qingbiao Li and Matthew Malencia and Amanda Prorok},
   booktitle = {ICLR 2022 - 10th International Conference on Learning Representations},
   title = {Graph neural network guided local search
for the traveling salesperson problem},
   year = {2022}
}

@inproceedings{Ma2023search,
   author = {Yining Ma and Zhiguang Cao and Yeow Meng Chee},
   booktitle = {Advances in Neural Information Processing Systems},
   title = {Learning to Search Feasible and Infeasible Regions of Routing Problems with Flexible Neural k-Opt},
   year = {2023}
}

@inproceedings{Roberto20202-opt,
   author = {Paulo Roberto and O Da Costa and Jason Rhuggenaath and Yingqian Zhang and Alp Akcay},
   booktitle = {Proceedings of Machine Learning Research},
   pages = {465-480},
   title = {Learning 2-opt Heuristics for the Traveling Salesman Problem via Deep Reinforcement Learning},
   year = {2020}
}

@InProceedings{Cheng2023select,
  title = 	 {Select and Optimize: Learning to solve large-scale TSP instances},
  author =       {Cheng, Hanni and Zheng, Haosi and Cong, Ya and Jiang, Weihao and Pu, Shiliang},
  booktitle = 	 {Proceedings of The 26th International Conference on Artificial Intelligence and Statistics},
  pages = 	 {1219--1231},
  year = 	 {2023},
  editor = 	 {Ruiz, Francisco and Dy, Jennifer and van de Meent, Jan-Willem},
  volume = 	 {206},
  series = 	 {Proceedings of Machine Learning Research},
  month = 	 {25--27 Apr},
  publisher =    {PMLR},
  url = 	 {https://proceedings.mlr.press/v206/cheng23a.html}
}

@inproceedings{Fu2021generalize,
  title={Generalize a small pre-trained model to arbitrarily large tsp instances},
  author={Fu, Zhang-Hua and Qiu, Kai-Bin and Zha, Hongyuan},
  booktitle={Proceedings of the AAAI conference on artificial intelligence},
  volume={35},
  number={8},
  pages={7474--7482},
  year={2021}
}

@inproceedings{Hou2023generalize,
  title={Generalize learned heuristics to solve large-scale vehicle routing problems in real-time},
  author={Hou, Qingchun and Yang, Jingwei and Su, Yiqiang and Wang, Xiaoqing and Deng, Yuming},
  booktitle={The Eleventh International Conference on Learning Representations},
  year={2023}
}

@article{Li2021delegate,
  title={Learning to delegate for large-scale vehicle routing},
  author={Li, Sirui and Yan, Zhongxia and Wu, Cathy},
  journal={Advances in Neural Information Processing Systems},
  volume={34},
  pages={26198--26211},
  year={2021}
}

@inproceedings{Kim2021collaborative,
   author = {Minsu Kim and Jinkyoo Park and Joungho Kim},
   isbn = {9781713845393},
   booktitle = {Advances in Neural Information Processing Systems},
   pages = {10418-10430},
   title = {Learning Collaborative Policies to Solve NP-hard Routing Problems},
   year = {2021}
}

@article{Zheng2024udc,
  title={UDC: A unified neural divide-and-conquer framework for large-scale combinatorial optimization problems},
  author={Zheng, Zhi and Zhou, Changliang and Xialiang, Tong and Yuan, Mingxuan and Wang, Zhenkun},
  journal={Advances in Neural Information Processing Systems},
  volume={37},
  pages={6081--6125},
  year={2024}
}

@inproceedings{Zong2022rbg,
  title={Rbg: Hierarchically solving large-scale routing problems in logistic systems via reinforcement learning},
  author={Zong, Zefang and Wang, Hansen and Wang, Jingwei and Zheng, Meng and Li, Yong},
  booktitle={Proceedings of the 28th ACM SIGKDD Conference on Knowledge Discovery and Data Mining},
  pages={4648--4658},
  year={2022}
}

@inproceedings{Ye2024glop,
  title={GLOP: Learning Global Partition and Local Construction for Solving Large-scale Routing Problems in Real-time},
  author={Ye, Haoran and Wang, Jiarui and Liang, Helan and Cao, Zhiguang and Li, Yong and Li, Fanzhang},
  booktitle={Proceedings of the AAAI Conference on Artificial Intelligence},
  year={2024},
}

@Article{Ramos2020,
author={Ramos, T{\^a}nia Rodrigues Pereira
and Gomes, Maria Isabel
and Barbosa-P{\'o}voa, Ana Paula},
title={A new matheuristic approach for the multi-depot vehicle routing problem with inter-depot routes},
journal={OR Spectrum},
year={2020},
month={Mar},
day={01},
volume={42},
number={1},
pages={75-110},
issn={1436-6304},
doi={10.1007/s00291-019-00568-7},
url={https://doi.org/10.1007/s00291-019-00568-7}
}

@article{Crevier2007,
title = {The multi-depot vehicle routing problem with inter-depot routes},
journal = {European Journal of Operational Research},
volume = {176},
number = {2},
pages = {756-773},
year = {2007},
issn = {0377-2217},
doi = {https://doi.org/10.1016/j.ejor.2005.08.015},
url = {https://www.sciencedirect.com/science/article/pii/S0377221705006983},
author = {Benoit Crevier and Jean-François Cordeau and Gilbert Laporte}
}

@ARTICLE{Li2024mdmta,
  author={Li, Jinqi and Tian Dai, Bing and Niu, Yunyun and Xiao, Jianhua and Wu, Yaoxin},
  journal={IEEE Transactions on Intelligent Transportation Systems}, 
  title={Multi-Type Attention for Solving Multi-Depot Vehicle Routing Problems}, 
  year={2024},
  volume={25},
  number={11},
  pages={17831-17840},
  doi={10.1109/TITS.2024.3413077}}

@article{Wouda2024,
  doi = {10.1287/ijoc.2023.0055},
  url = {https://doi.org/10.1287/ijoc.2023.0055},
  year = {2024},
  volume = {36},
  number = {4},
  pages = {943--955},
  publisher = {INFORMS},
  author = {Niels A. Wouda and Leon Lan and Wouter Kool},
  title = {{PyVRP}: a high-performance {VRP} solver package},
  journal = {INFORMS Journal on Computing},
}

@article{Uchoa2017,
   author = {Eduardo Uchoa and Diego Pecin and Artur Pessoa and Marcus Poggi and Thibaut Vidal and Anand Subramanian},
   doi = {10.1016/j.ejor.2016.08.012},
   issn = {03772217},
   issue = {3},
   journal = {European Journal of Operational Research},
   month = {3},
   pages = {845-858},
   title = {New benchmark instances for the Capacitated Vehicle Routing Problem},
   volume = {257},
   year = {2017},
}

@article{vanDerMaaten2008,
  author = {van der Maaten, Laurens and Hinton, Geoffrey},
  journal = {Journal of Machine Learning Research},
  pages = {2579--2605},
  title = {Visualizing Data using {t-SNE} },
  url = {http://www.jmlr.org/papers/v9/vandermaaten08a.html},
  volume = 9,
  year = 2008
}

@article{Cordeau1997,
  title={A tabu search heuristic for periodic and multi-depot vehicle routing problems},
  author={Jean-François Cordeau and Michel Gendreau and Gilbert Laporte},
  journal={Networks},
  year={1997},
  volume={30},
  pages={105-119}
}

@article{Braekers2016,
title = {The vehicle routing problem: State of the art classification and review},
journal = {Computers \& Industrial Engineering},
volume = {99},
pages = {300-313},
year = {2016},
issn = {0360-8352},
doi = {https://doi.org/10.1016/j.cie.2015.12.007},
url = {https://www.sciencedirect.com/science/article/pii/S0360835215004775},
author = {Kris Braekers and Katrien Ramaekers and Inneke {Van Nieuwenhuyse}}
}

@article{Pillac2013,
title = {A review of dynamic vehicle routing problems},
journal = {European Journal of Operational Research},
volume = {225},
number = {1},
pages = {1-11},
year = {2013},
issn = {0377-2217},
doi = {https://doi.org/10.1016/j.ejor.2012.08.015},
url = {https://www.sciencedirect.com/science/article/pii/S0377221712006388},
author = {Victor Pillac and Michel Gendreau and Christelle Guéret and Andrés L. Medaglia}
}

@inproceedings{Yu2020,
author = {Yu, Tianhe and Kumar, Saurabh and Gupta, Abhishek and Levine, Sergey and Hausman, Karol and Finn, Chelsea},
title = {Gradient surgery for multi-task learning},
year = {2020},
isbn = {9781713829546},
publisher = {Curran Associates Inc.},
address = {Red Hook, NY, USA},
booktitle = {Proceedings of the 34th International Conference on Neural Information Processing Systems},
articleno = {489},
numpages = {13},
location = {Vancouver, BC, Canada},
series = {NIPS '20}
}

@inproceedings{
Ha2017,
title={HyperNetworks},
author={David Ha and Andrew M. Dai and Quoc V. Le},
booktitle={International Conference on Learning Representations},
year={2017},
url={https://openreview.net/forum?id=rkpACe1lx}
}

@article{Jiang2026,
title = {Dual-branch distance-aware neural solver for cross-distribution vehicle routing problems},
journal = {Neurocomputing},
volume = {679},
pages = {133239},
year = {2026},
issn = {0925-2312},
doi = {https://doi.org/10.1016/j.neucom.2026.133239},
url = {https://www.sciencedirect.com/science/article/pii/S0925231226006363},
author = {Yi-Bo Jiang and Ding-Yue Tao and Wei-Jie Chen and Fan-Rong Xu}
}

@article{Lei2022,
title = {Solve routing problems with a residual edge-graph attention neural network},
journal = {Neurocomputing},
volume = {508},
pages = {79-98},
year = {2022},
issn = {0925-2312},
doi = {https://doi.org/10.1016/j.neucom.2022.08.005},
url = {https://www.sciencedirect.com/science/article/pii/S092523122200978X},
author = {Kun Lei and Peng Guo and Yi Wang and Xiao Wu and Wenchao Zhao}
}



\end{document}